\documentclass[11pt]{article}

\usepackage[preprint]{acl}

\definecolor{submission}{HTML}{000099}

\usepackage{times}
\usepackage{latexsym}

\usepackage[T1]{fontenc}

\usepackage[utf8]{inputenc}

\usepackage{microtype}

\usepackage{inconsolata}

\usepackage{graphicx}

\usepackage{amssymb}
\usepackage{amsmath}
\usepackage{multirow}
\usepackage{booktabs}
\usepackage{multicol}   
\usepackage{colortbl}   
\usepackage{xcolor}     
\usepackage{arydshln}
\usepackage{amssymb}
\usepackage{pifont}
\usepackage{multicol}
\usepackage{algorithm}
\usepackage[noend]{algpseudocode}
\usepackage{arydshln}
\usepackage{booktabs}
\usepackage{makecell}
\usepackage{amsmath, amssymb}
\usepackage{multirow}
\usepackage{adjustbox}
\usepackage{tabularx}
\usepackage{enumitem}
\usepackage{minted} 
\usepackage{url}
\usepackage{cuted}
\usepackage[most]{tcolorbox}
\usepackage{xcolor}
\usepackage{courier}
\usepackage{caption}

\title{Lost in Recovery: Demystifying Recovery Behavior of\\ Computer-Use Agents via Goal-Oriented Action Replay}

\author{
Byeonghun Bang\textsuperscript{1}\thanks{Equal contribution.} \quad
Sangwoo Park\textsuperscript{1}\footnotemark[1] \quad
Jinheon Baek\textsuperscript{1} \quad
Kangsan Kim\textsuperscript{1} \quad \\
\textbf{Youngwan Lee\textsuperscript{1,2} \quad
Sung Ju Hwang\textsuperscript{1,3}} \\[0.5em]
\textsuperscript{1}KAIST\quad
\textsuperscript{2}ETRI\quad
\textsuperscript{3}DeepAuto.ai\\[0.5em]
\fontsize{9.5pt}{9.5pt}\selectfont\texttt{\{bh.bang, jackson0530, jinheon.baek, kangsan.kim, sungju.hwang\}@kaist.ac.kr
}\\
\fontsize{9.5pt}{9.5pt} \texttt{yw.lee@etri.re.kr
}
}

\begin{document}
\maketitle

\begin{abstract}
Recent advances in large (vision) language models have enabled computer-use agents that automate digital workflows via fine-grained keyboard and mouse operations. However, such agents remain fragile in long-horizon settings, where errors accumulate and executed actions leave unintended side effects, making the ability to recover from failures critical. Yet, existing remedies remain inadequate: environment resets are infeasible in proprietary settings, action replay is prohibitively expensive over long horizons, and API-based rollback is limited to web environments, leaving model-based recovery largely unexplored. To address this gap, we present \textsc{ReTrace}, a scalable evaluation framework for systematically studying recovery behavior in computer-use agents. We formalize \textit{Transition-Consistent Recovery}: identical downstream actions from the recovered and target states must yield identical outcomes. We operationalize this criterion through goal-oriented action replay following controlled error injection. Using \textsc{ReTrace}, we present the first systematic study of autonomous, model-based recovery, revealing that (i) ideal recovery can unlock model's latent task completion capabilities, (ii) current computer-use agents fail to exploit their latent recovery capabilities, and (iii) accurate model-based recovery yields measurable end-to-end gains. Together, these findings establish recovery as a fundamental capability for building reliable computer-use agents. Our code and dataset is available at \url{https://github.com/BangProx/ReTRACE}.
\end{abstract}

\section{Introduction}
The digital computing environment has become the primary interface through which users interact with software, supporting a wide spectrum of tasks ranging from simple actions (e.g., website logins) to complex workflows (document editing). Recent advances in large (vision) language models~\citep{bai2025qwen3, Singh2025OpenAIGS, Guo_2025, claude_sonnet} have made it possible to automate such workflows through \emph{computer-use agents}, which act in digital environments via fine-grained keyboard and mouse inputs~\citep{gur2023real, furuta2023multimodal, sager2025ai}.

\begin{figure}[t]
    \centering
    \includegraphics[]{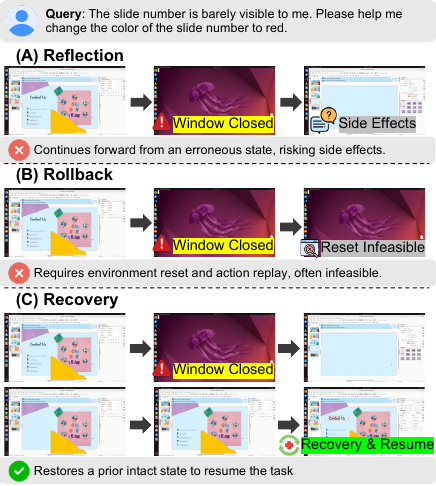}
    \vspace{-0.1in}
    \caption{Comparison of error-handling strategies in computer-use agents. (A) Reflection proceeds from the erroneous state, (B) rollback requires often-infeasible environment resets, while (C) recovery restores a prior intact state and resumes the task.}
    \label{fig:conceptual_figure.pdf}
    \vspace{-0.2in}
\end{figure}

However, computer-use agents remain fragile in long-horizon and sophisticated tasks, where early mistakes accumulate and propagate across subsequent steps~\citep{wang2026long}, and where executed actions may leave unintended side effects that should not have occurred (e.g., an agent organizing a directory may complete the task but silently overwrite files that the user wanted to keep). As a result, the ability to revert the environment to a previously intact state is critical for reliable operation. To address this, prior work explores two directions: inducing self-reflective behaviors through textual feedback or parameter updates from reward signals that refine forward actions~\citep{chae2025web, chae2025websheperd, wu2025backtrackagent, wang2025ui, zhang2026webarbiter, xue2026evocua, wang2026history}, and explicit test-time rollback through full environment resets followed by re-executing all prior actions (i.e., action replay)~\citep{koh2024tree, he2025branch} or API-based undo operations in web environments (e.g., browser back)~\citep{hu-etal-2025-webcot, zhang2026webrollback}, utilized as a heuristic component within an inference pipeline rather than subjected to systematic study.

Yet, these heuristic remedies fall short of providing general-purpose recovery. Environment resets are often infeasible in proprietary settings such as production websites and operating systems~\citep{regatti2021offline, nguyen2023provable, wang2026agent}, action replay~\citep{koh2024tree, he2025branch} inherits this requirement and becomes costly over long horizons, and API-based rollback~\citep{hu-etal-2025-webcot, zhang2026webrollback} is largely confined to web environments, leaving broader desktop settings uncovered.  Consequently, \textit{autonomous model-level recovery}, in which the agent itself restores a prior state without relying on such ad hoc mechanisms, emerges as an essential capability. However, whether current agents possess recovery capabilities, and how it affects overall agent performance remain largely unexplored. Therefore, we investigate several fundamental research questions: 

\vspace{0.2em}
\noindent\textbf{\textit{RQ1.}} \textit{Is recovery necessary to unlock models' latent capabilities?}

\vspace{0.2em}

\noindent\textbf{\textit{RQ2.}} \textit{Do current computer-use agents possess recovery capabilities under erroneous conditions, and are such capabilities latent or entirely absent?}

\vspace{0.2em}

\noindent\textbf{\textit{RQ3.}} \textit{Does accurate model-level recovery translate into measurable improvements in task completion?}

\vspace{0.2em}
\begin{figure*}[t!]
  \centering
  \includegraphics[width=0.97\textwidth]{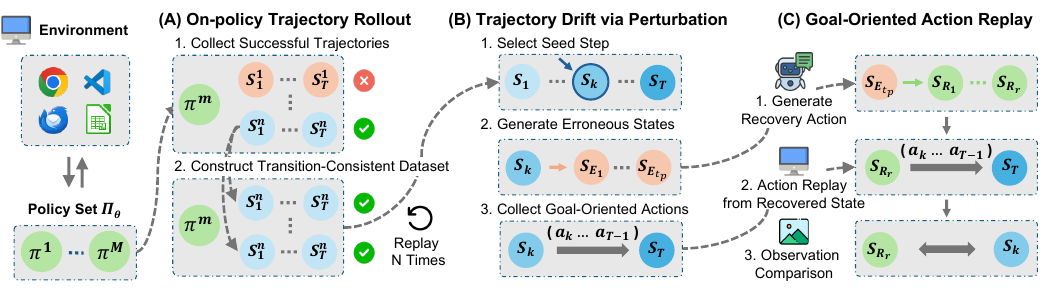}
  \vspace{-0.15in}
  \caption{Overview of \textsc{ReTrace}. (A) On-Policy Trajectory Rollout collects successful trajectories and replays each action sequence $N$ times to build a transition-consistent dataset. (B) Trajectory Drift via Perturbation selects a recovery target and perturbs the agent into an erroneous state. (C) Goal-Oriented Action Replay replays the remaining actions from the recovered state to test whether the agent reproduces the original task outcome.}
  \label{fig:main_figure}
  \vspace{-0.1in}
\end{figure*}
To address this, we introduce \textsc{ReTrace}, a pipeline that conducts the first systematic study of \textit{autonomous, model-based recovery} behavior in computer-use agents. Motivated by predictive state representation theory~\citep{littman2001predictive}, which represents states by the formalization of predicted futures, we formalize successful recovery as \emph{Transition-Consistent Recovery}: a recovered state matches its target if executing identical downstream actions from both states yields identical outcomes. This formulation captures latent changes beyond visual similarity by leveraging behavioral equivalence, operationalized via \emph{goal-oriented action replay}. Specifically, we replay each on-policy trajectory multiple times and retain only those with consistent successful outcomes, ensuring deterministic environment transitions. From these trajectories, we sample intermediate states as recovery targets, perturb them into erroneous states, and evaluate whether the original task outcome can be reproduced by replaying the remaining actions.

We then instantiate \textsc{ReTrace} on OSWorld~\citep{OSWorld} and WindowsAgentArena~\citep{bonatti2024windows} across closed- and open-source computer-use agents, and find that recovery plays a critical role in reliable agent operation. We first show that \emph{ideal recovery, in which agents are returned to an oracle prior state, boosts performance over naive environment resets (Section~\ref{main:ideal_gains_from_recovery}), identifying recovery as the key lever for unlocking agents' latent task-solving capacity (\textbf{RQ1})}. Yet current agents leave this lever unused, \emph{lacking both the tendency and capability to recover autonomously (Section~\ref{main:recovery_challenge}) (\textbf{RQ2})}. When equipped with explicit model-based recovery, however, recovery success climbs by more than an order of magnitude, revealing that the \emph{underlying capability is latent rather than absent (Section \ref{main:recovery_challenge}) \textbf{(RQ2)}}. Moreover, recovery quality in turn translates into downstream task success: \emph{accurate model-level recovery delivers measurable end-to-end gains by unlocking otherwise latent task-solving capacity, while weak recovery can actively derail the trajectory (\textbf{RQ3}) (Section~\ref{main:end2end_evaluation})}. These findings position accurate model-based recovery, together with reliable continuation from the recovered state, as a key step toward stronger computer-use agents.
\section{Related Works}

\paragraph{Computer-Use Agents}
Computer-use agents aim to complete diverse user tasks across web-based environments \citep{gur2023real, furuta2023multimodal, openai_operator, awadallah2025fara}, desktop operating systems such as Linux and Windows \citep{wang2025opencua, yang2025gta1, claude_sonnet}, and mobile environments \citep{ye2025mobile, xu2026mobile}. Pioneering works leverage LLMs over HTML DOM content \citep{liu2018reinforcement, kim2023language, zhou2023webarena}, while recent studies operate LVLMs over screenshot observations \citep{koh2024visualwebarena, bai2025qwen3, learnweak, wang2026history} and further reinforce backbone capabilities through reinforcement learning \citep{wang2025ui, xu2025mobilerl, xue2026evocua} or external memories \citep{zheng2023synapse, fu2024autoguide}. However, these efforts concentrate on strengthening the \emph{forward} decision policy, leaving long-horizon brittleness, where early mistakes accumulate and propagate downstream \citep{wang2026long}, largely unaddressed. In contrast, we focus on the systematic analysis of \emph{recovery} behavior in computer-use agents, a complementary but underexplored axis.

\paragraph{Agent Behaviors under Failure}
Prior work explores diverse error-handling strategies across complex interactive tasks. For example, \citet{lightman2024let, shinn2023reflexion, xi2024enhancingllmreasoningcritique} use outcome and process-level feedback to refine actions at test time in mathematical and code tasks \citep{Chen2022CodeTCG}. In long-horizon embodied settings, \citet{huang2022inner, wang2023voyager, wang2023describe} leverage external rewards to revise plans or acquire new skills, and \citet{li-etal-2025-generator} perform rollback through resetting the environment and executing identical actions. This problem has also been explored in computer-use agents: \citet{wang2024mobile, chae2025websheperd, wu2025backtrackagent} use external verifiers to induce reflective behaviors \citep{hu-etal-2025-webcot} for effective user task completion, \citet{koh2024tree, he2025branch} employ action replay to revert to prior states upon failure, and \citet{hu-etal-2025-webcot, zhang2026webrollback} utilize explicit rollback via go-back operations in Playwright, while \citet{lu2026beap} introduce semantic actions for state recovery within the DFS algorithm. However, previous works used these mechanisms as heuristics in inference pipelines rather than systematically analyzing their behavior and effects. In contrast, we uncover key findings on model-level recovery in computer-use agents using \textsc{ReTrace}, a scalable pipeline enabling the first systematic analysis of recovery behavior.

\section{Preliminary}
\label{preliminary:computer_agent_task_defintion}

\paragraph{Computer-Use Task} 
We formulate a computer-use agent (CUA) task as a Partially Observable Markov Decision Process~\citep{kaelbling1998planning}, $\mathcal{M} = (\mathcal{S}, \mathcal{A}, \mathcal{O}, \mathcal{E}, \mathcal{T}, \Omega)$, where a policy $\pi_\theta$ interacts with a digital environment $\mathcal{E}$ (e.g., Linux terminals, PowerPoint) to complete a user query $q$. Each state $s_t = (h_t, o_t) \in \mathcal{S}$ decomposes into a hidden component $h_t$ and an observation $o_t = \Omega(s_t) \in \mathcal{O}$, satisfying the Markov property. At each step, $\pi_\theta$ selects an action $a_t \in \mathcal{A}$, spanning high-level (\texttt{CLICK}) to low-level (\texttt{pyautogui.click}) operations, yielding $s_{t+1} \sim \mathcal{T}(\cdot \mid s_t, a_t)$.

\paragraph{Forward Policy and Trajectory} 
Since $\mathcal{E}$ is only partially observable, $\pi_\theta$ conditions on the observation–action history, $\pi_\theta(a_t \mid o_0, a_0, \dots, o_t)$, yielding a trajectory $\tau = \{(o_t, a_t)\}_{t=0}^{T}$. However, this forward formulation leaves unspecified how the agent should behave once early actions drive $\tau$ into an erroneous state from which $q$ can no longer be completed, the setting we study next.
\section{Accurate Recovery Matters}
\label{main:ideal_gains_from_recovery}
Here, we show through a preliminary experiment that accurate recovery yields measurable gains, establishing it as a powerful lever for reliable agents. For more details, please see Appendices \ref{appendix:ideal_gains_from_recovery} and \ref{appendix:Prompts}.

\paragraph{Expected Gains from Ideal Recovery}
Agents exhibit diverse trajectories across repeated rollouts of the same task, and this variance may imply substantial latent task-solving capability that any single rollout fails to exploit, motivating our hypothesis: reverting to an appropriate past state (\emph{recovery step}) can convert it into task success.
To validate this, we measure Success@5, the success rate over 5 independent continuations, when the forward policy resumes from different recovery steps.

\begin{figure}[t]
    \centering
    \includegraphics[width=0.975\columnwidth]{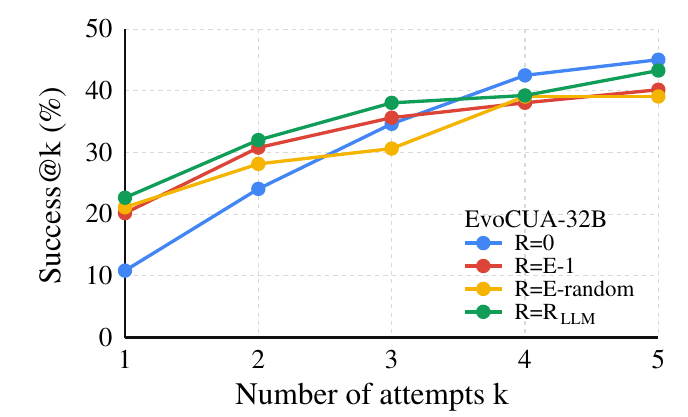}
    \vspace{-0.05in}
    \caption{Success@$\mathrm{k}$ across different recovery steps $\mathrm{R}$ under ideal recovery, using EvoCUA-32B as the forward policy on OSWorld.}
    \vspace{-0.1in}
    \label{fig:ideal_recovery_gain}
\end{figure}

\paragraph{Setup}
\label{main:ideal_gain_recovery_setup}
We focus on tasks that the forward policy fails to solve under standard rollouts, the cases where recovery can yield measurable gains, with EvoCUA-32B~\citep{xue2026evocua} as the forward policy on the OSWorld~\citep{OSWorld} benchmark. For each failed trajectory, we annotate the \emph{error step} $\mathrm{E}$ (where the agent first goes off track) using an LLM-based detector, and set the recovery step $\mathrm{R}$, from which the forward policy resumes, to one of: the initial state ($\mathrm{R=0}$, equivalent to a full environment reset), one step before the error ($\mathrm{R=E\text{-}1}$), a uniformly sampled past step ($\mathrm{R=E\text{-}\text{random}}$), or a step identified by an LLM-based recovery step detector ($\mathrm{R=R_{\text{LLM}}}$). We use Qwen3.6-Plus~\citep{qwen36plus} as the backbone for both detectors.

\paragraph{Results}
As shown in Figure~\ref{fig:ideal_recovery_gain}, every non-zero recovery strategy outperforms complete environment resets ($\mathrm{R=0}$), with the largest advantage observed in the low-$\mathrm{K}$ regime. This demonstrates that, \textit{under the same retry budget, recovering from the previous state utilizes each retry substantially more efficiently than restarting the task from scratch \textbf{(Finding 1)}}, resulting in significantly higher Success@$\mathrm{K}$. As $\mathrm{K}$ increases, the gap naturally diminishes because repeated restarts also accumulate additional opportunities for success. These findings therefore highlight the efficiency of accurate recovery in utilizing model's internal capabilities under constrained retry budgets.
\section{How Well Can Agents Recover?}
\label{main:recovery_challenge}

Here, we analyze the recovery behavior of current computer-use agents using the empirical pipeline illustrated in Figure \ref{fig:main_figure}. For more details, please see Appendices \ref{appendix:ReTrace}, \ref{appendix:human_evaluation}, \ref{appendix:Supplementary Experiments}, and \ref{appendix:Prompts}.
\subsection{Transition-Consistent Recovery}
\label{main:state_recovery_formulation}
Given a trajectory $\tau = \{(s_t, a_t)\}_{t=0}^{T}$ produced by a policy $\pi$, the recovery task is defined as a function $f$ that maps an arbitrary state $s_t$ back to a past target state, $f(s_t) = s_\mathrm{target} = s_{t-\mathrm{d}}$, where $\mathrm{d}$ is the recovery distance. Whether this mapping succeeds is judged by a predefined recovery oracle $\mathcal{R}$, which returns a binary indicator of recovery success or failure, $\mathcal{R}(f(s_t), s_\mathrm{target}) \in \{0, 1\}$. Recalling from Section \ref{preliminary:computer_agent_task_defintion} that each state decomposes as $s_t = (h_t, o_t)$ into a hidden component $h_t$ and an observation $o_t$, $\mathcal{R}$ should verify that $f(s_t)$ matches $s_\mathrm{target}$ in both components.
A natural instantiation of $\mathcal{R}$ is to compare the observations of $f(s_t)$ and $s_\mathrm{target}$ directly, e.g., via pixel- or feature-level matching. However, this surface criterion is unreliable, since latent components of the environment may be implicitly altered without any visible change~\citep{shani2004resolving} (e.g., a hidden configuration file modified while the screenshot remains identical), causing observation-only oracles to falsely declare success on states that nonetheless diverge in their downstream effects. To capture such latent discrepancies, we draw on the theory of predictive state representations~\citep{littman2001predictive}, which defines state equivalence through identical transition dynamics, and formalize \textit{transition-consistent recovery}. Specifically, in a controlled environment $\mathcal{E}$ with a deterministic transition function $\mathcal{T}_{\mathcal{E}}$, we declare a recovery successful when executing the same downstream action $a_{t-\mathrm{d}}$ from the recovered state reproduces the next state along the original trajectory, defined as $\mathcal{R}_{\mathrm{TC}} = \mathbb{I}\big[\mathcal{T}_{\mathcal{E}}(f(s_t), a_{t-\mathrm{d}}) = s_{t-\mathrm{d}+1}\big]$ for $(s_{t-\mathrm{d}}, a_{t-\mathrm{d}}, s_{t-\mathrm{d}+1}) \in \tau$, which could extend to the multi-step setting by recursively chaining transitions along $\tau$, as $\mathcal{R}_{\mathrm{TC}} = \mathbb{I}\big[\mathcal{T}_{\mathcal{E}}(\cdots \mathcal{T}_{\mathcal{E}}(f(s_\mathrm{t}), a_{\mathrm{t-d}}) \cdots, a_{\mathrm{t-1}}) = s_\mathrm{t}\big].$

To complement this hidden-state check, we additionally enforce observation-level agreement to guard against cases where the visible interface itself diverges from the target. Formally, we define $\mathcal{R}_{\mathrm{O}} = \mathbb{I}[\, o = \hat{o} \,]$, where $(h, o) = s_\mathrm{target}$ and $(\hat{h}, \hat{o}) = f(s_t)$. The final recovery oracle then combines both, $\mathcal{R} = \mathcal{R}_{\mathrm{TC}} \wedge \mathcal{R}_{\mathrm{O}}$, jointly evaluating recovery on the hidden and observable axes.

\subsection{Recovery Evaluation}

\textsc{ReTrace} evaluates recovery in three stages (Figure \ref{fig:main_figure}): the first two construct a transition-consistent evaluation set, and the third measures recovery on it via goal-oriented action replay.
\paragraph{On-Policy Trajectory Rollout} 
As the first stage of \textsc{ReTrace} (Figure \ref{fig:main_figure} (A)), evaluating transition-consistent recovery requires a target state $s_{\mathrm{target}}$ and an error state $s_{\mathrm{error}}$ for a trajectory $\tau$, together with a deterministic transition function $\mathcal{T}$. To probe each policy on its own behavior, $\tau$ must remain on-policy with the evaluated $\pi_{\theta}$. We thus begin by rolling out each $\pi_{\theta} \in \Pi$ on every user query $\mathrm{q} \in \mathcal{Q}$, collecting ground-truth trajectories $\tau_{\mathrm{gt}} = \{(s_t, a_t)\}_{t=0}^{T}$ whose success is determined by an oracle $\Phi_{\mathrm{q}}$. For each $\tau_{\mathrm{gt}}$, we sample seed time steps $t_{\mathrm{seed}} \in \{1,\dots,T\}$ and retain only successful trajectories with $\Phi_{\mathrm{q}}(\tau_{\mathrm{gt}}) = 1$, forming an evaluation set $\mathcal{D}$ for each $\pi_{\theta}$.

We further require a reliable transition function, grounded in the formal definition of state recovery in Section~\ref{main:state_recovery_formulation}. To this end, inspired by prior work that approximates the latent model distribution through iterative sampling of model outputs~\citep{wu2025invisible}, we construct an amortized deterministic transition function $\hat{\mathcal{T}}$ for a given action trajectory $\tau_{\mathrm{gt}}$. Specifically, we replay the identical action sequence $N$ times and retain only those that yield $N$ successful outcomes, thereby ensuring reliable state transitions. This finally yields a transition-consistent trajectory dataset $\mathcal{D}_{\mathrm{TC}}$ for each policy $\pi_{\theta} \in \Pi$.
\begin{figure*}[t!]
\centering
\includegraphics[width=\textwidth]{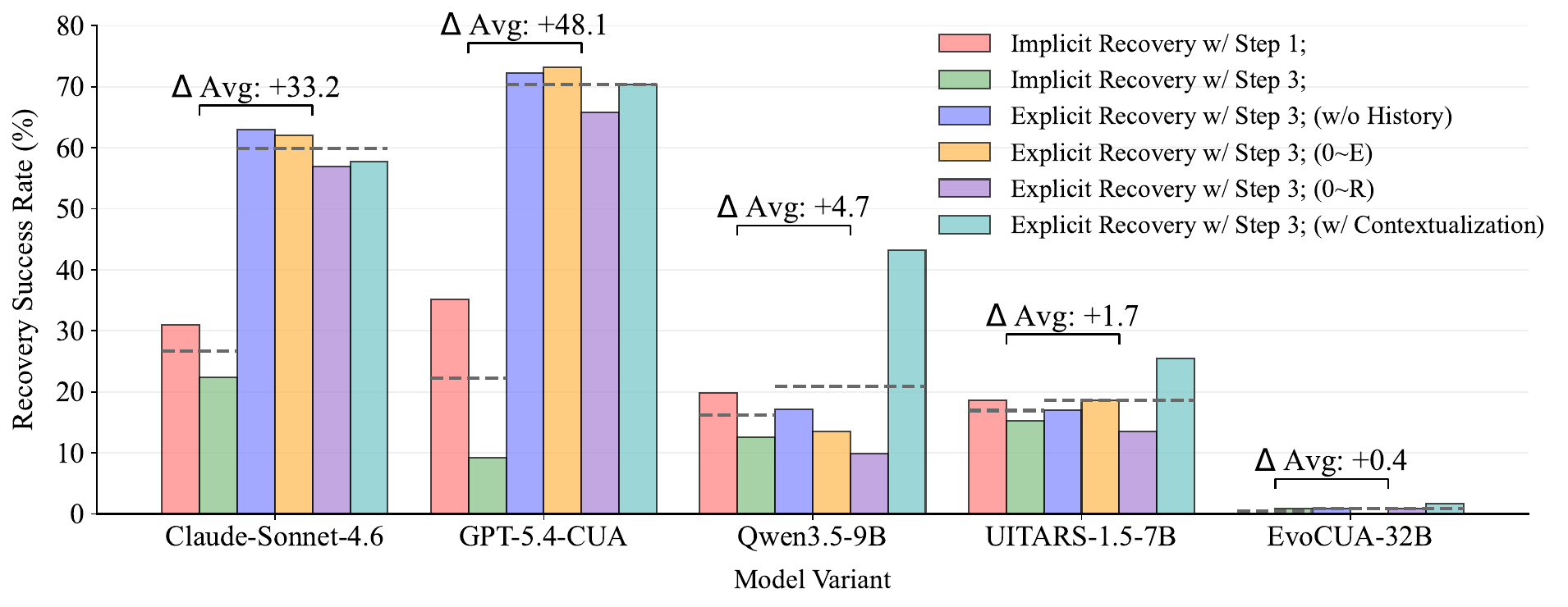} 
\vspace{-0.35in}
\captionof{figure}{Comparison of short-horizon recovery strategies across computer-use agents on OSWorld.}
\label{fig:osworld_shorthorizon}

\begin{minipage}[c]{0.32\textwidth}
    \centering
    \includegraphics[width=\textwidth]{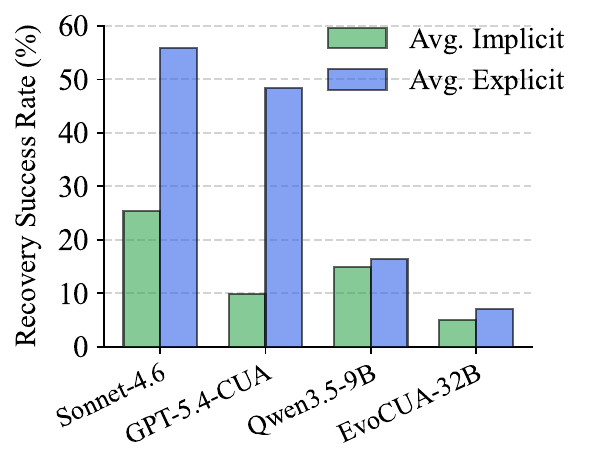} 
    \vspace{-0.3in}
    \captionof{figure}{Comparison of short-horizon recovery strategies across models on WindowsAgentArena.}
    \label{fig:warena_shorthorizon}
\end{minipage}
\hfill
\begin{minipage}[c]{0.32\textwidth}
    \centering
    \includegraphics[width=\textwidth]{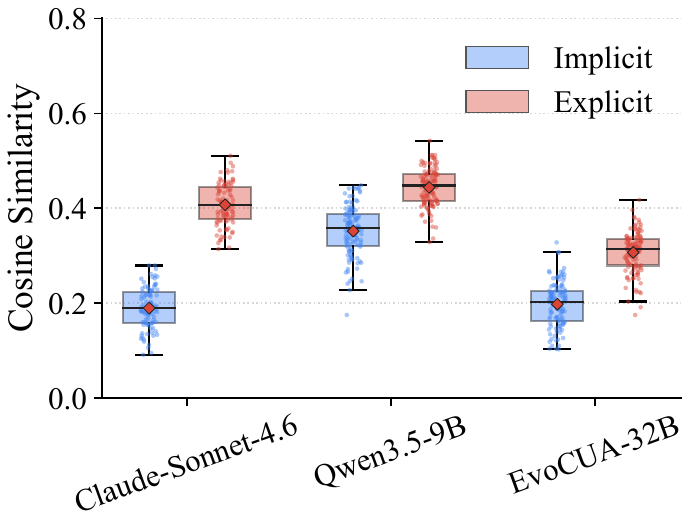} 
    \vspace{-0.3in}
    \captionof{figure}{Recovery intention under implicit vs explicit recovery, measured by the similarity between agent rationales and recovery prompts.}
    \label{fig:recovery_intention}
\end{minipage}
\hfill
\begin{minipage}[c]{0.31\textwidth}

    \vspace{0.1in}
    \setlength{\tabcolsep}{1pt}
    \renewcommand{\arraystretch}{1.4}

    {\small
    \resizebox{\textwidth}{!}{%
    \begin{tabular}{lc}
    \toprule
    \textbf{\small Baselines}  & \textbf{\small Win Rate (\%)} \\
    \midrule
    \textbf{w/o $\mathcal{R}_{\mathrm{O}}$} & 82.0\% \\
    \textbf{w/o $\mathcal{R}_{\mathrm{TC}}$} & 63.0\% \\
    \textbf{GPT-5.4 Evaluator} & 54.0\% \\
    \bottomrule
    \end{tabular}%
    }   
    }
    \captionof{table}{Average win rate of \textsc{ReTrace} against alternative evaluation strategies in matching human judgments, where human annotations serve as the ground truth. For more details, see Appendix \ref{appendix:human_evaluation}.}
    \label{tab:evaluation_preference_eval}
\end{minipage}
\vspace{-0.15in}
\end{figure*}

\paragraph{Trajectory Drift via Perturbation} Given the transition-consistent dataset, for $(\tau_{\mathrm{gt}}, t_{\mathrm{seed}}, \hat{\mathcal{T}}) \in \mathcal{D}_{\mathrm{TC}}$ we next require a ground-truth error state $s_{\mathrm{error}}$ (Figure \ref{fig:main_figure} (B)). Taking $s_{t_{\mathrm{seed}}} \in \tau_{\mathrm{gt}}$ as the target state $s_{\mathrm{target}}$, we obtain $s_{\mathrm{error}}$ by drifting the trajectory into erroneous conditions with respect to the user query $\mathrm{q}$ via a perturbation policy $\pi_{\theta}^{\mathrm{p}}$. Concretely, we first replay the action sequence of $\tau_{\mathrm{gt}}$ in $\mathcal{E}$ using $\hat{\mathcal{T}}$ up to step $t_{\mathrm{seed}}$, with the LLM verifier $h_{\theta}$ confirming that $s_{t_{\mathrm{seed}}}$ is reproduced. From this state, $\pi_{\theta}^{\mathrm{p}}$ rolls out a perturbation trajectory $\tau_{\mathrm{p}} = \big\{(s_{t_{\mathrm{seed}}},\; a_{t_{\mathrm{seed}}}),\; \ldots,\; (s_{t_{\mathrm{seed}} + t_\mathrm{p}}) \big\}$ up to the perturbation limit $t_\mathrm{p}$. The oracle $\Phi_{\mathrm{q}}$ then assigns the final state $s_{t_{\mathrm{seed}} + t_\mathrm{p}}$ as $s_{\mathrm{error}}$ when it is erroneous, yielding the transition-consistent recovery evaluation set $\mathcal{D}_{\mathrm{R}}$ for each policy $\pi_{\theta} \in \Pi$. By construction, every instance in $\mathcal{D}_{\mathrm{R}}$ pairs a verified off-track error state $s_{\mathrm{error}}$ with its reachable target $s_{\mathrm{target}}$, enabling a controlled evaluation of agent recovery under genuine failure.

\paragraph{Goal-Oriented Action Replay} 
Given the recovery evaluation set $\mathcal{D}_{\mathrm{R}}$, we initiate recovery from each $s_{\mathrm{error}} \in \mathcal{D}_{\mathrm{R}}$, which serves as the starting state for the recovery policy $\pi_{\theta}^{\mathrm{R}}$ (Figure \ref{fig:main_figure} (C)). Since $s_{\mathrm{error}}$ is conditioned on $s_{\mathrm{target}}$, we first replay $\tau_{\mathrm{gt}}$ up to $s_{\mathrm{target}}$ and then $\tau_{\mathrm{p}}$ for $t_{\mathrm{p}}$ steps, using the LLM verifier $h_{\theta}$ to confirm faithful reconstruction of $s_{\mathrm{target}}$ and $s_{\mathrm{error}}$, respectively. With $s_{\mathrm{error}}$ thus reconstructed, the recovery policy $\pi_{\theta}^{\mathrm{R}}$ executes from it for at most $\mathrm{r}$ steps, producing the recovery trajectory $\tau_{\mathrm{R}} = \{(s_t, a_t)\}_{t=0}^{T}$ and the recovered state $s_T$. However, single-step transition-consistent recovery is impractical, as defining oracle annotations for numerous intermediate states is infeasible. Therefore, following the multi-step formulation of transition-consistent recovery in Section~\ref{main:state_recovery_formulation}, we instead leverage the pre-configured oracle $\Phi$ for user-query completion, extending the objective to test whether replaying the remaining actions of $\tau_{\mathrm{gt}}$ from the recovered state reproduces the original task outcome. The recovery oracle is thus defined as $\mathcal{R}_{\mathrm{TC}}=\mathbb{I}\big[\Phi_{\mathrm{q}}(\mathcal{T}_{\mathcal{E}}(\cdots \mathcal{T}_{\mathcal{E}}(f(s_t), a_\mathrm{t-d}), \ldots, a_T))=1\big]$, together with $\mathcal{R}_{\mathrm{O}}$ for observation-level matching.

\subsection{Recovery Policy}

Given an error state, a recovery policy acts to restore a past target state. We study two instantiations, differing in whether recovery is elicited from the forward agent itself or from a dedicated policy.

\paragraph{Implicit Recovery} 
The implicit recovery policy reuses the forward decision policy as is, i.e., $\pi_{\theta}^{\mathrm{R}}=\pi_{\theta}$, applying no recovery-specific instruction or parameter update. The same agent simply continues to act from the error state, which probes whether it spontaneously produces restorative actions toward prior states as an intrinsic behavior. 

\paragraph{Explicit Recovery} 
In contrast, the explicit recovery policy $\pi_{\theta}^{\mathrm{R}}$ is an external policy specifically designed to revert the agent to a target recovery state $s_\mathrm{target}$. Formally, this recovery process is defined as $\pi_{\theta}^{\mathrm{R}}(a_t \mid s_t, s_\mathrm{target})$, which, following the POMDP formulation in Section~\ref{preliminary:computer_agent_task_defintion}, can be expressed as $\pi_{\theta}^{\mathrm{R}}(a_t \mid o_\mathrm{target}, (o_0, a_0, \dots, o_t))$. Unlike the forward decision policy that begins from a fixed starting point, recovery may start from an arbitrary state along the forward trajectory. Moreover, since the target state itself lies within that trajectory, its hidden component cannot be inferred from $o_\mathrm{target}$ alone; we thus introduce an additional recovery context $\mathcal{C}$ to represent the target hidden state $h_\mathrm{target} \in s_\mathrm{target}$, leading to the final explicit recovery formulation $\pi_{\theta}^{\mathrm{R}}(a_t \mid o_\mathrm{target}, \mathcal{C}, (o_0, a_0, \dots, o_t))$.

\subsection{Experiments}
\label{sec:recovery_evaluation_experiments}

\subsubsection{Setup}
We now describe the experimental setup, including datasets, models, and inference configurations.
\paragraph{Dataset} 
We evaluate recovery on two computer-use benchmarks, OSWorld~\citep{OSWorld} and WindowsAgentArena~\citep{bonatti2024windows}, each deployed in an isolated environment and providing a user query together with an oracle evaluator that determines whether its execution succeeds. 
\paragraph{Models} 
We evaluate a range of state-of-the-art computer-use agents. As closed-source models, we use Claude Sonnet 4.6~\citep{claude_sonnet} and GPT-5.4-Computer-Use~\citep{gpt54}, which achieve top performance on OSWorld. For open-source models, we consider two post-trained agentic models, EvoCUA-32B~\citep{xue2026evocua} and UI-TARS-1.5-7B~\citep{qin2025ui}, together with Qwen3.5-9B-Thinking~\citep{qwen3.5}, a strong open-source model without task-specific post-training.
\paragraph{Inference Configurations} 
We run experiments under \textbf{short horizon} and \textbf{long horizon} settings, with recovery distance $\mathrm{d}$ set to 1 and 5, respectively. We further vary the maximum recovery step budget $\mathrm{r} \in \{1, 3, 10\}$.
\begin{figure*}[t!]
\vspace{-0.1in}
\centering

\begin{minipage}[t]{0.32\textwidth}
\centering
\raisebox{-\height}{\includegraphics[width=\textwidth]{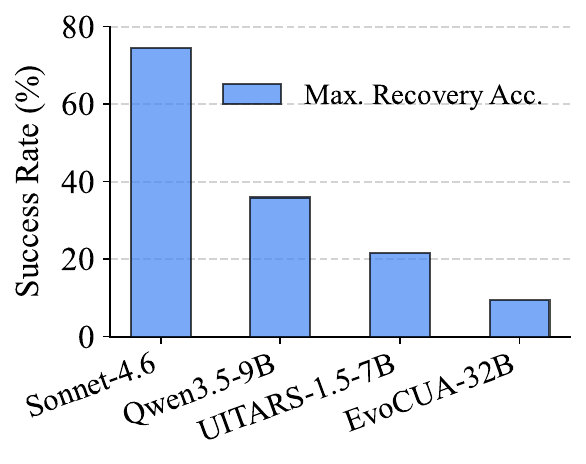}}
\vspace{-0.15in}
\captionof{figure}{Maximum long-horizon recovery success rate among four different explicit recovery strategies on OSWorld.}
\label{fig:long_horizon_recovery_success_rate}
\end{minipage}\hfill
\begin{minipage}[t]{0.32\textwidth}
\vspace{0pt}
\centering
\vspace{0.06in}
\raisebox{-\height}{\resizebox{\linewidth}{!}{%
\renewcommand{\arraystretch}{1.1}
\begin{tabular}{lcc}
\toprule
\begin{tabular}[c]{@{}l@{}}\textbf{Recovery} \\ \textbf{Context}\end{tabular} & \textbf{SR (\%)} $\uparrow$ & \textbf{$\Delta$} $\uparrow$ \\
\midrule
Baseline & $39.38 \pm 1.17$ & --- \\
\midrule
Contextualizer & $41.06 \pm 3.05$ & \textcolor{teal}{$+1.68$} \\
0$\sim$R & $\mathbf{44.26 \pm 1.90}$ & \textcolor{teal}{$\mathbf{+4.88}$} \\
0$\sim$E & $40.49 \pm 0.95$ & \textcolor{teal}{$+1.11$} \\
w/o History & $40.2 \pm 1.84$ & \textcolor{teal}{$+0.82$} \\
\bottomrule
\end{tabular}%
}}
\vspace{0.0in}
\captionof{table}{End-to-end success rate (SR) across recovery strategies, where $\Delta$ is the change over the no-recovery baseline, with three independent runs.}
\label{tab:recovery_potential}
\end{minipage}\hfill
\begin{minipage}[t]{0.31\textwidth}
\centering
\raisebox{-\height}{\includegraphics[width=0.97\textwidth]{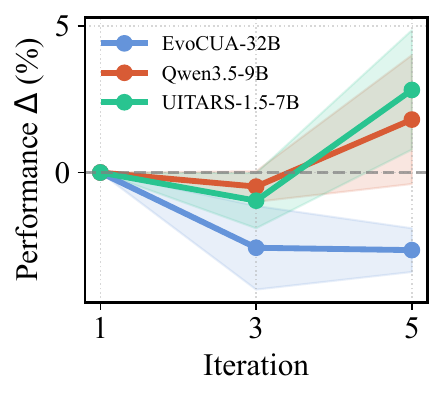}}
\vspace{-0.15in}
\captionof{figure}{Performance change ($\Delta$) across recovery iterations under \textit{Recovery Target Contextualization}.}
\label{fig:max_recovery_iteration}
\vspace{-0.15in}
\end{minipage}
\vspace{-0.15in}
\end{figure*}
\paragraph{Recovery Context} 
We consider several recovery policy configurations with different contexts $\mathcal{C}$. For the implicit recovery policy, we provide the observation–action history from the first step up to the error step, denoted \textbf{0-E}. For the explicit recovery policy, we evaluate four variants: \textbf{0-E}; \textbf{0-R}, which instead provides the observation–action history up to the target recovery step $s_{\mathrm{target}}$; \textbf{w/o History}, which excludes the history from the context; and \textbf{w/ Recovery Target Contextualization}, which contextualizes the transition history leading up to the target recovery with an LLM summarizer to reduce context~\citep{zhou2025mem1, ye2025agentfold}.

\subsubsection{Results and Analysis}
We now present and analyze the recovery evaluation results across both benchmarks and all agents.
\paragraph{Agents Fail to Recover Prior States} 
Our results in Figures~\ref{fig:osworld_shorthorizon} and \ref{fig:warena_shorthorizon} show that even the strongest computer-use agents, Claude-Sonnet-4.6 and GPT-5.4-CUA, have limited recovery capability, while the strongest open-source model, EvoCUA-32B, exhibits a near-complete inability to restore previous states, with an average recovery rate below 1\%. \textit{This suggests that current forward policies lack both the intrinsic capability and the behavioral tendency to autonomously perform restorative actions under failure \textbf{(Finding 2)}}. The analysis in Figure~\ref{fig:recovery_intention} reinforces this: rationales generated during implicit recovery show substantially lower similarity to the ground-truth recovery prompts, indicating that agents rarely even attempt recovery.
\begin{table}[t]
\centering
\footnotesize 

\setlength{\tabcolsep}{8pt} 

\renewcommand{\arraystretch}{0.9} 
\begin{tabular}{llc}
\toprule
\textbf{Model} & \textbf{Setting} & \textbf{SR (\%)} $\uparrow$ \\
\midrule
\multirow{2}{*}{Sonnet-4.6}
    & Baseline      & $70.27 \pm 0.70$ \\
    & w/ Recovery   & $71.81 \pm 1.51$ \\
\midrule
\multirow{2}{*}{Qwen3.5-9B}
    & Baseline      & $38.76 \pm 0.64$ \\
    & w/ Recovery   & $41.56 \pm 4.14$ \\
\midrule
\multirow{2}{*}{UI-TARS-1.5-7B}
    & Baseline      & $17.25 \pm 1.55$ \\
    & w/ Recovery   & $15.74 \pm 0.65$ \\
\midrule
\multirow{2}{*}{EvoCUA-32B}
    & Baseline      & $54.88 \pm 4.39$ \\
    & w/ Recovery   & $51.55 \pm 2.54$ \\
\bottomrule
\end{tabular}
\vspace{-0.05in}
\caption{End-to-end task success rate (SR) on OSWorld across different models with the \textit{Recovery Target Contextualization} strategy, over two independent runs.}
\vspace{-0.1in}
\label{tab:end_to_end}
\end{table}
\begin{table*}[t!]
\centering
\renewcommand{\tabcolsep}{3pt} 
\vspace{-0.05in}
{\small
\begin{tabular}{
>{\centering\arraybackslash}m{0.10\textwidth}
>{\centering\arraybackslash}m{0.12\textwidth}
>{\centering\arraybackslash}m{0.20\textwidth}
>{\centering\arraybackslash}m{0.20\textwidth}
>{\arraybackslash}m{0.12\textwidth}
>{\arraybackslash}m{0.12\textwidth}
}
\toprule
\footnotesize \textbf{Error Type} &
\footnotesize \textbf{Instruction} & 
\footnotesize \textbf{Target State} & 
\footnotesize \textbf{Restored State} & 
\footnotesize \hspace{0.15in}\textbf{Action} &
\footnotesize \hspace{0.1in} \textbf{Problem} \\ 
\midrule

\footnotesize \textbf{Spurious Perception} &
\scriptsize \textbf{Align} the text of the first textbox on slide 3 to the \textbf{right}, on slide 4 to the center, and on slide 5 to the left. &
\begin{minipage}{\linewidth}
    \centering
    \includegraphics[width=1\linewidth]{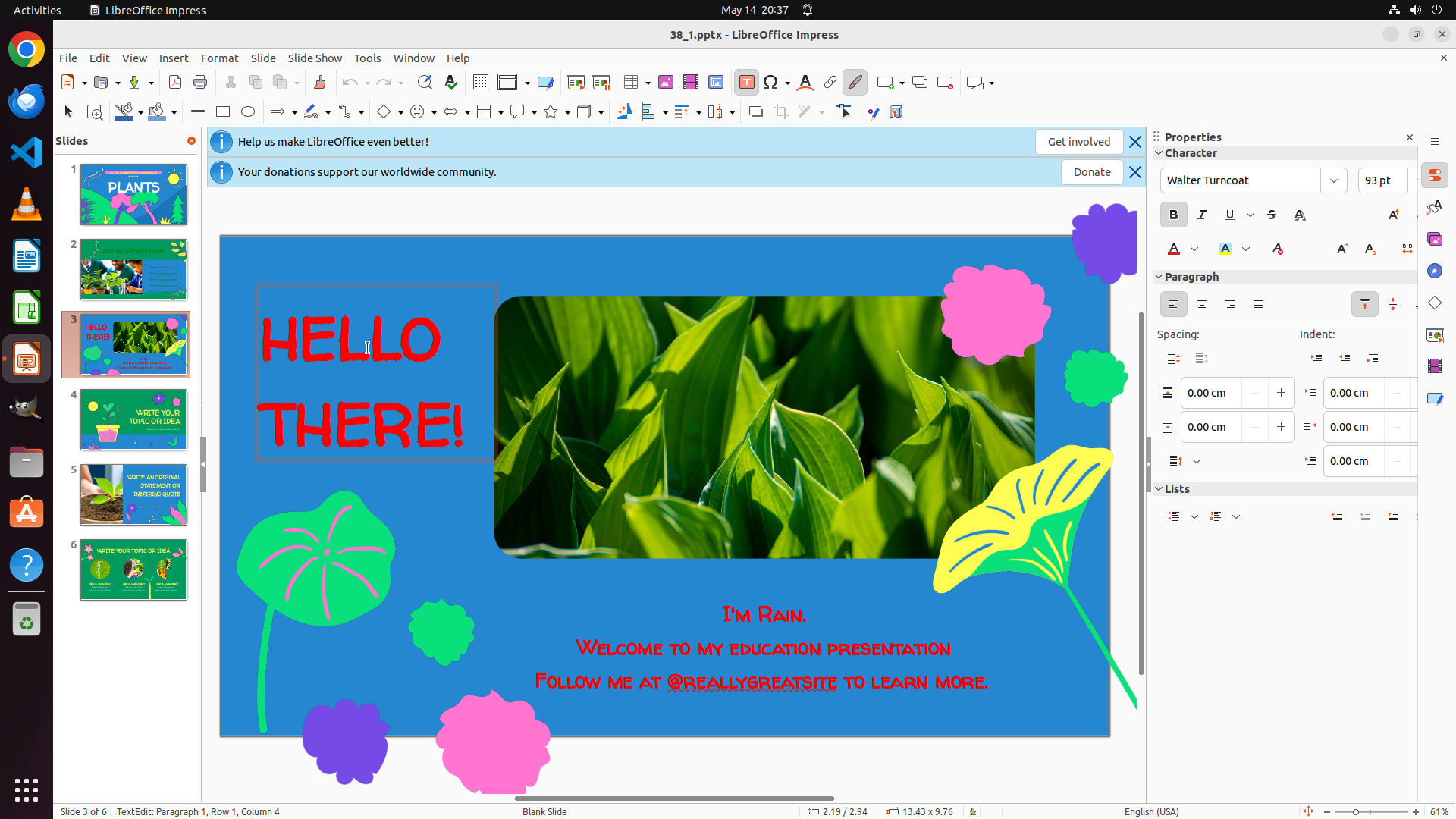}
\end{minipage} &
\begin{minipage}{\linewidth}
    \centering
    \includegraphics[width=1\linewidth]{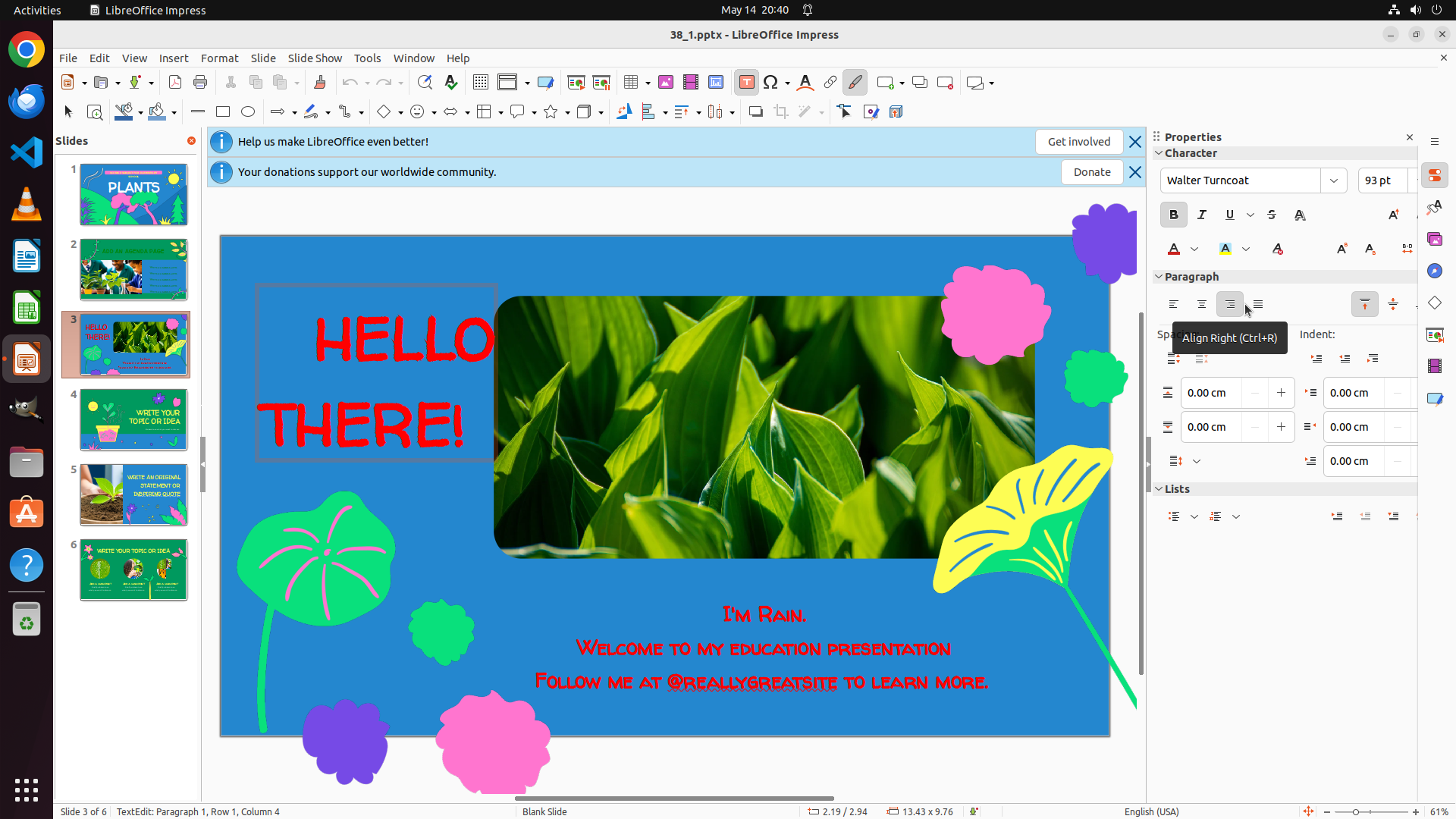}
\end{minipage} &
\scriptsize \textbf{[click]} on the \textbf{hovering alignment button} to align slide 3 the text to the right. &
\scriptsize \textbf{Fixates on to the hovering UI element in recovered state}, \textbf{skipping [drag]} on the text before clicking the alignment button.  
\tabularnewline
\midrule

\footnotesize \textbf{State Tracking Failure} &
\scriptsize Change the font to \textbf{"Times New Roman"} throughout the text. &
\begin{minipage}{\linewidth}
    \centering
    \includegraphics[width=1\linewidth]{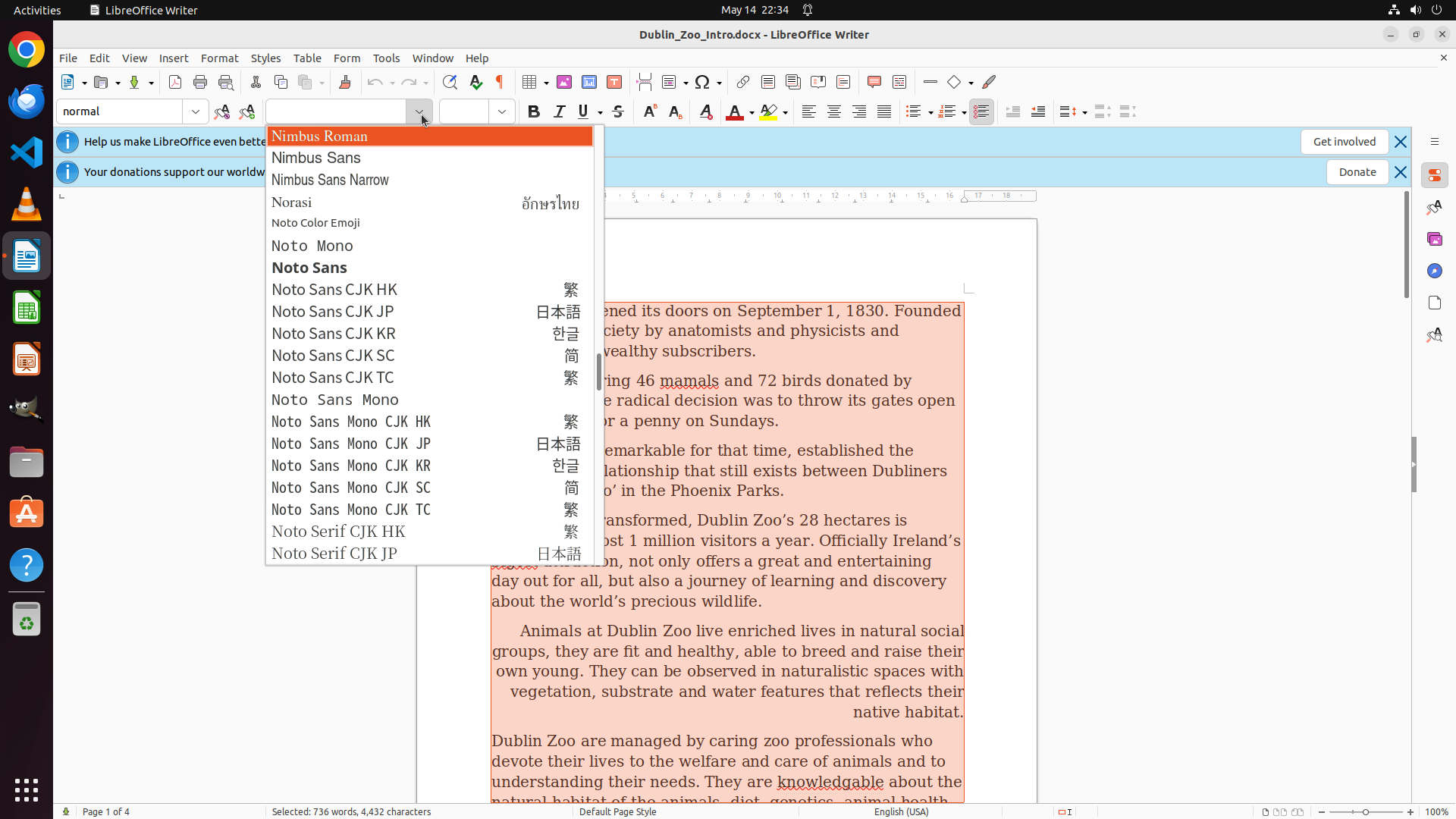}
\end{minipage} &
\begin{minipage}{\linewidth}
    \centering
    \includegraphics[width=1\linewidth]{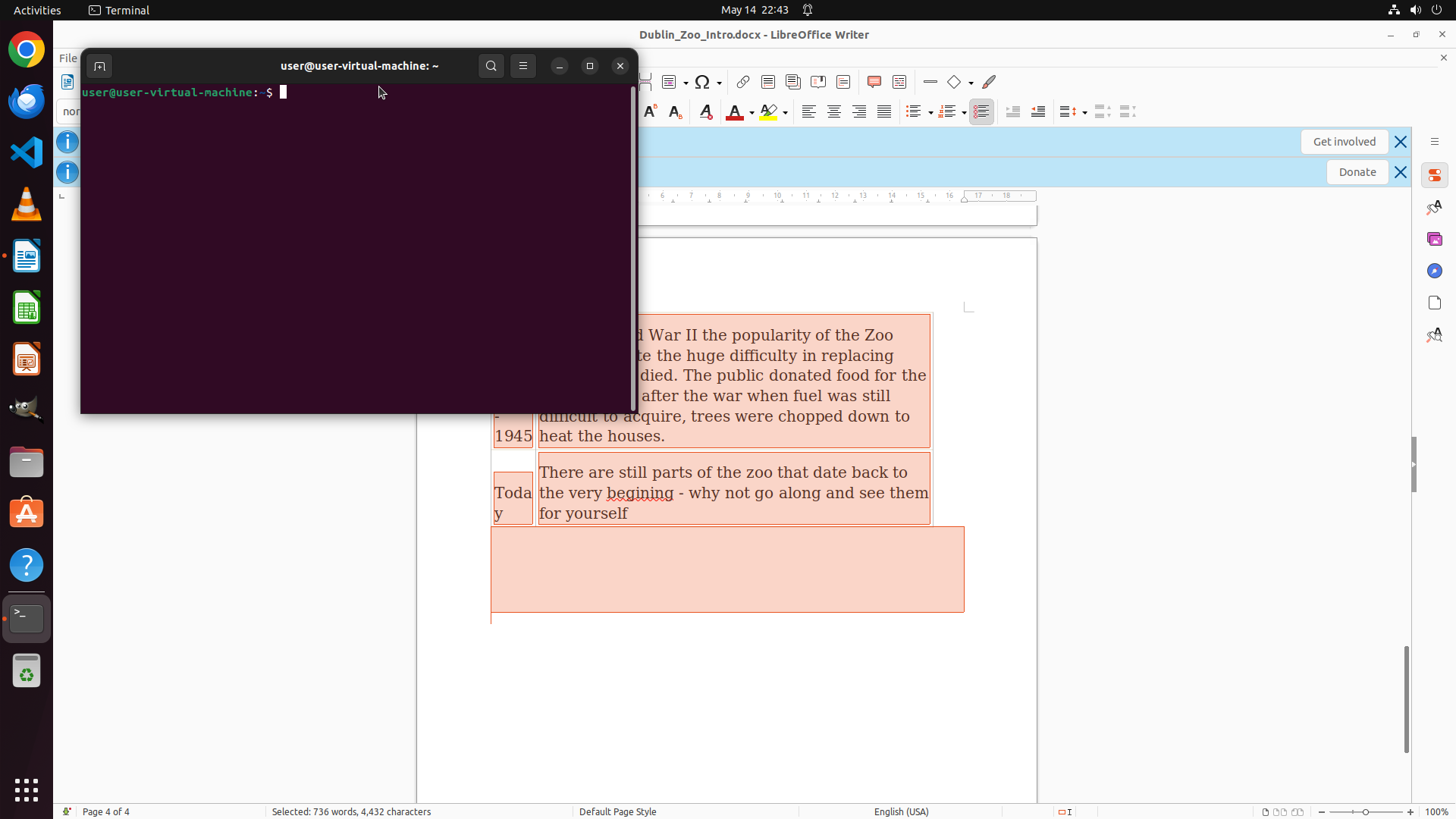}
\end{minipage} &
\scriptsize \textbf{[scroll down]} to find Times New Roman, since font selection interface is opened. &
\scriptsize \textbf{Hallucinates the restored state due to failed state synchronization} with its previous states. 
\tabularnewline
\bottomrule
\end{tabular}
}
\vspace{-0.1in}
\caption{Qualitative analysis of erroneous agent behaviors induced by imperfect state restoration.}
\vspace{-0.05in}
\label{tab:qualitative_analysis}
\end{table*}
\paragraph{Analysis on Explicit Recovery} 
The results in Figures \ref{fig:osworld_shorthorizon} and \ref{fig:warena_shorthorizon} further show that explicit model-based recovery substantially outperforms implicit recovery, by an average margin of 17.86\% across the Linux and Windows environments. In addition, the largest improvements are observed for generalist computer-use agents, such as Claude-Sonnet-4.6, GPT-5.4-CUA, and Qwen-3.5-9B. \textit{These results suggest that the recovery capability of current computer-use agents is latent rather than absent, but such capabilities are not effectively elicited during standard forward task execution \textbf{(Finding 3)}}. In contrast, the gains are considerably smaller for post-trained open-source agents, suggesting that \textit{extensive post-training on one-directional task-completion trajectories may weaken their inherent recovery capability \textbf{(Finding 4)}}. Figure~\ref{fig:long_horizon_recovery_success_rate} shows the same trend in long-horizon settings: generalist agents exhibit substantially greater recovery potential, whereas post-trained specialists consistently struggle with explicit recovery. Nevertheless, even the most capable models combined with basic explicit recovery strategies still fail to achieve complete recovery under our evaluation settings, suggesting that substantial research remains in developing more robust recovery mechanisms
for computer-use agents. 

\paragraph{Efficacy of \textsc{ReTrace}}
To assess how reliably \textsc{ReTrace} judges recovery success, we conduct a human evaluation study against three baselines: two ablations of our oracle $\mathcal{R}$ with either the image-wise matching component $\mathcal{R}_O$ or the transition-consistent component $\mathcal{R}_{\mathrm{TC}}$ removed, and GPT-5.4, a state-of-the-art model with extensive computer-use knowledge. We recruit 12 annotators to judge whether each recovered state aligns with the target state, with 50 samples per variant and each annotator rating 25 of them. As shown in Table~\ref{tab:evaluation_preference_eval}, \textsc{ReTrace} agrees with human judgments more often than all baselines, including GPT-5.4, confirming the reliability of its recovery oracle $\mathcal{R}$. We further report an average inter-annotator agreement of 82\% and a Cohen’s kappa~\citep{cohen1960coefficient} of 0.45, indicating reliable agreement.

\section{End-to-End Evaluation of Recovery}
\label{main:end2end_evaluation}

In this section, we analyze how recovery quality affects the end-to-end task performance of computer-use agents. For more details, please see Appendices~\ref{appendix:end2end_evaluation} and~\ref{appendix:Prompts}.

\subsection{Setup}
We follow the evaluation protocols of Sections~\ref{main:ideal_gains_from_recovery} and~\ref{sec:recovery_evaluation_experiments}. Specifically, the forward and recovery agents share the same backbone model, on which we run the explicit recovery strategies.

\subsection{Results and Analysis}
\label{main:end2end_quantitative_analysis}

\paragraph{Potential Gains from Model-Based Recovery}
To validate the potential benefits of model-based recovery, we analyze its performance gains across diverse recovery settings. As shown in Table~\ref{tab:recovery_potential}, model-based recovery improves Qwen3.5-9B over the no-recovery baseline under every recovery setting, with an average gain of 2.12\% in the end-to-end setting. Notably, these gains hold despite the suboptimal recovery success rates observed in Figures~\ref{fig:osworld_shorthorizon} and~\ref{fig:warena_shorthorizon}, \emph{indicating that even imperfect recovery can begin to unlock the latent capabilities of current computer-use agents, and that more reliable recovery mechanisms hold substantial promise for improving the overall performance and robustness \textbf{(Finding 5)}}.

\paragraph{Recovery Performance Effects on Task Completion}
The results in Table~\ref{tab:end_to_end} show that models with weaker recovery capabilities (UI-TARS-1.5-7B, EvoCUA-32B) suffer performance degradation during end-to-end inference, underscoring the importance of accurate recovery behavior. In addition, Figure~\ref{fig:max_recovery_iteration} demonstrates that allowing more recovery attempts yields further gains for models with comparably better recovery performance (Qwen-3.5-9B, UI-TARS-1.5-7B). EvoCUA-32B, in contrast, exhibits severely limited explicit recovery capability: while even a single successful recovery can substantially improve downstream task performance, its consistently low recovery success rate means that additional attempts more often derail the trajectory into increasingly erroneous states. This suggests that \emph{accurate model-level recovery delivers performance gains, while weak recovery can actively derail the trajectory \textbf{(Finding 6)}}.

\subsection{Qualitative Analysis}
In Section~\ref{main:end2end_quantitative_analysis}, we showed that poor recovery performance can severely degrade downstream task completion, in contrast to the idealized settings of Section~\ref{main:ideal_gains_from_recovery}. To understand why, we examine the rationales that forward agents generate after recovering to incorrect states, and categorize the resulting behaviors into two major patterns (Table~\ref{tab:qualitative_analysis}). Additional examples are provided in Appendix~\ref{appendix:additional_qualitive_analysis}.

\paragraph{Spurious Perception}
Our first observation is that agents \textit{fixate on elements distinct from the target state} and rely on them when choosing their next action. For example, in Table~\ref{tab:qualitative_analysis}, the agent immediately clicks the right-alignment button because the cursor happens to hover over it in the restored state, skipping the drag operation on the text that should have come first and thereby deviating from the trajectory needed for correct task completion.

\paragraph{State Tracking Failure}
Meanwhile, agents also hallucinate actions by assuming the restored state matches the target, failing to register the discrepancy when the two diverge. For example, as shown in Table~\ref{tab:qualitative_analysis}, although the restored state is masked by the terminal interface and no longer shows the font-selection UI, the agent acts as if that interface were still open, revealing the side effects of imperfect state restoration.

\section{Conclusion} 
In this paper, we conducted the first systematic study of autonomous, model-level recovery behavior in computer-use agents through \textsc{ReTrace}, a scalable pipeline that formalizes successful restoration as \emph{Transition-Consistent Recovery}, where a recovered state matches its target if identical downstream actions yield identical outcomes, and operationalizes this via \emph{goal-oriented action replay}. Empirical evaluations across closed and open-source agents reveal that current agents lack both the tendency and capability to recover autonomously, while accurate model-based recovery substantially closes this gap and translates into measurable end-to-end gains, despite characteristic failure modes such as spurious perception and state-tracking errors. We believe these findings position accurate model-based recovery as a key step toward reliable computer-use agents.
\section*{Limitations}
\textsc{ReTrace} is designed to measure and analyze recovery capabilities and behaviors rather than to directly improve them. Our explicit recovery settings can only elicit capabilities already present in the underlying model and cannot compensate for fundamental deficiencies in GUI understanding, state tracking, or action planning. Consequently, inaccurate recovery by specialist agents with weaker recovery capabilities may induce additional trajectory drift and reduce task success. Potential directions for addressing this gap include recovery-aware post-training on erroneous states, target states, and corresponding corrective action trajectories, as well as auxiliary objectives for tracking latent environment states and predicting action-induced transitions. \textsc{ReTrace} can provide the evaluation protocol and controlled recovery scenarios needed to develop and assess such methods, but designing and validating them is beyond the scope of the present work.

\section*{Ethical Considerations}
Our work aims to advance the reliability of computer-use agents by systematically studying their recovery behavior, contributing positively to the safe and dependable deployment of digital automation. However, we acknowledge the dual-use nature of such research, as more capable agents could in principle be repurposed for harmful automation. To mitigate such risks, we conduct all experiments in sandboxed virtual environments using publicly available benchmarks in accordance with their licenses, without interaction with real user accounts or personal data, and recommend pairing deployments with appropriate safeguards.

\bibliography{custom}
\appendix

\clearpage

\appendix

\section{\textsc{ReTrace} Details}\label{appendix:ReTrace}
Here, we provide implementation details on the evaluation set construction and recovery evaluation in \textsc{ReTrace} framework. We provide statistics of the filtered trajectories in Table \ref{appendix:dataset_construction_statistics}, and distribution of domains within our recovery evaluation set in Table \ref{appendix:recovery_evaluation_domain_composition}.

\subsection{Evaluation Set Construction}
\label{appendix:evaluation_set_construction}
\paragraph{Benchmarks} We primarily leverage two publicly available benchmarks consisting of both web tasks and non-web tasks: OSWorld \footnote{\url{https://github.com/xlang-ai/OSWorld.git}}~\citep{OSWorld}, released under the Apache License 2.0, and WindowsAgentArena~\citep{bonatti2024windows}, released under the MIT License. Both benchmarks consist of non-web tasks with significantly higher ratio compared to web-based tasks, which we demonstrate such ratio in Table \ref{tab:web_nonweb_ratio} that underscores the importance of model-level recovery, where web-based apis such as Playwright is not supported. To accelerate evaluation, we execute multiple environments in parallel, and for WindowsAgentArena, we utilize the processed environments provided by OS-Symphony\footnote{\url{https://github.com/OS-Copilot/OS-Symphony.git}}~\citep{yang2026ossymphony}, released under the Apache License 2.0, to enable efficient large-scale evaluation set construction. Our use of these artifacts is consistent with their intended use as public benchmark and evaluation resources. We deploy all environments within isolated Docker containers and apply a 10-second post-action settle window to ensure that pending UI updates are fully rendered before capturing the next observation. For screenshot resolution, we use $(1920, 1080)$ for OSWorld and $(1280, 800)$ for WindowsAgentArena. Finally, we restrict our evaluation to tasks that do not require proxy or Google Drive settings, due to constraints on proxy server budget and account configuration, respectively.

\paragraph{Models} To construct the recovery evaluation set, we use five models: Claude-Sonnet-4.6~\citep{claude_sonnet}, GPT-5.4-CUA~\citep{gpt54}, Qwen-3.5-9B-Thinking~\citep{qwen3.5}, EvoCUA-32B~\citep{xue2026evocua}, and UI-TARS-1.5-7B~\citep{qin2025ui}. Claude-Sonnet-4.6 and GPT-5.4-CUA are configured with thinking mode and xhigh reasoning mode, respectively. We use the \textit{pyautogui} action space for all models except Claude-Sonnet-4.6, which uses the proprietary \textit{claude-computer use} action space, while providing raw screenshots as observations. We adopt the default prompts from OSWorld and adapt them to Windows environments for WindowsAgentArena evaluation (Figures 
\ref{app:fig:sonnet_system_prompt_osworld}, 
\ref{app:fig:sonnet_system_prompt_windows},
\ref{app:fig:system_prompt_osworld_evocua},
\ref{app:fig:system_prompt_windows_evocua},
\ref{app:fig:tool_definition_evocua},
\ref{app:fig:description_prompt_evocua},
\ref{app:fig:action_description_prompt_evocua},  
\ref{app:fig:uitars_action_space}, 
\ref{app:fig:system_prompt_osworld_gpt54}, 
\ref{app:fig:system_prompt_osworld_qwen},
\ref{app:fig:system_prompt_windows_gpt54}, 
\ref{app:fig:system_prompt_windows_qwen},
\ref{app:fig:action_description_prompt_qwen}, \ref{app:fig:uitars_user_prompt}, \ref{app:fig:tool_definition_qwen}). All open-source models are served locally using vLLM~\citep{kwon2023efficient} with bfloat16 precision through OpenAI-compatible HTTP endpoints. Experiments are conducted using a combination of one NVIDIA H200 GPU (141GB) and two RTX A6000 GPUs.
\begin{table}[t!]
\centering
\setlength{\tabcolsep}{2.8pt}
\vspace{0.05in}
\renewcommand{\arraystretch}{1.3}
\resizebox{0.48\textwidth}{!}{%
\begin{tabular}{lcc}
\toprule
\textbf{Benchmark} & \textbf{Web Tasks (\%)} & \textbf{Non-Web Tasks (\%)} \\
\midrule
OSWorld & 12.66 & 87.34 \\
WindowsAgentArena & 21.28 & 78.72 \\
\bottomrule
\end{tabular}
}
\caption{Ratio of web and non-web tasks in the OSWorld and WindowsAgentArena benchmarks.}
\label{tab:web_nonweb_ratio}
\end{table}
\paragraph{On-Policy Rollout} For each model, we construct trajectories for the corresponding tasks in each benchmark. We then create an annotated dataset by selecting potential starting steps within each trajectory, while filtering out trajectories containing multiple actions within a single step or erroneous trajectories in which actions are missing from the agent output, as such cases can complicate the annotation process. Subsequently, the action sequences from each trajectory are replayed consecutively, and we retain only the trajectories that achieve a successful oracle evaluation in three consecutive replays (N=3) to minimize the effect from environment stochasticity (Transition-Consistent Filtering). This process empirically ensures deterministic transition dynamics, resulting in a filtered evaluation set in which the starting step of each trajectory is guaranteed to exhibit consistent transitions while retaining 64.4\% of recovery trajectories (Table \ref{appendix:dataset_construction_statistics}). For OsWorld short-horizon evaluation settings using GPT-5.4-CUA, we randomly sample 600 samples for perturbation after transition-consistent filtering to keep the API computational cost manageable prior to subsequent filtering stage (Trajectory Perturbation).

\begin{table*}[t!]
\centering

\small
\setlength{\tabcolsep}{4pt}
\vspace{-0.03in}
\renewcommand{\arraystretch}{1.3}

\resizebox{0.99\textwidth}{!}{%

\begin{tabular}{ccccccc}
\toprule
& Dataset Filtering Step & Claude-Sonnet-4.6 & GPT-5.4-CUA & QWEN-3.5-9B & UI-Tars-1.5-7B & EvoCUA-32B \\
\midrule
\multirow{4}{*}{%
\rotatebox[origin=c]{90}{%
\fontsize{8pt}{8pt}\selectfont
\textbf{\shortstack{OsWorld}}
}}
& \# of traj. after formatting & 2733 & 2749 & 790 & 1277 & 1259 \\
& \# of traj. after transition-consistent filtering & 1921 & 1912 (600) & 574 & 227 & 924 \\
& \# of traj. after perturbation & 1335 & 334 & 397 & 133 & 314 \\
& \# of traj. after final set construction & 120 & 120 & 120 & 60 & 120 \\
\midrule
\multirow{4}{*}{\rotatebox[origin=c]{90}{\fontsize{8pt}{8pt}\selectfont\ \textbf{\shortstack{Windows\\AgentArena}}}}
& \# of traj. after formatting & 919 & 867 & 211 & N/A & 280 \\
& \# of traj. after transition-consistent filtering & 779 & 407 & 153 & N/A & 201 \\
& \# of traj. after perturbation & 506 & 200 & 88 & N/A & 122 \\
& \# of traj. after final set construction & 70 & 70 & 50 & N/A & 50 \\
\midrule
\bottomrule
\end{tabular}
}
\caption{Dataset statistics after each filtering step for the short-horizon settings.}
\label{appendix:dataset_construction_statistics}
\end{table*}
\paragraph{Trajectory Perturbation} Using the filtered evaluation set consisting of annotated starting steps from each trajectory, we replay the action sequences until reaching the designated starting step, while verifying that the resulting observation matches the original state. Subsequently, a perturbation agent sharing the same backbone as the forward agent is prompted (Figures \ref{app:fig:perturbation_prefix_message}, \ref{app:fig:openai_cua_perturbation_prefix_prompt}) to intentionally drift the trajectory under a maximum step budget identical to the recovery distance $\mathrm{d}$. For short-horizon recovery evaluation, we use a perturbation budget of 1 for all models, except GPT-5.4-CUA, for which we use a budget of 2 due to its strong tendency to generate a ``take a screenshot'' action as the initial step which maps to \texttt{time.sleep()}, even under explicit instructions. We then retain only trajectories that result in oracle failure according to the benchmark evaluators (while excluding samples that were crashed due to environmental failures or gpt verification failures). This process constructs a recovery-annotated dataset, where the original starting state serves as the target recovery state, and the final perturbed state serves as the error state.

\paragraph{Final Evaluation Set Construction} Finally, using the recovery-annotated dataset, we exclude problematic samples whose perturbation trajectories generate multiple actions within a single step or fail to produce actions at arbitrary steps, as such cases complicate annotation. For short-horizon recovery evaluation, we additionally remove samples without meaningful perturbation actions, such as \texttt{time.sleep()} or \texttt{WAIT}. For GPT-5.4-CUA, we exclusively retain samples where the first action is ``take a screenshot'' and the second action corresponds to an actual perturbation action, due to its strong tendency to begin trajectories with screenshot actions. Furthermore, we remove samples where the error state and target recovery state show significantly high similarity, while also assuring that duplicate samples with both same error state observation and target recovery state observation do not exist, adopting SSIM as core mechanism. After filtering, we randomly sample evaluation subsets. For short-horizon evaluation in OSWorld, we sample 120 instances for all models except UI-TARS-1.5-7B, for which we sample 60 due to the limited number of remaining trajectories. For long-horizon evaluation, we sample 80 instances for Qwen-3.5-9B and EvoCUA-32B, 60 for Claude-Sonnet-4.6, and UI-TARS-1.5-7B. Finally for WindowsAgentArena, we sample 70 instances for Claude-Sonnet-4.6 and GPT-5.4-CUA, and 50 for EvoCUA-32B and Qwen-3.5-9B, due to the different number of remaining samples after the filtering process. 

\begin{table*}[t!]
\centering

\small
\setlength{\tabcolsep}{4pt}
\vspace{-0.03in}
\renewcommand{\arraystretch}{0.8}

\resizebox{0.99\textwidth}{!}{%

\begin{tabular}{cccccccc}
\toprule
& Domain Composition (\%) & Original &Claude-Sonnet-4.6 & GPT-5.4-CUA & QWEN-3.5-9B & UI-Tars-1.5-7B & EvoCUA-32B \\
\midrule
\multirow{10}{*}{%
\rotatebox[origin=c]{90}{%
\fontsize{10pt}{10pt}\selectfont
\textbf{\shortstack{OsWorld}}
}}
& chrome & 6.64 & 4.17 & 2.50 & 6.67 & 25.00 & 0.00 \\
& gimp & 8.23 & 0.83 & 6.67 & 8.33 & 1.67 & 6.67 \\
& libreoffice\_calc & 14.87 & 26.67 & 14.17 & 7.50 & 0.00 & 17.50 \\
& libreoffice\_impress & 14.87 & 29.17 & 15.83 & 21.67 & 13.33 & 38.33 \\
& libreoffice\_writer & 7.28 & 5.83 & 8.33 & 5.83 & 13.33 & 14.17 \\
& multi\_apps & 23.73 & 16.67 & 35.00 & 27.50 & 3.33 & 9.17 \\
& os & 7.59 & 4.17 & 1.67 & 2.50 & 11.67 & 8.33 \\
& thunderbird & 4.75 & 4.17 & 1.67 & 11.67 & 11.67 & 0.00 \\
& vlc & 5.38 & 3.33 & 5.00 & 0.83 & 15.00 & 3.33 \\
& vscode & 6.65 & 5.00 & 9.17 & 7.50 & 5.00 & 2.50 \\
\midrule
\midrule
\multirow{12}{*}{%
\rotatebox[origin=c]{90}{%
\fontsize{10pt}{10pt}\selectfont
\textbf{\shortstack{WindowsAgentArena}}
}}
& file\_explorer & 13.48 & 14.29 & 17.14 & 24.00 & N/A & 48.00 \\
& microsoft\_paint & 2.13 & 4.29 & 0.00 & 0.00 & N/A & 0.00 \\
& libreoffice\_writer & 12.77 & 22.86 & 22.86 & 4.00 & N/A & 8.00 \\
& libreoffice\_calc & 17.02 & 11.43 & 18.57 & 0.00 & N/A & 6.00 \\
& settings & 3.55 & 1.43 & 0.00 & 16.00 & N/A & 14.00 \\
& vs\_code & 13.48 & 10.00 & 4.29 & 40.00 & N/A & 14.00 \\
& msedge & 9.22 & 7.14 & 7.14 & 0.00 & N/A & 0.00 \\
& vlc & 9.92 & 8.57 & 21.43 & 8.00 & N/A & 0.00 \\
& clock & 2.84 & 1.43 & 0.00 & 8.00 & N/A & 4.00 \\
& windows\_calc & 2.13 & 0.00 & 4.29 & 0.00 & N/A & 0.00 \\
& notepad & 1.42 & 5.71 & 1.43 & 0.00 & N/A & 0.00 \\
& chrome & 12.06 & 12.86 & 2.86 & 0.00 & N/A & 6.00 \\
\midrule
\bottomrule
\end{tabular}
}
\caption{Domain Composition ratio per recovery evaluation dataset (short horizon).}
\label{appendix:recovery_evaluation_domain_composition}
\end{table*}
\subsection{Recovery Evaluation}

\label{appendix:recovery_evaluation_experiments}
\paragraph{Benchmarks} We use the evaluation sets constructed in Section~\ref{appendix:evaluation_set_construction} to measure the recovery success rates of both implicit and explicit recovery policies, where the annotated evaluation sets are instantiated in both OSWorld and WindowsAgentArena, where we set the environment settings identical with Section~\ref{appendix:evaluation_set_construction}. 

\paragraph{Models} We evaluate both implicit and explicit recovery using the five models described in Section~\ref{appendix:evaluation_set_construction}. For Claude-Sonnet-4.6 and GPT-5.4-CUA, we use a temperature of 1.0 with thinking mode and xhigh reasoning mode enabled, respectively. For Qwen-3.5-9B and EvoCUA-32B, we use a temperature of 0.7, while UI-TARS-1.5-7B is evaluated with a temperature of 0.0. As the recovery target contextualizer, we use Claude-Sonnet-4.6 as the backbone state-transition summarizer with a temperature of 0.0 and a maximum token budget of 1024. All open-source backbones are served locally using vLLM within the same computing infrastructure described in Section~\ref{appendix:evaluation_set_construction}. Finally, we use GPT-5-Mini~\citep{Singh2025OpenAIGS} as the state-verification model with default OpenAI hyperparameters.

\paragraph{Recovery Context} For implicit recovery strategies, we provide the full history of observation–action pairs up to the error state, allowing the forward agent to infer the underlying state transitions, similar to real-world settings where agents interact continuously throughout the trajectory. In contrast, for explicit recovery strategies, we provide policies with various recovery contexts,  including the observation–action history up to the error state and the target recovery state, which we denote as \textbf{0-E} and \textbf{0-R}, respectively. We additionally consider settings where such observations are not provided, as well as settings where the observation–action pairs up to the goal state are summarized by an LLM contextualizer that generates a compact description of the target recovery state and the transitions leading to it. When constructing observation–action histories, we maintain aligned observation–action pairs and exclude the final observation resulting from the last action, since new observation is input as current observation. The only exception is GPT-5.4-CUA, for which the final observation is included in the history due to proprietary API technical constraints.

\paragraph{Prompts} We construct the system prompts for explicit recovery by concatenating the recovery prefix prompts (Figures~\ref{app:fig:recovery_prefix_message_1_E}, ~\ref{app:fig:recovery_prefix_message_1_G}) from Section~\ref{appendix:Prompts} with the default system prompts used for each model in their respective benchmark settings. Similarly, we construct explicit recovery instructions by combining the model-specific instructions provided in the default OSWorld benchmark with the recovery instruction prefixes from Figures~\ref{app:fig:recovery_instruction_1_E} and \ref{app:fig:recovery_instruction_1_G}. In contrast, for implicit recovery policies, we use the original default system prompts and instructions without any recovery-specific prefix prompts or instructions. Meanwhile, we use prompts in Figures \ref{app:fig:goal_contextualizer_system_prompt} and \ref{app:fig:goal_contextualizer_instruction} for recovery target contextualizer, while leveraging prompts in Figure \ref{app:fig:state_verification_prompt} for LLM -based state verification during recovery evaluation. 

\paragraph{Recovery Verification} In \textsc{ReTrace}, we evaluate recovery success along two axes: latent state recovery success and observation-wise state recovery success. For latent state recovery, we leverage our goal-oriented action replay framework, where recovery is considered unsuccessful if the benchmark oracle evaluator outputs 0 after executing the remaining actions from the ground-truth trajectory starting from the recovered state. For observation-wise recovery, we use SSIM~\citep{wang2004image} with a threshold of 0.99 to detect visual mismatches that may interfere with accurate evaluation. Furthermore, in our main recovery evaluation experiments, we exclusively sample tasks that are consistently executable across all variants without docker failures or state verification failures, thereby ensuring a fair and reliable comparison of recovery success rates.

\section{Additional Experiment Details}
Here, we explain the experimental settings and implementation details for experiments introduced in Sections \ref{main:ideal_gains_from_recovery} and \ref{main:end2end_evaluation}, along with details on Human Evaluation introduced in Table \ref{tab:evaluation_preference_eval}.

\label{appendix:Experiment_Details}
\subsection{Ideal Gains from Recovery}
\label{appendix:ideal_gains_from_recovery}
This experiment estimates how much end-to-end performance can improve when an agent is allowed to resume from an appropriate past state after a failure. It measures the potential gain from recovery under a controlled replay setting, where the target past state is reconstructed by action replay before the forward agent resumes execution.

\paragraph{Benchmark and model} We run the analysis on OSWorld~\citep{OSWorld} using EvoCUA-32B~\citep{xue2026evocua} as the forward model, following the setup from Section \ref{appendix:evaluation_set_construction}. We set the temperature to $0.7$, \texttt{max\_tokens} to $32768$, and \texttt{max\_steps} to $50$. We first collect standard forward rollouts and keep only failed instances whose \texttt{result.txt} score is $0$. For each failed instance, we identify an error step and select a corresponding recovery step using the recovery-step selection procedure described in Section~\ref{main:ideal_gain_recovery_setup}. The environment at the selected recovery step is reconstructed by replaying the saved action prefix up to that step. We measure Success@5 on 83 samples after excluding instances where action replay fails.

\paragraph{Fixed annotations} For each failed instance, we annotate an error step $\mathrm{E}$ once before running any recovery-step experiment. The annotation stage reads the saved trajectory, reconstructs the observation-action history, and calls the LLM error-step detector. When evaluating the LLM-selected recovery step, we also annotate a recovery step $\mathrm{R_{\text{LLM}}}$ once in the same preparation stage. These annotations are then materialized into a fixed run plan, so different recovery-step conditions are evaluated against the same underlying failed trajectory rather than re-detecting $\mathrm{E}$ or $\mathrm{R_{\text{LLM}}}$ at runtime.

\paragraph{Recovery Steps} We compare four starting steps. $\mathrm{R=0}$ restarts from the initial environment state. $\mathrm{R=E-1}$ resumes from the step immediately before the detected error. $\mathrm{R=E-\text{random}}$ samples a previous step with a fixed seed and reuses the same sampled step across runs for that instance. $\mathrm{R=R_{\text{LLM}}}$ uses the recovery step predicted by the recovery-step detector. For $\mathrm{R>0}$, the environment is reset and the saved action prefix up to $\mathrm{R}$ is replayed before the forward agent resumes. For $\mathrm{R=0}$, no prefix is replayed.

\paragraph{Metric}
For each instance and each condition, we draw five independent continuations from the selected recovery state, and report Success@$\mathrm{K}$ for $\mathrm{K}\in\{1,\ldots,5\}$. Each continuation begins with action-prefix replay to reconstruct the recovery state; because replay can occasionally desynchronize from the saved trajectory (e.g.,~due to UI timing), we retry replay up to three times per continuation. Success@$\mathrm{K}$  measures whether at least one of the $\mathrm{K}$ successfully reconstructed continuations from the same recovery state completes the task, and reflects how much of the agent's latent task-solving ability is exposed by recovery.

\subsection{End-to-End Evaluation}
\label{appendix:end2end_evaluation}

\paragraph{Benchmark} We evaluate on the OSWorld~\citep{OSWorld} benchmark, where we follow a consistent setting as \ref{appendix:evaluation_set_construction}. 

\paragraph{Forward agents} We use three open-source models as the forward agent: Qwen3.5-9B~\citep{qwen3.5} with thinking mode, UI-TARS-1.5-7B~\citep{qin2025ui}, and EvoCUA-32B~\citep{xue2026evocua}, following the setup in Section \ref{appendix:evaluation_set_construction}, where default system prompts and instructions in Section \ref{appendix:evaluation_set_construction} are provided. Within each end-to-end experiment, the recovery agent shares the same backbone as the forward agent; the two differ only in their prompts and step budgets. For all forward agents we set \texttt{max\_steps} to $50$. Temperatures are chosen from each model's official inference recipe except for EvoCUA-32B: its default temperature of $0.0$ produces deterministic outputs, which would cause the forward agent to repeat the same action when it resumes from a past state, so we set it to $0.7$ to preserve action diversity after recovery. The resulting values are $0.7$ for EvoCUA-32B, $0.7$ for Qwen3.5-9B-Thinking, and $1.0$ for UI-TARS-1.5-7B. The \texttt{max\_tokens} budget is $32768$ for EvoCUA-32B (to accommodate its long reasoning trace), $3000$ for Qwen3.5-9B-Thinking, and $500$ for UI-TARS-1.5-7B.

\paragraph{Detectors} Both the error-step detector and the recovery-step detector are LLM-based and share Qwen3.6-Plus~\citep{qwen36plus} as their backbone, accessed through the OpenRouter API\footnote{\url{https://openrouter.ai/}} with a temperature of 0.7 and custom system prompts (Figure \ref{app:fig:system_prompt_error_detector}, \ref{app:fig:system_prompt_recovery_detector}). The error detector is queried once per forward step. Once a recovery is triggered, the error detector is frozen for the next $5$ steps; this prevents the detector from re-interrupting the forward agent before any recovery-induced changes have a chance to take effect.

\paragraph{Recovery agent} The recovery agent is given a recovery step budget of $10$ steps per attempt and is triggered at most \texttt{max\_recovery\_iteration} times per episode, with \texttt{max\_recovery\_iteration} $\in\{1,3,5\}$, using the identical system prompt and instructions as Section \ref{appendix:recovery_evaluation_experiments}. Recovery steps are tracked separately from the forward agent's step budget. When all retries are exhausted, the forward agent continues without further recovery activation.

\paragraph{Post-recovery context} After recovery completes, the forward agent resumes from the recovery-step context where the message history is truncated to the recovery boundary, and a short transition string informing the agent that the environment has been recovered is prepended before the next forward decision.

\subsection{Human Evaluation}\label{appendix:human_evaluation}
When constructing the human evaluation sets, we sample recovery results from three representative models, including Claude-Sonnet-4.6, Qwen3.5-9B, and EvoCUA-32B, across both short- and long-horizon evaluation settings in OSWorld, as well as the short-horizon setting in WindowsAgentArena. We compare \textsc{ReTrace} against three baselines: variants of our oracle evaluator with either $\mathcal{R}_{O}$ or $\mathcal{R}_{\mathrm{TC}}$ removed, and GPT-5.4, a state-of-the-art model with extensive computer-use knowledge. We collect cases where the evaluation decisions produced by the oracle implemented in \textsc{ReTrace} differ from those of the alternative evaluators, and randomly sample 50 instances for each variant, resulting in a total of 150 evaluation samples in Table~\ref{tab:evaluation_preference_eval}.
For the image-wise matching evaluator, we provide only the recovery target screenshot and the screenshot after the final recovery step. For the Transition-Consistent evaluator, success is determined solely based on whether replaying the remaining actions from the recovered state successfully reaches the original task goal. For GPT-5.4, we provide the history of observation-action pairs leading up to the error step, the recovery trajectory generated by the recovery agent, the recovery target screenshot, the error-step screenshot, and the final recovered-state screenshot. GPT-5.4 is then prompted to determine whether the semantic meaning of the recovered state aligns with the target recovery state, following the same context and inputs provided to human annotators for a fair comparison (See Figure \ref{app:fig:human_evaluation_prompt} for prompt details).

After constructing the evaluation set, we split the 50 samples for each variant into two subsets and assigned two human annotators to each subset. In total, we recruited 12 annotators (4 annotators for each variant) who were researchers based in South Korea and highly proficient in computer-use tasks and voluntarily consented to participate in the study, with each annotator compensated approximately 20 USD for their participation, provided with detailed guideline for objective evaluation (Figure \ref{app:fig:Human_Evaluation_Guideline}). Each pair of annotators independently evaluated whether the recovery target state had been successfully restored, labeling each sample in a binary success-or-failure manner consistent with our evaluators, while optionally indicating whether their decision was made despite stochastic environment dynamics (e.g., different timestamps or pop-ups) when the recovery process was believed to be successful, allowing us to analyze the effect of such environment dynamics on oracle judgments in \textsc{ReTrace} (Table \ref{tab:effect_of_environment_dynamics}).

Using these annotations as reference labels, we compared the decisions from our oracle evaluator against those from the alternative evaluators and measured their alignment with human judgments. Specifically, we independently computed the win rate of our oracle evaluator against each baseline for every annotator pair, and report the average win rate across all four subsets per variant in Table~\ref{tab:evaluation_preference_eval}. Across annotator pairs, we additionally observed an average agreement rate of 82\% and a Cohen’s kappa coefficient of 0.45~\citep{cohen1960coefficient}.

\section{Supplementary Experiments}\label{appendix:Supplementary Experiments}
Here, we provide additional experiments to further support the results and analysis in the main paper.
\begin{figure*}[t!]

  \centering
  \includegraphics[width=\linewidth]{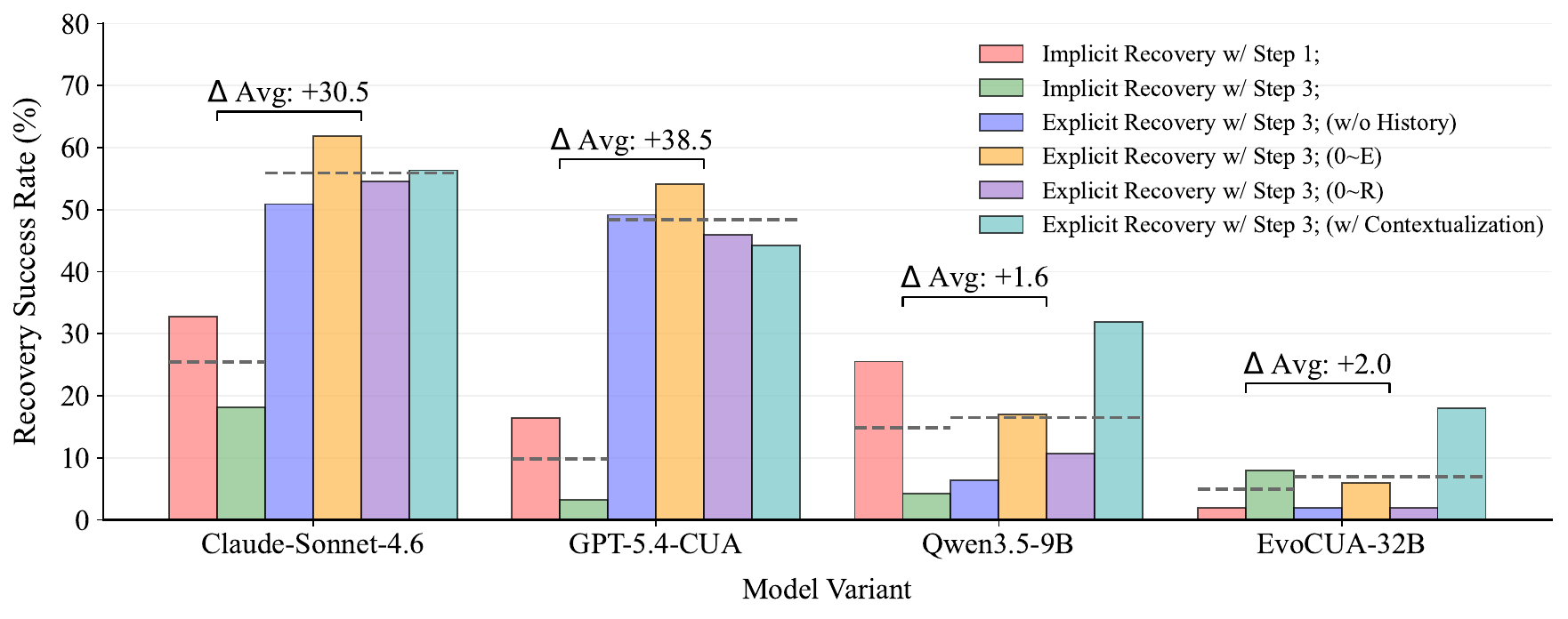}
  
  \caption{Expanded Results on WindowsAgentArena from Figure \ref{fig:warena_shorthorizon}.}

  \label{fig:windowsagentarena_expanded}

\end{figure*}
\paragraph{Additional Results on WindowsAgentArena} In Figure \ref{fig:windowsagentarena_expanded}, we present the full results on WindowsAgentArena, reporting the individual recovery success rates of each policy model. These results complement the average recovery success rates of the implicit and explicit recovery variants reported in Figure \ref{fig:warena_shorthorizon}. The results are consistent with the overall trend observed across our experiments: current computer-use agents possess latent recovery capabilities rather than lacking them entirely, but they do not reliably invoke these capabilities during inference \textit{\textbf{(Finding 2, Finding 3)}}. Moreover, the results further support our \textit{\textbf{Finding 4}}, where generalist agents exhibit stronger recovery capabilities overall. Specifically, Claude-Sonnet-4.6, GPT-5.4-CUA, and Qwen-3.5-9B achieve higher recovery success rates under implicit recovery and generally exhibit larger performance gains when explicitly instructed to perform recovery. These additional analyses further substantiate the claims and major findings presented in our original paper.
\begin{table}[t!]
\centering

\small
\setlength{\tabcolsep}{10pt}
\renewcommand{\arraystretch}{1.5}

\resizebox{0.48\textwidth}{!}{%
\begin{tabular}{lcccc}
\toprule
\textbf{Model} 
& \textbf{Max 2} 
& \textbf{Max 4} 
& \textbf{Max 5} 
& \textbf{Max 10} \\
\midrule

\textbf{Claude-Sonnet-4.6}
& 26.98 & 13.34 & 8.76 & 6.37 \\

\textbf{GPT-5.4-CUA}
& 17.85 & 6.30 & 5.81 & 5.15 \\

\textbf{Qwen3.5-9B}
& 12.04 & 7.29 & 6.12 & 5.29 \\

\textbf{EvoCUA-32B}
& 0.86 & 2.00 & 0.86 & 1.44 \\

\textbf{UI-TARS-1.5-7B}
& 7.26 & 7.81 & 12.31 & 7.82 \\

\bottomrule
\end{tabular}%
}
\caption{Implicit recovery success rate (\%) across diverse max recovery step budgets in OSWorld benchmark. We report the mean success rate across three different trials from three independently sampled evaluation sets.}
\label{tab:implicit_recovery_iteration}
\end{table}

\paragraph{Robust Suboptimal Implicit Recovery} The results in Table~\ref{tab:implicit_recovery_iteration} show that current computer-use agents struggle to perform restorative actions across different recovery budgets. Moreover, the recovery success rates of most agents consistently decrease as the maximum number of recovery steps increases, suggesting that simply allowing more steps does not resolve their recovery limitations. Notably, however, success rates do not drop to zero, suggesting that agents retain some latent ability to perform restorative actions but fail to reliably leverage it under failure scenarios. These findings further support our main-paper conclusion that current computer-use agents struggle to effectively realize their restorative capabilities when recovering from failures (\textbf{\textit{Finding 2}, \textit{Finding 3}}).

\begin{table}[t!]
\centering

\small
\setlength{\tabcolsep}{3pt}
\renewcommand{\arraystretch}{1.0}

\resizebox{0.48\textwidth}{!}{%

\begin{tabular}{ccccc}
\toprule
& Temperature & Max Tokens & Short Horizon & Long Horizon \\
\midrule
\multirow{4}{*}{%
\rotatebox[origin=c]{90}{%
\fontsize{9pt}{9pt}\selectfont
\textbf{\shortstack{EvoCUA\\-32B}}
}}
& 1.0 & 32768 & 0.00 & 7.94\\
& 1.0 & 16384  & 1.75 & 4.76\\
& 0.0 & 32768 & 3.51 & 20.63 \\
& 0.0 & 16384  & 4.39 & 6.35 \\
\midrule
\multirow{4}{*}{\rotatebox[origin=c]{90}{\fontsize{9pt}{9pt}\selectfont\ \textbf{\shortstack{UI-Tars-\\1.5-7B}}}}
& 1.0 & 1000 & 31.67 & 25.00 \\
& 1.0 & 500  & 26.67 & 26.67 \\
& 0.0 & 1000 & 26.67 & 21.67 \\
& 0.0 & 500  & 26.67 & 13.33 \\
\bottomrule
\end{tabular}
}
\caption{Recovery Success Rate (\%) across two different models with different hyper parameters using \textit{Recovery Target Contextualization} in OSWorld benchmark.}
\label{appendix:hyperparameter_ablation_study}
\end{table}
\begin{table}[t!]
\centering

\small
\setlength{\tabcolsep}{4pt}
\renewcommand{\arraystretch}{1.1}

\resizebox{0.49\textwidth}{!}{%

\begin{tabular}{ccccc}
\toprule
& Temperature & Max Tokens & Top\_P & Success Rate (\%) \\
\midrule
\multirow{5}{*}{%
\rotatebox[origin=c]{90}{%
\fontsize{10pt}{10pt}\selectfont
\textbf{\shortstack{EvoCUA\\-32B}}
}}
& 0.7 & 32768 & 0.95 & 18.00  \\
& 1.0 & 32768 & 0.95 & 14.00  \\
& 1.0 & 16384 & 0.95 & 20.00 \\
& 0.0 & 32768 & 0.95 & 20.00 \\
& 0.0 & 16384 & 0.95 & 18.00 \\
\bottomrule
\end{tabular}
}
\caption{Recovery Success Rate (\%) of EvoCUA-32B with different hyper parameters using \textit{Recovery Target Contextualization} in WindowsAgentArena benchmark.}

\label{appendix:hyperparameter_ablation_study_windowsagentarena}
\end{table}
\begin{figure*}[t!]
  \centering
  \includegraphics[width=\linewidth]{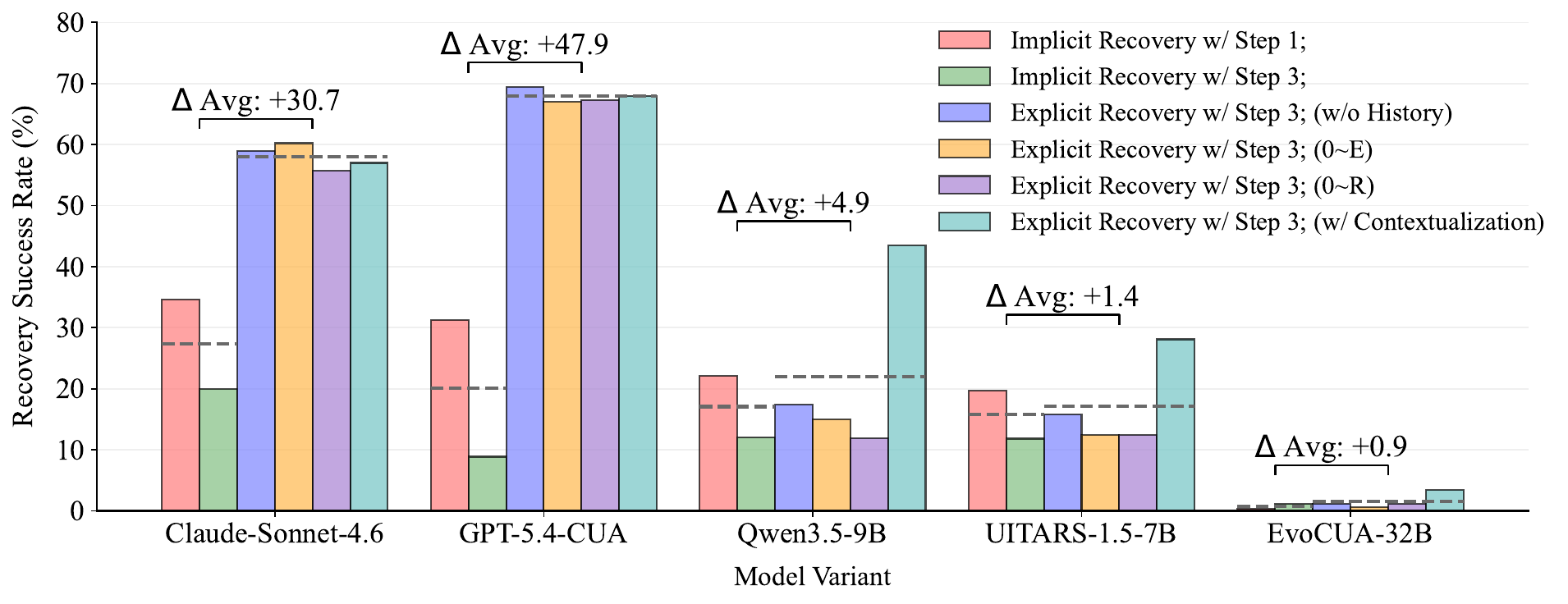}
  \caption{Three-run average recovery success rates for each recovery variant across three randomly sampled recovery evaluation sets in \textsc{ReTrace}.}
  \label{fig:osworld_shorthorizon_dataset_robust}

\end{figure*}
\paragraph{Hyperparameter Ablation Study} To further validate the robustness of our findings, we provide an additional analysis of the recovery capabilities of state-of-the-art computer-use agents under different hyperparameter settings in Table \ref{appendix:hyperparameter_ablation_study}, using the \textit{Recovery Target Contextualization} setting. The results consistently show that EvoCUA-32B exhibits limited recovery capability even under explicit recovery prompting, compared to UI-TARS-1.5-7B across a wide range of hyperparameter configurations. Furthermore, the recovery success rate also remains robust across diverse hyper parameter settings for EvoCUA in WindowsAgentArena (Table \ref{appendix:hyperparameter_ablation_study_windowsagentarena}). This finding highlights the inherent limitation of the EvoCUA-32B post-trained model, despite being extensively trained on offline trajectory data. In other words, even with explicit recovery guidance, the model remains unable to reliably restore previous states regardless of the hyper parameter settings. These observations further reinforce the results presented in Figure \ref{fig:osworld_shorthorizon} and Figure \ref{fig:long_horizon_recovery_success_rate}.

\paragraph{Dataset Robustness Analysis} In Figure \ref{fig:osworld_shorthorizon_dataset_robust}, we show that the trend observed in Figure \ref{fig:osworld_shorthorizon} remains consistent across different randomly sampled recovery evaluation sets. We randomly sample two additional recovery evaluation sets for respective policy model, each containing the same number of samples as the original set, and measure the recovery success rate of each policy model on all three sets (including one used in our main experiments). We then compute the macro-average success rate across the three evaluation sets for each recovery variant of each policy model. The results closely mirror the trend observed in Figure \ref{fig:osworld_shorthorizon}: explicit recovery instructions yield substantial gains over implicit recovery, demonstrating that current computer-use agents possess recovery capabilities but fail to reliably invoke them during inference \textit{\textbf{(Finding 2, Finding 3)}}. Furthermore, post-trained models such as EvoCUA exhibit nearly no recovery capability, while UI-TARS shows only modest improvements with explicit recovery instructions. In contrast, generalist agents (Claude, GPT-5.4, and Qwen) exhibit substantial gains from explicit recovery instructions. These results further support our claim, suggesting that extensive post-training on one-directional task-completion trajectories may suppress or weaken the inherent recovery capabilities of computer-use agents \textit{\textbf{(Finding 4)}}. Furthermore, in Table \ref{tab:implicit_recovery_iteration}, we show that across three randomly sampled recovery evaluation sets, current computer-use agents rarely have restorative behavior (\textbf{\textit{Finding 2}}), and its performance is robust regardless of evaluation set.

\begin{table}[t]
\centering
\small
\setlength{\tabcolsep}{10pt}
\renewcommand{\arraystretch}{1.5}

\begin{tabular}{lc}
\toprule
\textbf{Oracle} & \textbf{Unaffected Rate (\%)} \\
\midrule
\textsc{ReTrace (All)} & 86.0 (\%) \\
\bottomrule
\end{tabular}
\caption{Analysis on the rate of oracle evaluations unaffected by stochastic environmental dynamics.}
\label{tab:effect_of_environment_dynamics}
\end{table} 
\paragraph{Effect of Environment Dynamics on \textsc{ReTrace}} In addition, we investigate the impact of stochastic transition dynamics, such as pop-ups, rendering variations, and dynamic timestamps, on the judgments of our oracle evaluation framework. Specifically, we conduct a human study following the same protocol as in Table~\ref{tab:evaluation_preference_eval} and Appendix~\ref{appendix:human_evaluation}. Annotators are asked to determine whether a successfully recovered state differs from the target state only due to ignorable environmental variations. Since recovery failures identified by human annotators are not attributed to environmental dynamics (see the evaluation guidelines in Figure~\ref{app:fig:Human_Evaluation_Guideline}), this evaluation quantifies how often our oracle incorrectly judges a successful recovery as failure due to stochastic environmental variations. To obtain a reliable estimate, we consider only cases where both annotators agree on a judgment on both recovery success and environmental effect. From these cases, we sample instances where the human judgment disagrees with \textsc{ReTrace}, and both agree that there exists stochastic environmental dynamics. Finally, we eliminate the same recovery evaluation set distributed across the independent evaluation sets (GPT-5.4 evaluator, w/o $\mathcal{R}_{\mathrm{O}}$, w/o $\mathcal{R}_{\mathrm{TC}}$) from Table \ref{tab:evaluation_preference_eval} and compute micro-average ratio.  The results in Table~\ref{tab:effect_of_environment_dynamics} show that \textsc{ReTrace} is largely robust to stochastic environmental dynamics: 86.0\% of the \textsc{ReTrace} judgments were favorable decisions unaffected by stochastic environmental dynamics. Furthermore, since this analysis is performed on a challenging subset consisting exclusively of disagreement cases between different oracles, the overall adverse effect of stochastic environmental dynamics on the judgments of \textsc{ReTrace} is expected to be even smaller in practice.

\paragraph{Analysis on the $\mathcal{R}_{\mathrm{O}}$ and $\mathcal{R}_{\mathrm{TC}}$ disagreements} Furthermore, we analyze the quality of the judgments made by \textsc{ReTrace} in cases where the two oracles, $\mathcal{R}_{\mathrm{O}}$ and $\mathcal{R}_{\mathrm{TC}}$, disagree. Specifically, we sample disagreement cases between $\mathcal{R}_{\mathrm{O}}$ and $\mathcal{R}_{\mathrm{TC}}$ from the human evaluation sets (Tables~\ref{tab:evaluation_preference_eval} and~\ref{tab:effect_of_environment_dynamics}). Following the protocol in Table~\ref{tab:effect_of_environment_dynamics}, we consider only samples for which both human annotators within each evaluation subset agree on a recovery success evaluation. We then measure the agreement rate between human judgments and oracle decisions when using $\mathcal{R}_{\mathrm{O}}$, and $\mathcal{R}_{\mathrm{TC}}$ alone, and both oracles jointly, among such samples using the micro-average protocol demonstrated in Table \ref{tab:effect_of_environment_dynamics}. The results in Table \ref{tab:disagreement_robustness_analysis} demonstrate that the joint use of two oracles provides complementary strength and that the combination of them yields higher agreement with human judgments than either oracle alone, compared to utilizing each oracle in isolation, further indicating that even in disagreement situations, \textsc{ReTrace} can provide objective evaluation results. 

\begin{table}[t]
\centering
\small
\setlength{\tabcolsep}{10pt}
\renewcommand{\arraystretch}{1.5}

\begin{tabular}{lc}
\toprule
\textbf{Oracle} & \textbf{Alignment Rate (\%)} \\
\midrule
\textsc{$\mathcal{R}_{\mathrm{TC}}$} & 38.8 (\%) \\
\textsc{$\mathcal{R}_{\mathrm{O}}$} & 61.2 (\%) \\
\textsc{ReTrace} & 81.2 (\%) \\
\bottomrule
\end{tabular}
\caption{Analysis on the alignment rate of oracle judgments with human annotations using samples under $\mathcal{R}_{\mathrm{O}}$ and $\mathcal{R}_{\mathrm{TC}}$ disagreement.}
\label{tab:disagreement_robustness_analysis}
\end{table}
\begin{table}[t]
\centering
\small
\setlength{\tabcolsep}{10pt}
\renewcommand{\arraystretch}{1.5}

\begin{tabular}{lc}
\toprule
\textbf{Threshold} & \textbf{Alignment Rate (\%)} \\
\midrule
\textsc{0.99} & 56.9 (\%) \\
\textsc{0.95} & 33.9 (\%) \\
\textsc{0.91} & 26.6 (\%) \\
\bottomrule
\end{tabular}
\caption{Analysis of the Effect of the SSIM Threshold on the Alignment Between \textsc{$\mathcal{R}_{\mathrm{O}}$} and Human Judgments.}
\label{tab:effect_of_threshold_on_judgment}
\end{table}

\paragraph{Analysis of \textsc{ReTrace}'s Sensitivity to the SSIM Threshold} Finally, we analyze the sensitivity of $\mathcal{R}_{\mathrm{O}}$ to the SSIM threshold, which is set to 0.99 by default. Table~\ref{tab:effect_of_threshold_on_judgment} reports the alignment between the judgments of $\mathcal{R}_{\mathrm{O}}$ and human annotations under different SSIM thresholds. Following the protocol used in Tables~\ref{tab:effect_of_environment_dynamics} and \ref{tab:disagreement_robustness_analysis}, we sample a common set of instances for which both human annotators agree on whether the recovery is successful, and measure the proportion of agreement between $\mathcal{R}_{\mathrm{O}}$ and the corresponding human judgments across different thresholds among such samples. The results show that lowering the SSIM threshold causes $\mathcal{R}_{\mathrm{O}}$ to overestimate recovery success, substantially reducing its alignment with human annotations. These findings justify our design choice of adopting a strict SSIM threshold of 0.99 in \textsc{ReTrace} to ensure high-fidelity screenshot-level matching.

\onecolumn
\twocolumn
\clearpage
\section{Qualitative Analyses}
\label{appendix:additional_qualitive_analysis}
Here, we provide additional qualitative examples that explain the behavior pattern of computer agents under suboptimal recovery demonstrated in Table \ref{tab:additional_qualitative_analysis}.
\vspace{0.1in}
\begin{table}[H]
\centering
\renewcommand{\tabcolsep}{3pt} 
\small
\begin{tabular}{
>{\centering\arraybackslash}m{0.10\textwidth}
>{\centering\arraybackslash}m{0.13\textwidth}
>{\centering\arraybackslash}m{0.22\textwidth}
>{\centering\arraybackslash}m{0.22\textwidth}
>{\arraybackslash}m{0.13\textwidth}
>{\arraybackslash}m{0.13\textwidth}
}
\toprule
\footnotesize \textbf{Error Type} &
\footnotesize \textbf{Instruction} & 
\footnotesize \textbf{Target State} & 
\footnotesize \textbf{Restored State} & 
\footnotesize \hspace{0.15in}\textbf{Action} &
\footnotesize \hspace{0.1in} \textbf{Problem} \\ 
\midrule

\footnotesize \textbf{Spurious Perception} &
\scriptsize \textbf{Align} the text of the first textbox on slide 3 to the \textbf{right}, on slide 4 to the center, and on slide 5 to the left. &
\begin{minipage}{\linewidth}
    \centering
    \includegraphics[width=1\textwidth]{Figures/QualitativeAnalysis/SpuriousAttention/goal_state.png}
\end{minipage} &
\begin{minipage}{\linewidth}
    \centering
    \includegraphics[width=1\textwidth]{Figures/QualitativeAnalysis/SpuriousAttention/final_state_after_recovery.png}
\end{minipage} &
\scriptsize \textbf{[click]} on the \textbf{hovering alignment button} to align slide 3 the text to the right. &
\scriptsize \textbf{Fixates on to the hovering UI element in recovered state}, \textbf{skipping [drag]} on the text before clicking the alignment button.  
\tabularnewline
\midrule

\footnotesize \textbf{Spurious Perception} &
\scriptsize Please help me \textbf{change the color theme of VS Code} to Visual Studio Dark. &
\begin{minipage}{\linewidth}
    \centering
    \includegraphics[width=1\textwidth]{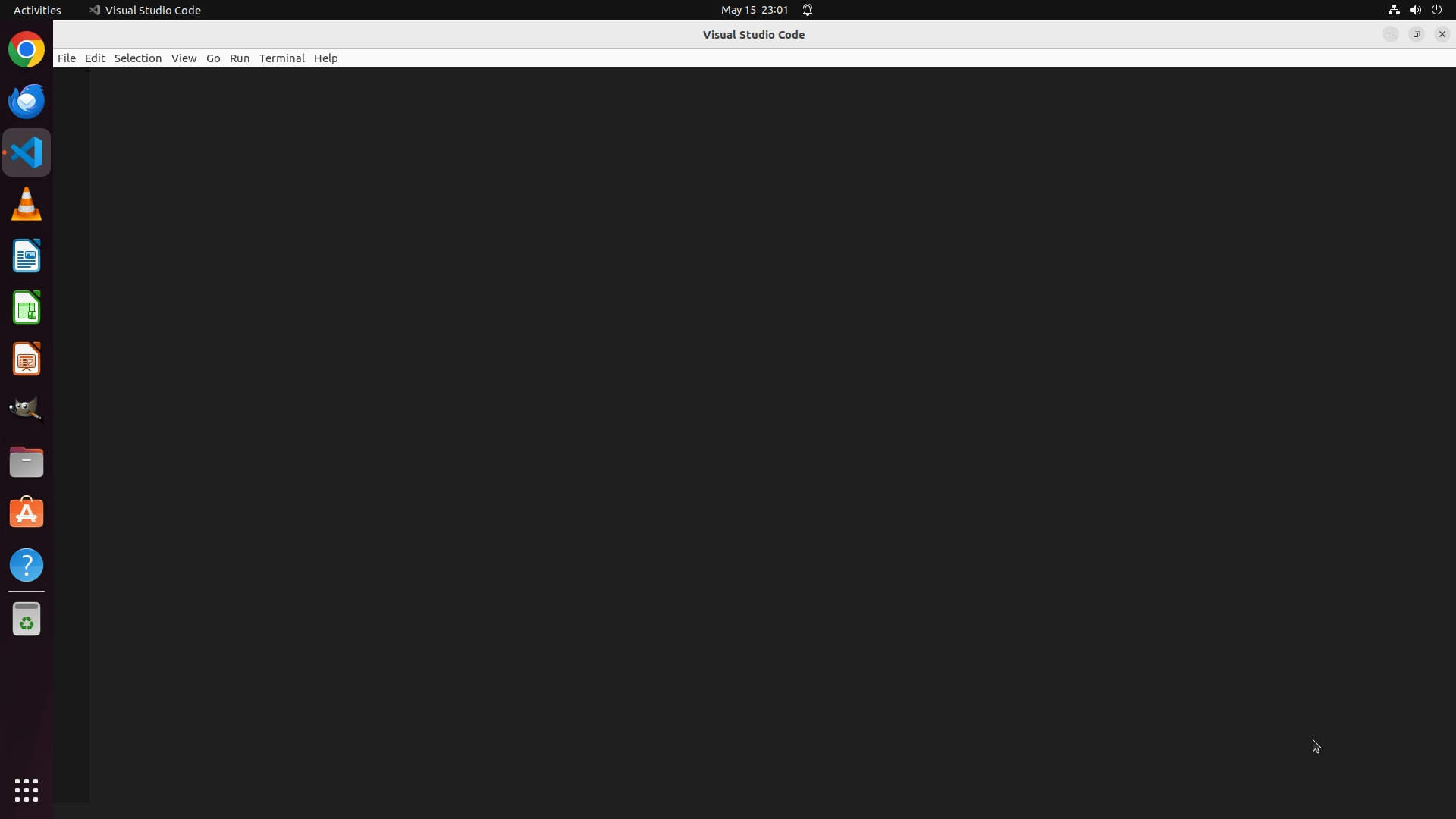}
\end{minipage} &
\begin{minipage}{\linewidth}
    \centering
    \includegraphics[width=1\textwidth]{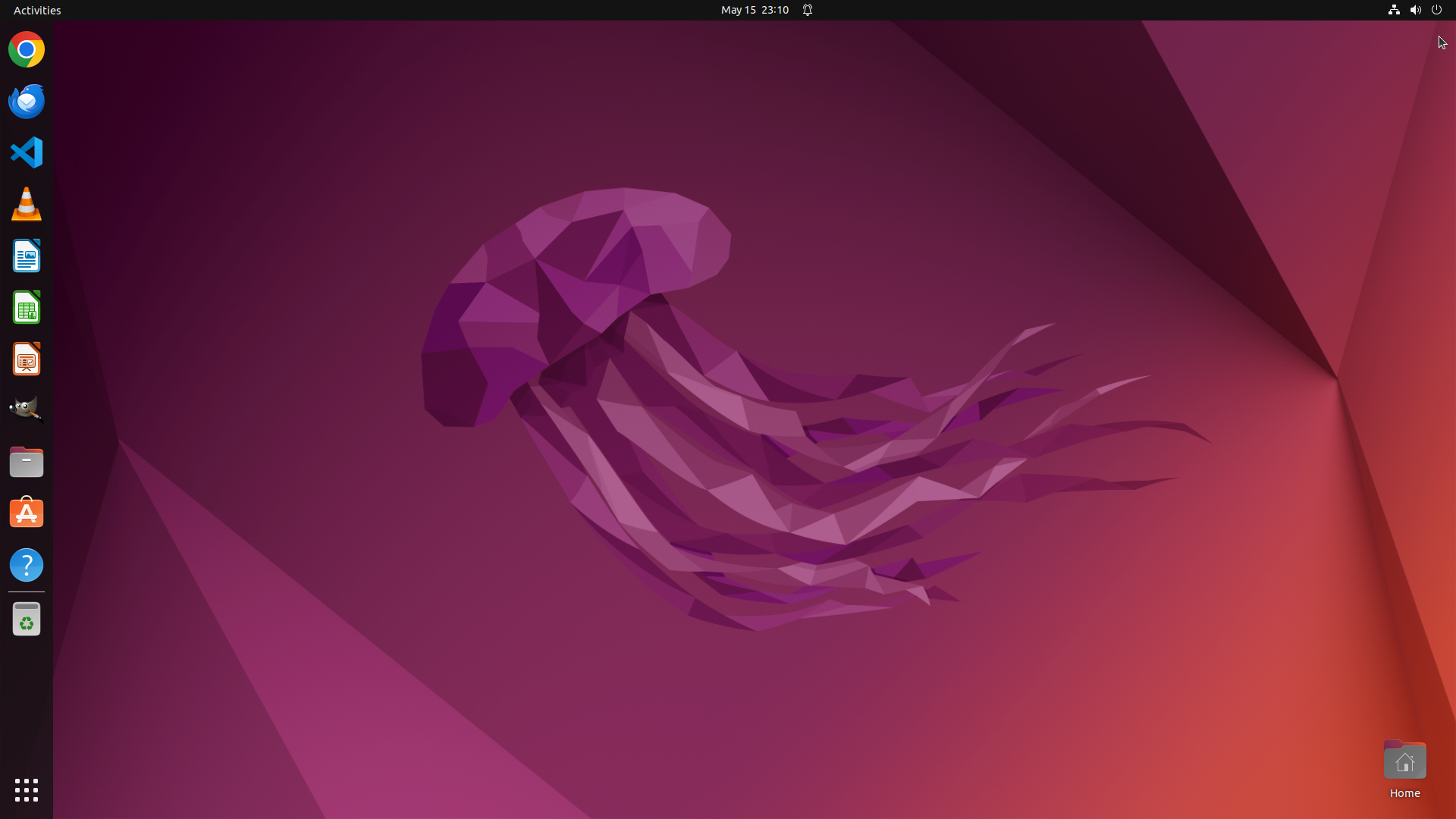}
\end{minipage} &
\scriptsize \textbf{[click]} \textbf{Activities element} in the top left corner to launch vscode application &
\scriptsize \textbf{Fixates on to the UI element} appearing in the restored state, unable to correctly click vscode application after restoration.
\tabularnewline
\midrule

\footnotesize \textbf{Spurious Perception} &
\scriptsize Could you help me \textbf{remove the dock} on the left side of the screen in the GIMP? &
\begin{minipage}{\linewidth}
    \centering
    \includegraphics[width=1\textwidth]{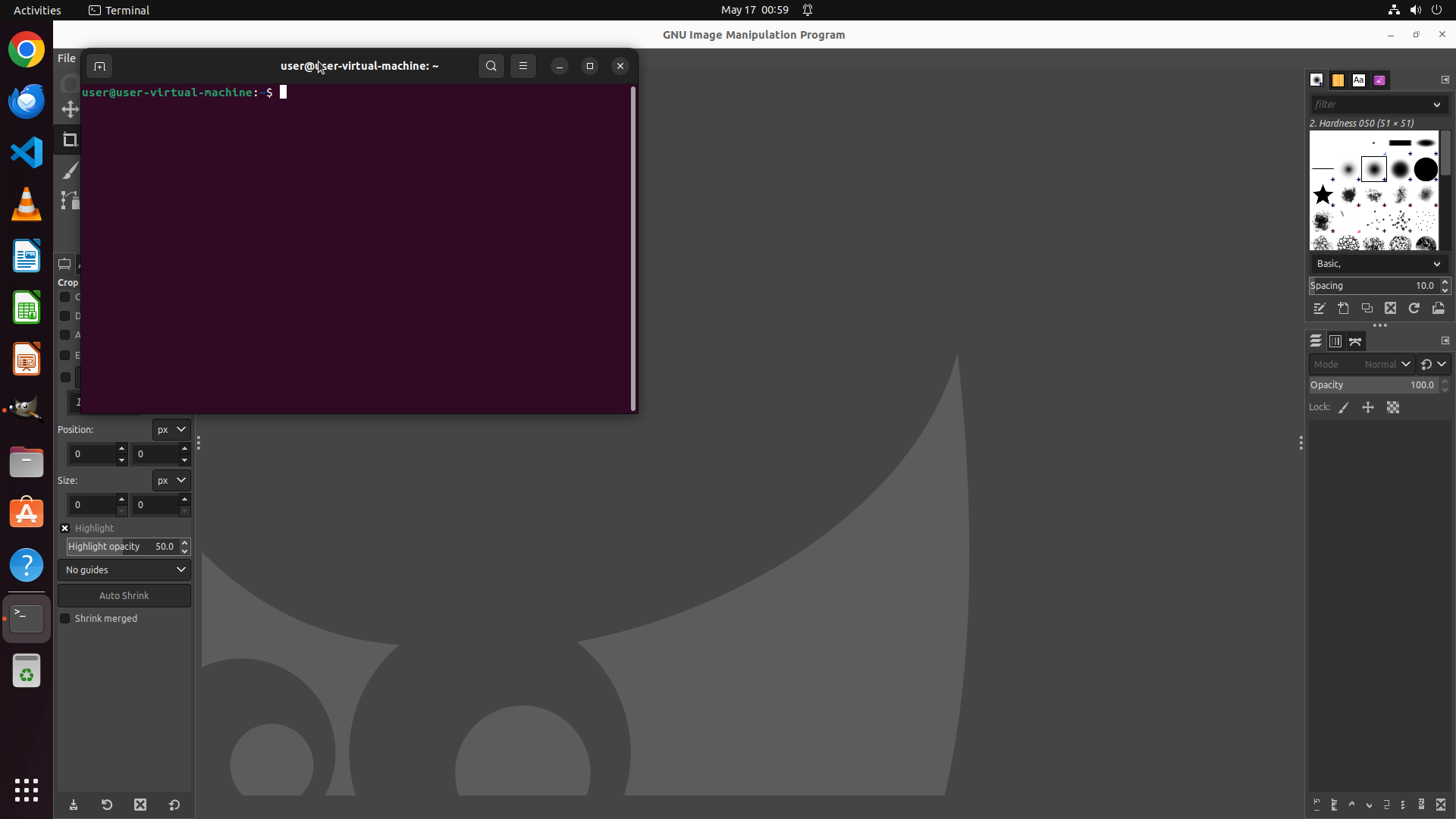}
\end{minipage} &
\begin{minipage}{\linewidth}
    \centering
    \includegraphics[width=1\textwidth]{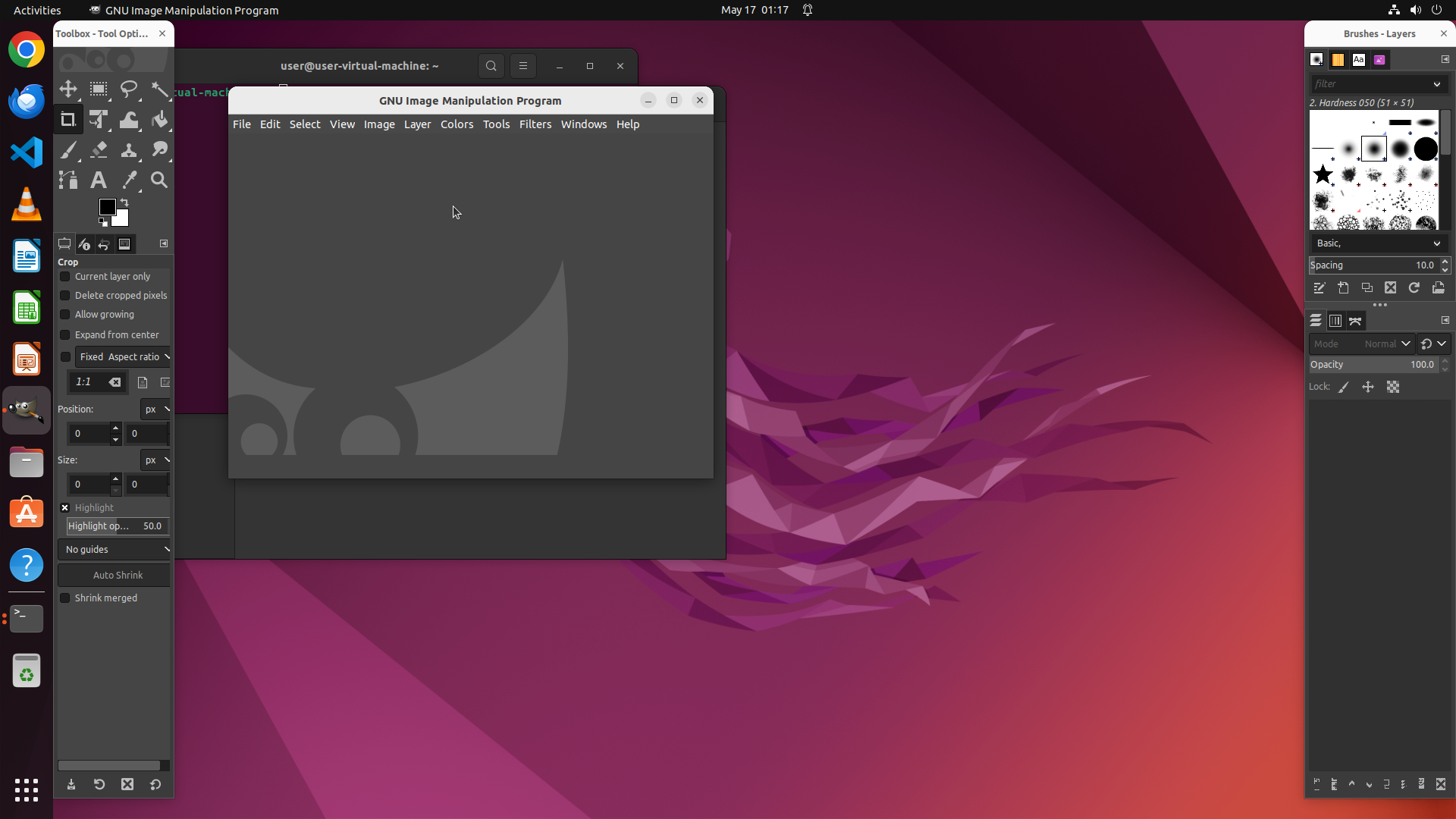}
\end{minipage} &
\scriptsize \textbf{[click]} on windows option in the right window to explore options for removing the dock &
\scriptsize \textbf{Fixates on to the new windows option irrelevant to removing the dock option}, fails to generate correct actions due to incorrect state restoration. 
\tabularnewline
\midrule

\footnotesize \textbf{State Tracking Failure} &
\scriptsize Change the font to \textbf{"Times New Roman"} throughout the text. &
\begin{minipage}{\linewidth}
    \centering
    \includegraphics[width=1\textwidth]{Figures/QualitativeAnalysis/StateTrackingFailures/goal_state.png}
\end{minipage} &
\begin{minipage}{\linewidth}
    \centering
    \includegraphics[width=1\textwidth]{Figures/QualitativeAnalysis/StateTrackingFailures/final_state_after_recovery.png}
\end{minipage} &
\scriptsize \textbf{[scroll down]} to find Times New Roman, since font selection interface is opened. &
\scriptsize \textbf{Hallucinates the restored state due to failed state synchronization} with its previous states. 
\tabularnewline

\midrule

\footnotesize \textbf{State Tracking Failure} &
\scriptsize Help me \textbf{freeze the range A1:B1} on this sheet to keep the headers always visible. &
\begin{minipage}{\linewidth}
    \centering
    \includegraphics[width=1\textwidth]{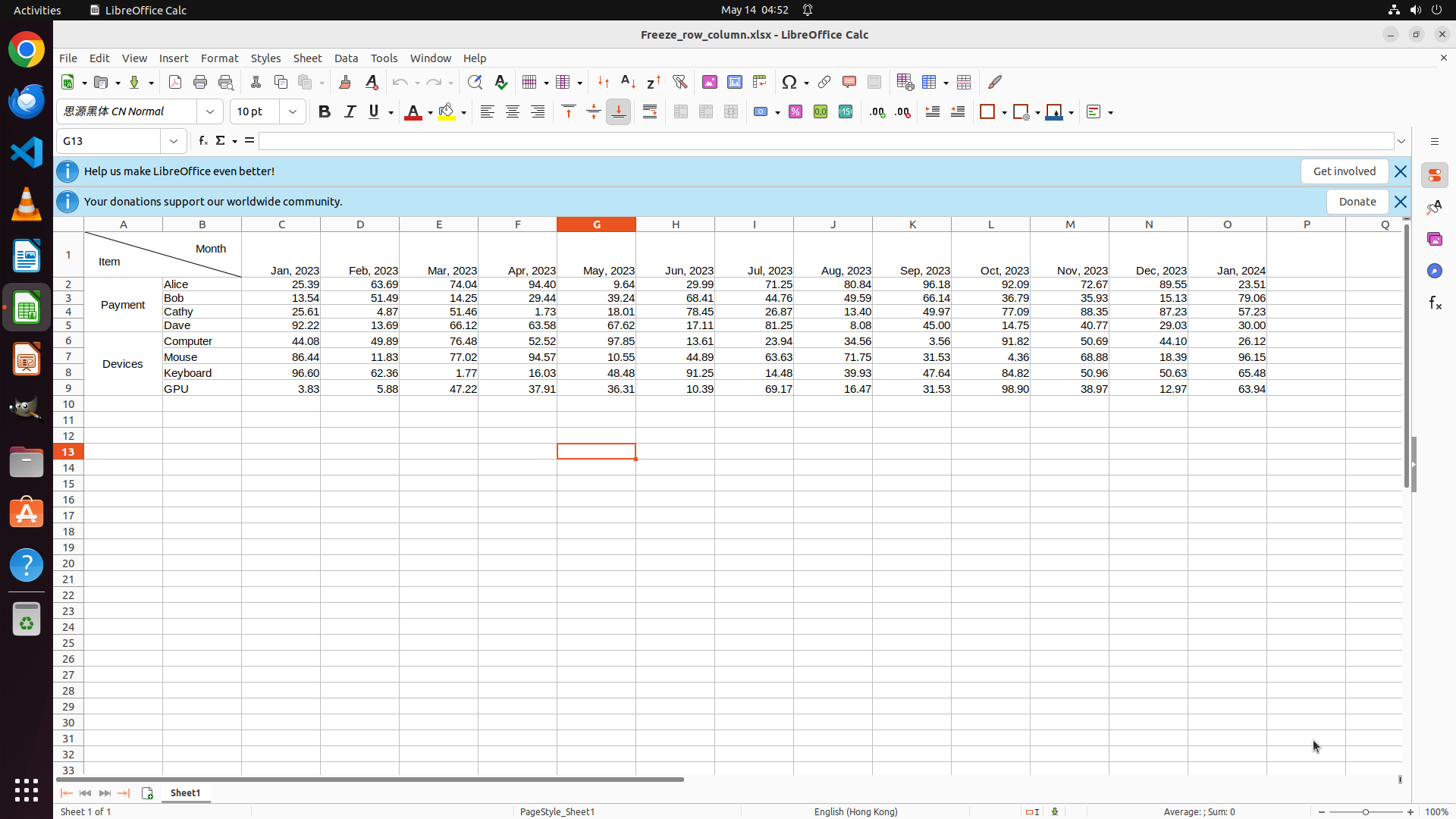}
\end{minipage} &
\begin{minipage}{\linewidth}
    \centering
    \includegraphics[width=1\textwidth]{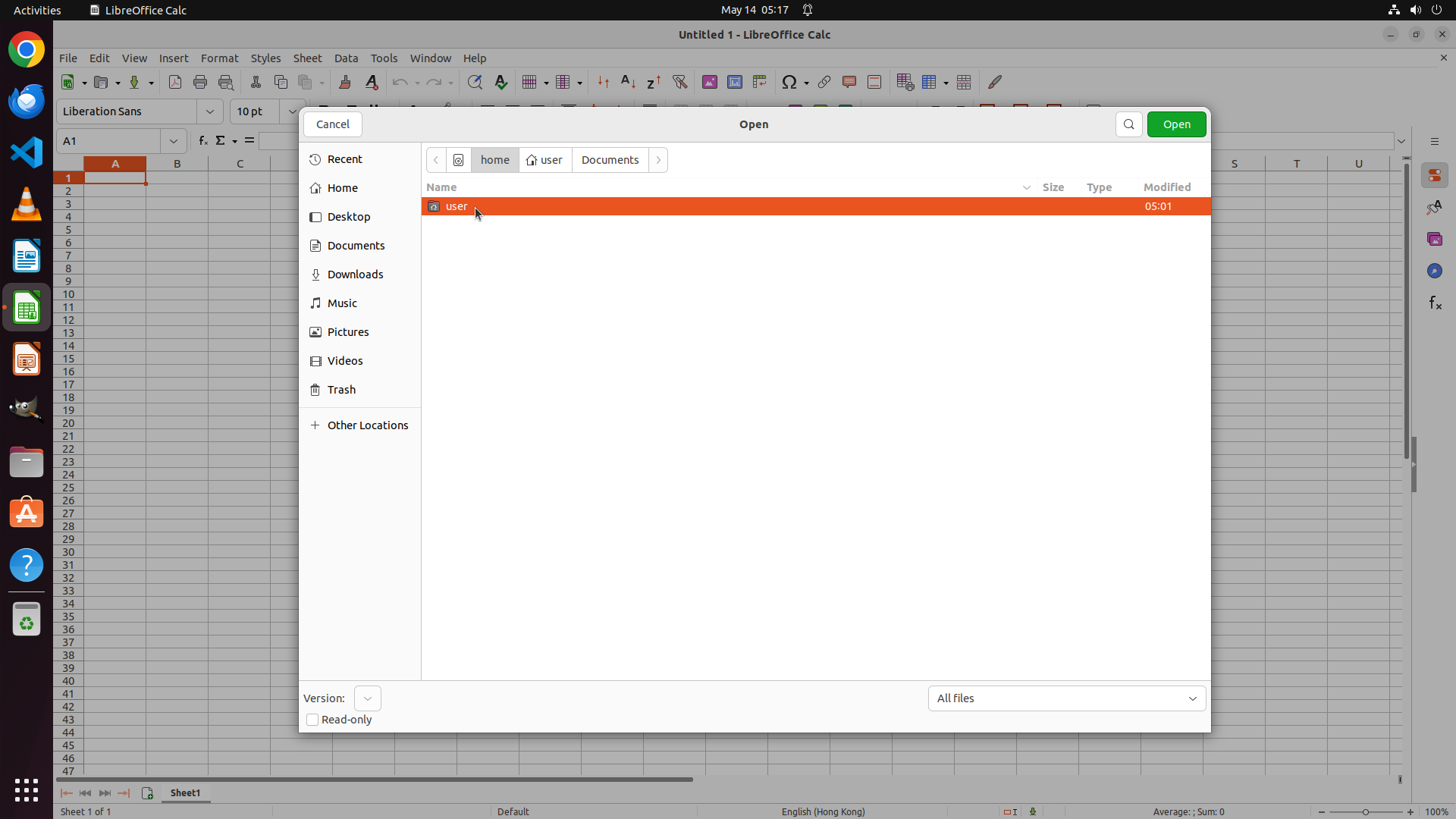}
\end{minipage} &
\scriptsize \textbf{[Drag]} from column A1 to B1, in order to accurately select the area we want to freeze &
\scriptsize \textbf{Hallucinates that the restored state allows clicking the range A1 to B1, when in fact, cannot be dragged}, failing to synchronize with the actual prior state.
\tabularnewline
\midrule
\footnotesize \textbf{State Tracking Failure} &
\scriptsize Please \textbf{modify VS Code's settings to disable error} reporting for Python missing imports. &
\begin{minipage}{\linewidth}
    \centering
    \includegraphics[width=1\textwidth]{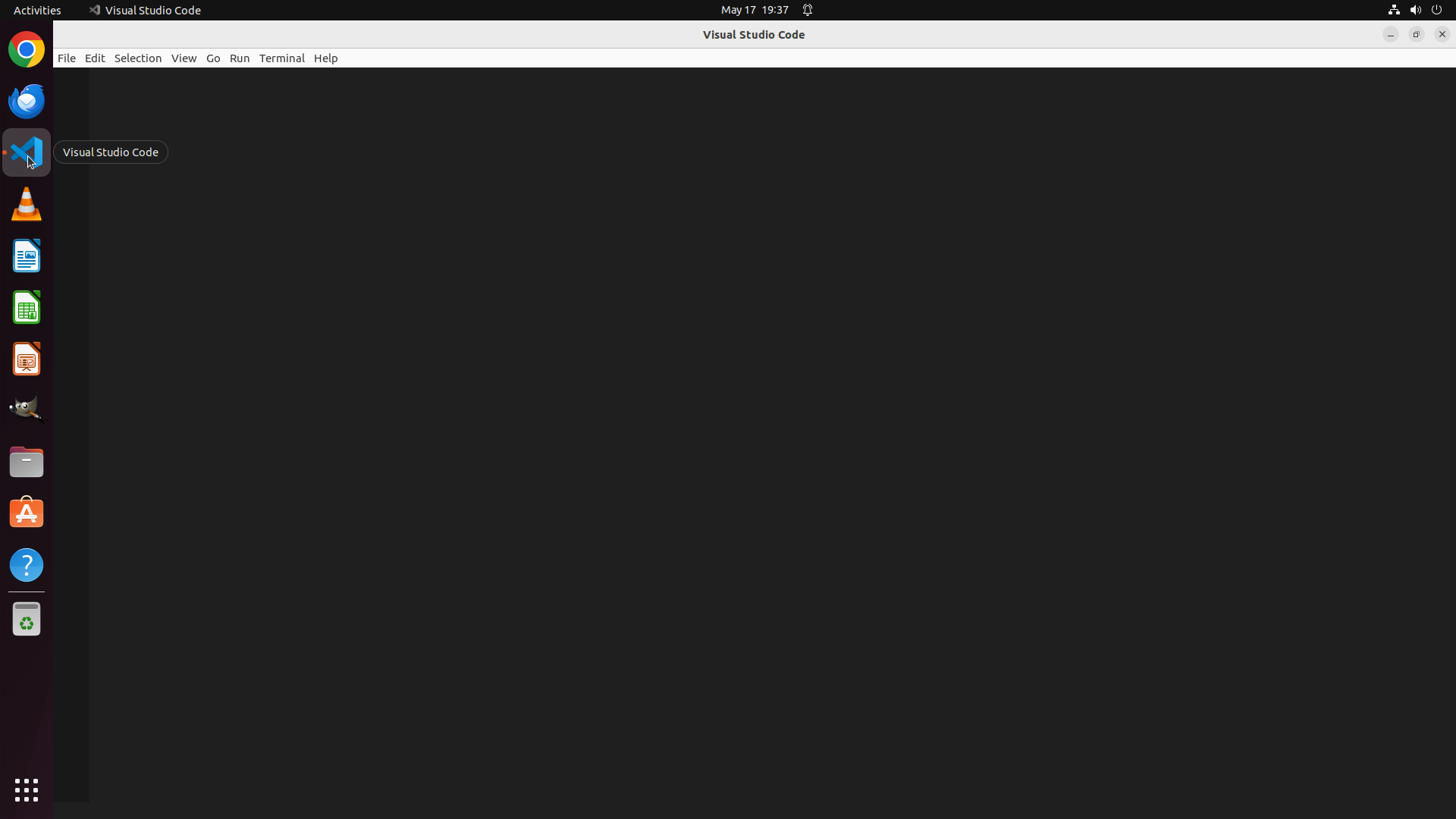}
\end{minipage} &
\begin{minipage}{\linewidth}
    \centering
    \includegraphics[width=1\textwidth]{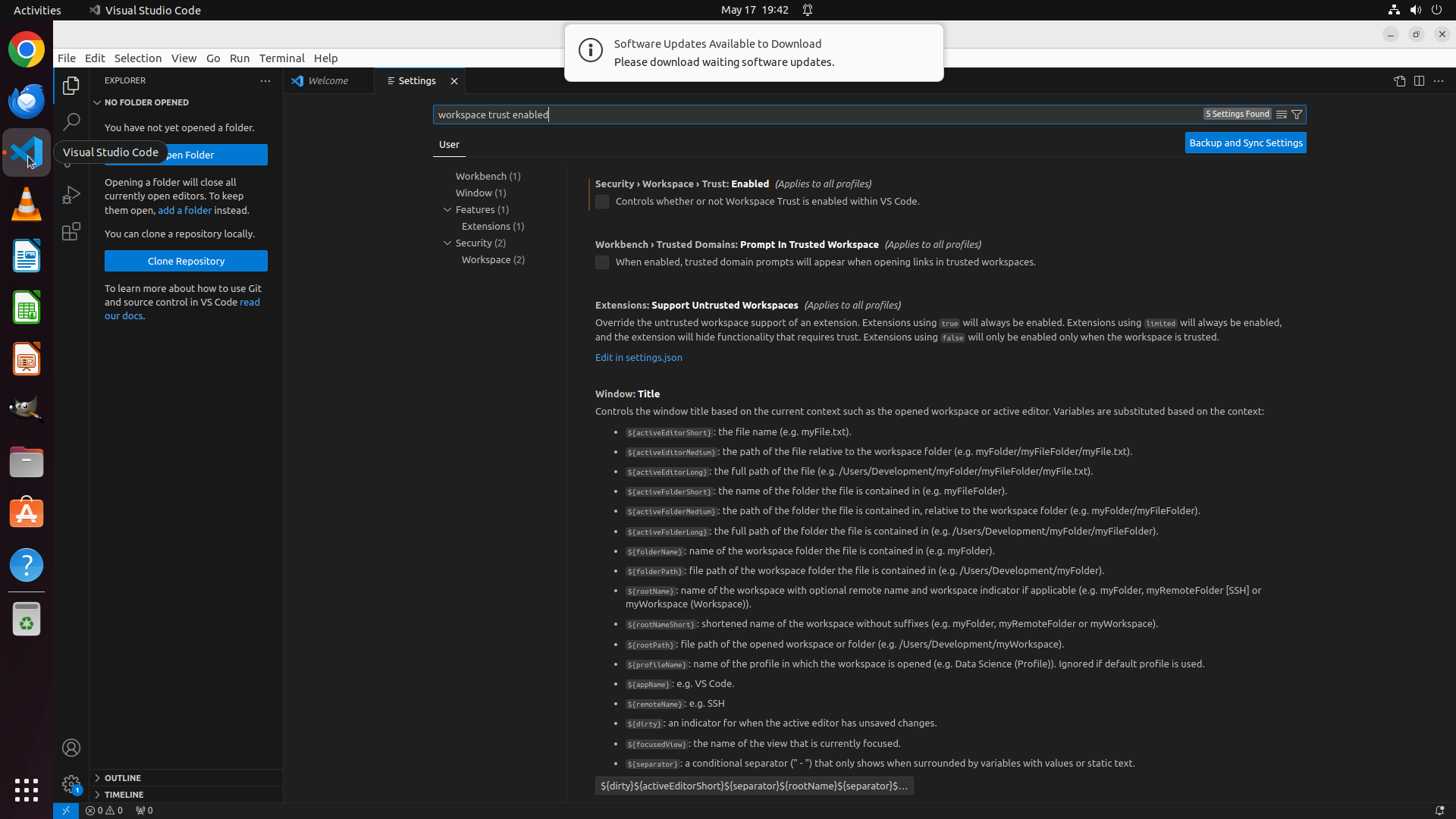}
\end{minipage} &
\scriptsize \textbf{[hotkey(key='ctrl   shift   p')]} I checked the results, but there's nothing in the search box. I'm aware of a better solution. &
\scriptsize \textbf{Hallucinates that the restored state has blank results, just like the recovery target state}, leading to completely novel behavior. 
\tabularnewline

\midrule
\footnotesize \textbf{State Tracking Failure} &
\scriptsize Find flights from Seattle to New York on 5th next month and only show those that can be purchased with miles. &
\begin{minipage}{\linewidth}
    \centering
    \includegraphics[width=1\textwidth]{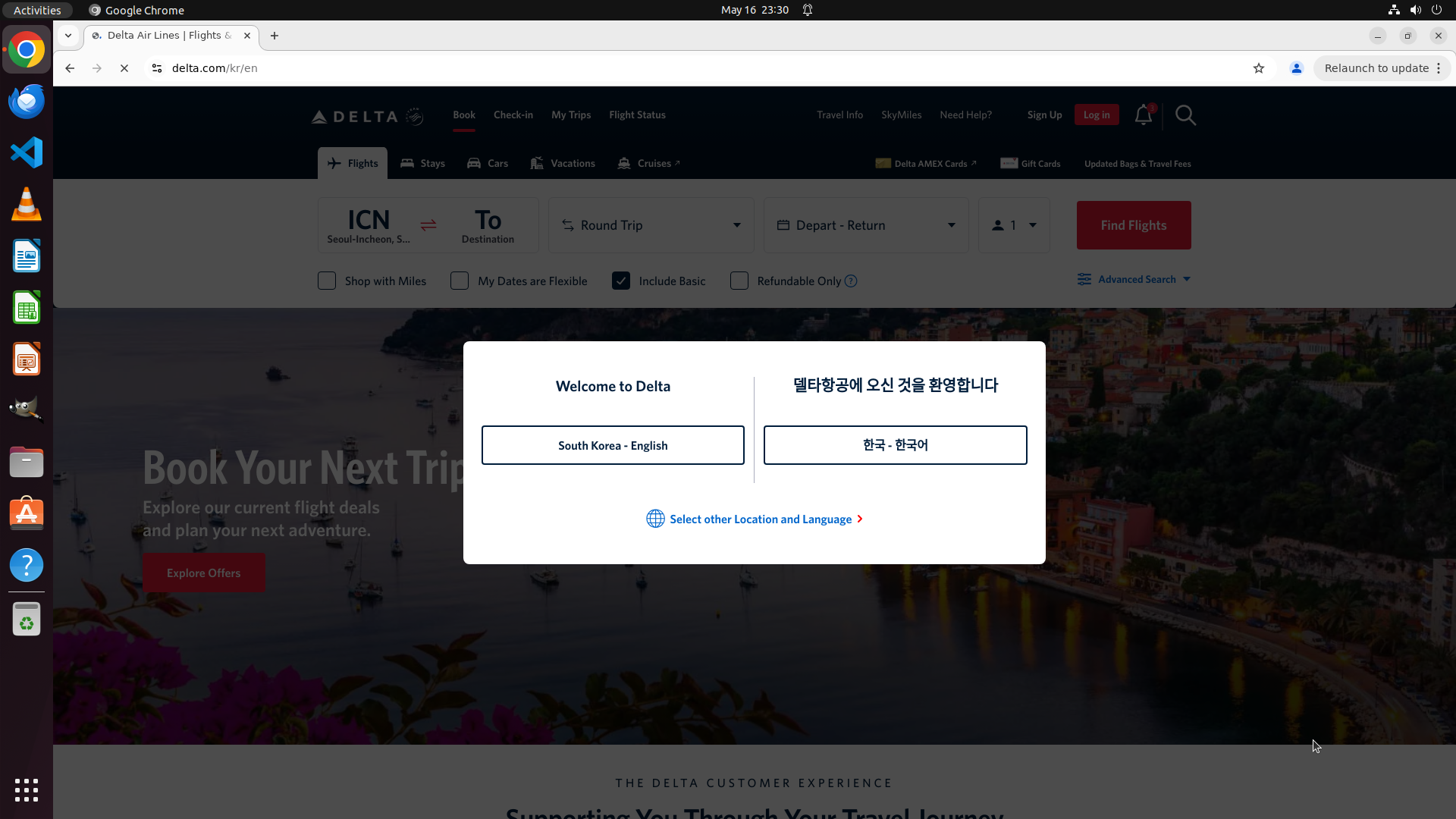}
\end{minipage} &
\begin{minipage}{\linewidth}
    \centering
    \includegraphics[width=1\textwidth]{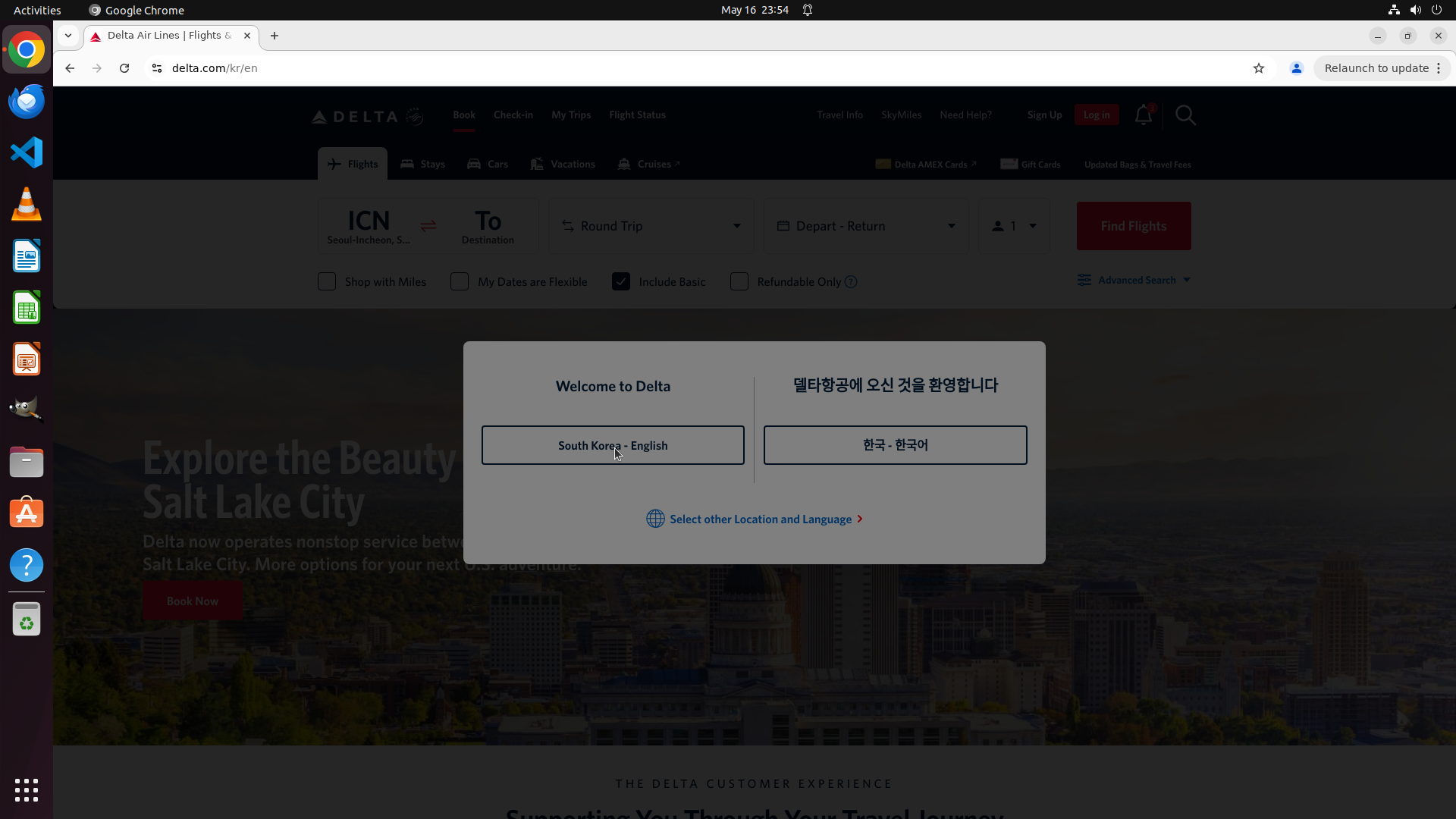}
\end{minipage} &
\scriptsize \textbf{[click]} I located the button labeled "South Korea - English" and clicked on it to continue booking my flight. &
\scriptsize \textbf{Hallucinates the current state as clickable based on the target-state observation}, failing to correctly synchronize with the actual system state.

\tabularnewline
\midrule
\footnotesize \textbf{State Tracking Failure} &
\scriptsize Set the background color to yellow for any slide that contains one or more images of real people, and set the title of slide 2 as "Let's start". &
\begin{minipage}{\linewidth}
    \centering
    \includegraphics[width=1\textwidth]{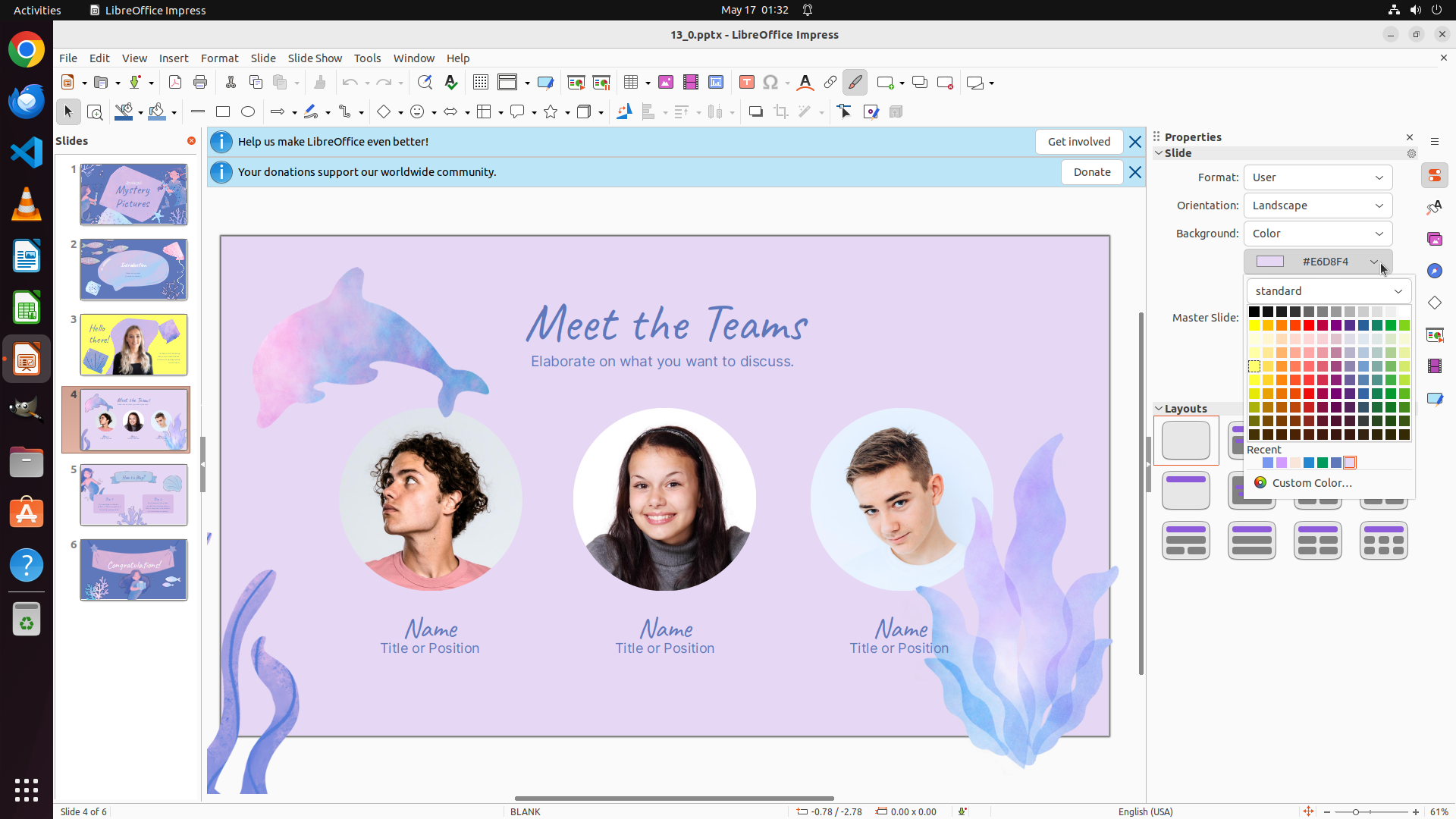}
\end{minipage} &
\begin{minipage}{\linewidth}
    \centering
    \includegraphics[width=1\textwidth]{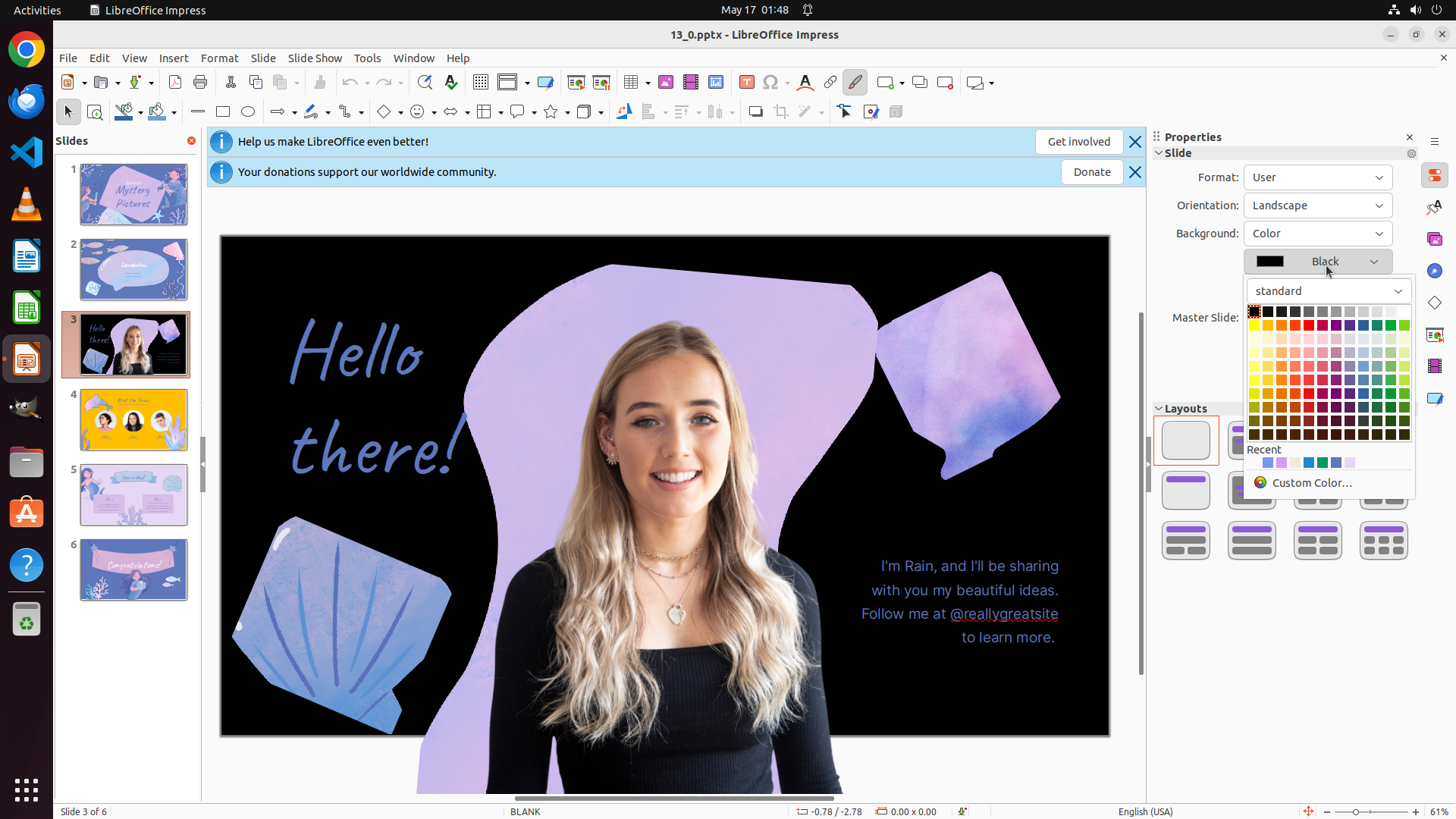}
\end{minipage} &
\scriptsize \textbf{[click]} I've already changed the background on the third slide to yellow, it's time to change fourth one. &
\scriptsize Hallucinates the current state as being on the fourth slide instead of third, unable to synchronize with the erroneous recovered state. 
\tabularnewline

\bottomrule
\end{tabular}
\onecolumn
\caption{Additional Qualitative analysis of erroneous agent behaviors induced by imperfect state restoration.}
\twocolumn
\label{tab:additional_qualitative_analysis}
\end{table}
\label{appendix:Qualitative Analysis}

\clearpage
\section{Human Evaluation Guidelines}
In this section, we provide the guidelines for human
evaluation provided to the participants in Table
\ref{tab:evaluation_preference_eval}.
\newtcolorbox{promptbox68}{
  colback=gray!4,
  colframe=black!70,
  boxrule=1.2pt,
  arc=3pt,
  left=7pt,
  right=7pt,
  top=1pt,
  bottom=1pt,
  title={Human Evaluation Guideline},
  width=\linewidth,
  fontupper=\ttfamily\small,
  enhanced,
  breakable,
}

\begin{promptbox68}
You will be provided with:\\
- Goal Screenshot\\
- Error-Step Screenshot\\
- Final Recovery Screenshot\\
- Original Instruction\\
- Recovery Trajectory\\
- Original Execution Trajectory\\
- Your task is to determine whether the recovery process successfully restores the environment to the intended goal state.\\

Evaluation Objective\\
- Evaluate whether the Final Recovery Screenshot is functionally equivalent to the Goal Screenshot.\\
    - Success (1): the environment is correctly restored\\
    - Failure (0): the environment is not properly restored\\

Evaluation Principles\\
1. Prioritize Visual Similarity\\
The primary signal is the visual similarity between the Final Recovery Screenshot and the Goal Screenshot. Please carefully compare the screenshots at a fine-grained, segment-level detail, including interaction-dependent UI elements. Pay close attention to whether:\\
- active windows or pages differ,\\
- selected content or tabs change,\\
- focused text boxes differ,\\
- dropdowns or menus are opened differently,\\
- cursor positions or hover states change,\\
- highlighted or selected UI elements differ,\\
- drag-and-drop states change, or other interaction-sensitive regions are inconsistent.\\
Mark as Failure (0) if important visual or interaction-level differences remain.\\
**Important** Minor differences caused by dynamic environments (e.g., notifications, pop-ups, timestamps, loading indicators, rendering variations) can be ignored if the underlying functional state remains equivalent.\\

2. Verify Functional Recovery\\
The recovered state should allow the original task to continue correctly.\\
Even if screenshots appear visually similar, mark as Failure (0) if:-\\
- important interaction context is missing,\\
- unintended changes remain,\\
- incorrect files/pages/windows are active,\\
or the operational state is not properly restored.\\
Use the recovery trajectory and original trajectory to understand:\\
- what actions changed the environment,\\
- whether erroneous states were actually reverted,\\
- and whether the recovery process is consistent with the intended goal state.\\
Trajectory information is especially important when screenshots appear visually similar but the           actual operational state may still differ.\\

Final Output\\
Binary Label: 1 for Success\\
Binary Label: 0 for Failure \\

Final Output (Optional If Recovery is Successful)\\
If you believe that the recovery is successful, please determine whether the successful recovery decision was despite ignorable observation differences caused by minor dynamic environmental changes such as notifications, pop-ups, timestamps, loading indicators, and rendering variations.\\
\\
Binary Label: 1 for YES\\
Binary Label: 0 for NO

\end{promptbox68}
\vspace{-0.15in}
{%
\captionof{figure}{Human Evaluation Guideline}
\label{app:fig:Human_Evaluation_Guideline}
}

\section{Prompts}
Here, we provide the prompts in our experiments.
\label{appendix:Prompts}
\newtcolorbox{promptbox455}{
  colback=pink!4,
  colframe=black!70,
  boxrule=1.2pt,
  arc=3pt,
  left=10pt,
  right=10pt,
  top=7pt,
  bottom=7pt,
  title=System Prompt for Claude-Sonnet-4.6 Forward Agent in OsWorld,
  width=\linewidth,
  fontupper=\ttfamily\small,
  enhanced,
  breakable,
}

\begin{promptbox455}
<SYSTEM\_CAPABILITY>\\
* You are utilising an Ubuntu virtual machine using x86\_64 architecture with internet access.\\
* You can feel free to install Ubuntu applications with your bash tool. Use curl instead of wget.
* To open browser, please just click on the Chrome icon.  Note, Chrome is what is installed on your system.\\
* Using bash tool you can start GUI applications, but you need to set export DISPLAY=:1 and use a subshell. For example "(DISPLAY=:1 xterm \&)". GUI apps run with bash tool will appear within your desktop environment, but they may take some time to appear. Take a screenshot to confirm it did.\\
* When using your bash tool with commands that are expected to output very large quantities of text, redirect into a tmp file and use str\_replace\_editor or `grep -n -B <lines before> -A <lines after> <query> <filename>` to confirm output.\\
* When viewing a page it can be helpful to zoom out so that you can see everything on the page.  Either that, or make sure you scroll down to see everything before deciding something isn't available.\\
* DO NOT ask users for clarification during task execution. DO NOT stop to request more information from users. Always take action using available tools.\\
* When using your computer function calls, they take a while to run and send back to you.  Where possible/feasible, try to chain multiple of these calls all into one function calls request.\\
* TASK FEASIBILITY: You can declare a task infeasible at any point during execution - whether at the beginning after taking a screenshot, or later after attempting some actions and discovering barriers. Carefully evaluate whether the task is feasible given the current system state, available applications, and task requirements. If you determine that a task cannot be completed due to:\\
  - Missing required applications or dependencies that cannot be installed\\
  - Insufficient permissions or system limitations\\
  - Contradictory or impossible requirements\\
  - Any other fundamental barriers that make completion impossible\\
  Then you MUST output exactly "[INFEASIBLE]" (including the square brackets) anywhere in your response to trigger the fail action. The system will automatically detect this pattern and terminate the task appropriately.\\
* The current date is \{current\_date\}.\\
* Home directory of this Ubuntu system is '/home/user'.\\
* If you need a password for sudo, the password of the computer is \{password\}.\\
</SYSTEM\_CAPABILITY>\\
\\
<IMPORTANT>\\
* If the item you are looking at is a pdf, if after taking a single screenshot of the pdf it seems that you want to read the entire document instead of trying to continue to read the pdf from your screenshots + navigation, determine the URL, use curl to download the pdf, install and use pdftotext to convert it to a text file, and then read that text file directly with your StrReplaceEditTool.\\
</IMPORTANT>

\end{promptbox455}
\vspace{-0.15in}
{
\captionof{figure}{System Prompt for Claude-Sonnet-4.6 Forward Agent in OsWorld}
\label{app:fig:sonnet_system_prompt_osworld}
}
\newtcolorbox{promptbox460}{
  colback=pink!4,
  colframe=black!70,
  boxrule=1.2pt,
  arc=3pt,
  left=10pt,
  right=10pt,
  top=6pt,
  bottom=6pt,
  title=System Prompt for Claude-Sonnet-4.6 Forward Agent in WindowsAgentArena,
  width=\linewidth,
  fontupper=\ttfamily\small,
  enhanced,
  breakable,
}

\begin{promptbox460}

<SYSTEM\_CAPABILITY>\\
* You are utilising a Windows virtual machine using x86\_64 architecture with internet access.\\
* To open browser, please just click on the Chrome icon.  Note, Chrome is what is installed on your system.\\
* When viewing a page it can be helpful to zoom out so that you can see everything on the page.  Either that, or make sure you scroll down to see everything before deciding something isn't available.\\
* When you want to open some applications on Windows, please use Double Click on it instead of clicking once.\\
</SYSTEM\_CAPABILITY>

\end{promptbox460}
\vspace{-0.15in}
{
\captionof{figure}{System Prompt for Claude-Sonnet-4.6 Forward Agent in WindowsAgentArena}
\label{app:fig:sonnet_system_prompt_windows}
}

\newtcolorbox{promptbox456}{
  colback=pink!4,
  colframe=black!70,
  boxrule=1.2pt,
  arc=3pt,
  left=10pt,
  right=10pt,
  top=10pt,
  bottom=10pt,
  title=System Prompt for Evocua-32B forward agent in OsWorld,
  width=\linewidth,
  fontupper=\ttfamily\small,
  enhanced,
  breakable,
}

\begin{promptbox456}
\# Tools\\
You may call one or more functions to assist with the user query.\\
\\
You are provided with function signatures within <tools></tools> XML tags:\\

<tools>\\
\{tools\_xml\}\\
</tools>\\
\\
For each function call, return a json object with function name and arguments within <tool\_call></tool\_call> XML tags:\\

<tool\_call>\\
\{"name": <function-name>, "arguments": <args-json-object>\}\\
</tool\_call>\\
\\
The current date is \{date\}\\
If you need a password for sudo, the password of the computer is \{client\_password\}\\
\\
\# Response format\\
Response format for every step:\\
1) Action: a short imperative describing what to do in the UI.\\
2) A single <tool\_call>...</tool\_call> block containing only the JSON: \{"name": <function-name>, "arguments": <args-json-object>\}.\\
\\
Rules:\\
- Output exactly in the order: Action, <tool\_call>.\\
- Be brief: one sentence for Action.\\
- Do not output anything else outside those parts.\\
- If finishing, use action=terminate in the tool call.
\end{promptbox456}
\vspace{-0.2in}
{
\captionof{figure}{System Prompt for Evocua-32B forward agent in OsWorld}
\label{app:fig:system_prompt_osworld_evocua}
}
\newtcolorbox{promptbox461}{
  colback=pink!4,
  colframe=black!70,
  boxrule=1.2pt,
  arc=3pt,
  left=10pt,
  right=10pt,
  top=10pt,
  bottom=10pt,
  title=System Prompt used for Evocua-32B forward agent in WindowsAgentArena,
  width=\linewidth,
  fontupper=\ttfamily\small,
  enhanced,
  breakable,
}

\begin{promptbox461}
\# Tools \\
You may call one or more functions to assist with the user query.\\

You are provided with function signatures within <tools></tools> XML tags:\\

<tools>\\
\{tools\_xml\}\\
</tools>\\
\\
For each function call, return a json object with function name and arguments within <tool\_call></tool\_call> XML tags:\\

<tool\_call>\\
\{"name": <function-name>, "arguments": <args-json-object>\}\\
</tool\_call>\\
\\
\# Response format
\\
Response format for every step:\\
1) Action: a short imperative describing what to do in the UI.\\
2) A single <tool\_call>...</tool\_call> block containing only the JSON: \{"name": <function-name>, "arguments": <args-json-object>\}.\\
\\
Rules:\\
- Output exactly in the order: Action, <tool\_call>.\\
- Be brief: one sentence for Action.\\
- Do not output anything else outside those parts.\\
- If finishing, use action=terminate in the tool call.
\end{promptbox461}

\vspace{-0.15in}
{
\captionof{figure}{System Prompt used for Evocua-32B forward agent in WindowsAgentArena}
\label{app:fig:system_prompt_windows_evocua}
}
\newtcolorbox{promptbox465}{
  colback=pink!4,
  colframe=black!70,
  boxrule=1.2pt,
  arc=3pt,
  left=10pt,
  right=10pt,
  top=10pt,
  bottom=10pt,
  title=Tool description prompt used for Evocua-32B agent,
  width=\linewidth,
  fontupper=\ttfamily\small,
  enhanced,
  breakable,
}

\begin{promptbox465}
\{
"type": "function", \\
\hspace*{1em} "function": \{\\
\hspace*{2em} "name\_for\_human": "computer\_use",\\
\hspace*{2em} "name": "computer\_use", \\
\hspace*{2em} "description": description\_prompt,\\
\hspace*{2em} "parameters": \{\\
\hspace*{3em} "properties": \{\\
\hspace*{4em} "action": \{\\
\hspace*{5em} "description": S2\_ACTION\_DESCRIPTION,\\
\hspace*{5em} "enum": ["key", "type", "mouse\_move", "left\_click", "left\_click\_drag", \\
\hspace*{5em} "right\_click", "middle\_click", "double\_click", "triple\_click", \\ 
\hspace*{5em} "scroll", "wait", "terminate", "key\_down", "key\_up"], \\
\hspace*{5em} "type": "string"\\
\hspace*{4em} \},\\
\hspace*{4em} "keys": \{\\
\hspace*{5em} "description": "Required only by `action=key`.",\\
\hspace*{5em} "type": "array"\\
\hspace*{4em} \}, \\
\hspace*{4em} "text": \{\\
\hspace*{5em} "description": "Required only by `action=type`.", \\
\hspace*{5em} "type": "string"\\
\hspace*{4em} \}, \\
\hspace*{4em} "coordinate": \{\\
\hspace*{5em} "description": "The x,y coordinates for mouse actions.", \\
\hspace*{5em} "type": "array"\\
\hspace*{4em} \}, \\
\hspace*{4em} "pixels": \{\\
\hspace*{5em} "description": "The amount of scrolling.", \\
\hspace*{5em} "type": "number"\\
\hspace*{4em} \}, \\
\hspace*{4em} "time": \{"description": "The seconds to wait.", "type": "number"\}, \\
\hspace*{4em} "status": \{\\
\hspace*{5em} "description": "The status of the task.", \\
\hspace*{5em} "type": "string", \\
\hspace*{5em} "enum": ["success", "failure"]\\
\hspace*{4em} \}\\
\hspace*{3em} \}, \\
\hspace*{3em} "required": ["action"],\\ 
\hspace*{3em} "type": "object"\\
\hspace*{2em} \}, \\
\hspace*{2em} "args\_format": "Format the arguments as a JSON object."\\
\hspace*{2em} \}\\
\}
\end{promptbox465}

\vspace{-0.15in}
{
\captionof{figure}{Tool description prompt used for Evocua-32B agent}
\label{app:fig:tool_definition_evocua}
}
\newtcolorbox{promptbox458}{
  colback=pink!4,
  colframe=black!70,
  boxrule=1.2pt,
  arc=3pt,
  left=10pt,
  right=10pt,
  top=10pt,
  bottom=10pt,
  title=Description Prompt used for Evocua-32B agent,
  width=\linewidth,
  fontupper=\ttfamily\small,
  enhanced,
  breakable,
}

\begin{promptbox458}
Use a mouse and keyboard to interact with a computer, and take screenshots.\\
* This is an interface to a desktop GUI. You must click on desktop icons to start applications.\\
* Some applications may take time to start or process actions, so you may need to wait and take successive screenshots to see the results of your actions. E.g. if you click on Firefox and a window doesn't open, try wait and taking another screenshot.\\
\{resolution\_info\}\\
* Whenever you intend to move the cursor to click on an element like an icon, you should consult a screenshot to determine the coordinates of the element before moving the cursor.\\
* If you tried clicking on a program or link but it failed to load even after waiting, try adjusting your cursor position so that the tip of the cursor visually falls on the element that you want to click.\\
* Make sure to click any buttons, links, icons, etc with the cursor tip in the center of the element. Don't click boxes on their edges unless asked.
\end{promptbox458}
{
\vspace{-0.15in}
\captionof{figure}{Description Prompt used for Evocua-32B agent}
\label{app:fig:description_prompt_evocua}
}
\newtcolorbox{promptbox457}{
  colback=pink!4,
  colframe=black!70,
  boxrule=1.2pt,
  arc=3pt,
  left=10pt,
  right=10pt,
  top=10pt,
  bottom=10pt,
  title=Action description prompt used for Evocua-32B agent,
  width=\linewidth,
  fontupper=\ttfamily\small,
  enhanced,
  breakable,
}

\begin{promptbox457}
* `key`: Performs key down presses on the arguments passed in order, then performs key releases in reverse order.\\
* `key\_down`: Press and HOLD the specified key(s) down in order (no release). Use this for stateful holds like holding Shift while clicking.\\
* `key\_up`: Release the specified key(s) in reverse order.\\
* `type`: Type a string of text on the keyboard.\\
* `mouse\_move`: Move the cursor to a specified (x, y) pixel coordinate on the screen.\\
* `left\_click`: Click the left mouse button at a specified (x, y) pixel coordinate on the screen.\\
* `left\_click\_drag`: Click and drag the cursor to a specified (x, y) pixel coordinate on the screen.\\
* `right\_click`: Click the right mouse button at a specified (x, y) pixel coordinate on the screen.\\
* `middle\_click`: Click the middle mouse button at a specified (x, y) pixel coordinate on the screen.\\
* `double\_click`: Double-click the left mouse button at a specified (x, y) pixel coordinate on the screen.\\
* `triple\_click`: Triple-click the left mouse button at a specified (x, y) pixel coordinate on the screen.\\
* `scroll`: Performs a scroll of the mouse scroll wheel.\\
* `hscroll`: Performs a horizontal scroll (mapped to regular scroll).\\
* `wait`: Wait specified seconds for the change to happen.\\
* `terminate`: Terminate the current task and report its completion status.
* `answer`: Answer a question.
\end{promptbox457}
{
\vspace{-0.15in}
\captionof{figure}{Action description prompt used for Evocua-32B agent}
\label{app:fig:action_description_prompt_evocua}
}
\newtcolorbox{promptbox619}{
  colback=pink!4,
  colframe=black!70,
  boxrule=1.2pt,
  arc=3pt,
  left=10pt,
  right=10pt,
  top=5pt,
  bottom=5pt,
  title=Action space used for UI-Tars-1.5-7B forward agent,
  width=\linewidth,
  fontupper=\ttfamily\small,
  enhanced,
  breakable,
}

\begin{promptbox619}
click(start\_box='<|box\_start|>\\(x1,y1)<|box\_end|>')\\
left\_double(start\_box='<|box\_start|>\\(x1,y1)<|box\_end|>')\\
right\_single(start\_box='<|box\_start|>\\(x1,y1)<|box\_end|>')\\
drag(start\_box='<|box\_start|>\\(x1,y1)<|box\_end|>', \\ end\_box='<|box\_start|>\\(x3,y3)<|box\_end|>')
hotkey(key='')\\
type(content='') \#If you want to submit your input, use " \\n" at the end of `content`.\\
scroll(start\_box='<|box\_start|>\\(x1,y1)<|box\_end|>', direction='down or up or right or left')\\
wait() \#Sleep for 5s and take a screenshot to check for any changes.\\
finished(content='xxx') \# Use escape characters \\', \\", and \\n in content part to ensure we can parse the content in normal python string format.
\end{promptbox619}

{
\vspace{-0.15in}
\captionof{figure}{Action space used for UI-Tars-1.5-7B forward agent}
\label{app:fig:uitars_action_space}
}
\newtcolorbox{promptbox459}{
  colback=pink!4,
  colframe=black!70,
  boxrule=1.2pt,
  arc=3pt,
  left=10pt,
  right=10pt,
  top=5pt,
  bottom=5pt,
  title=System prompt used for GPT-5.4-CUA agent in OsWorld benchmark,
  width=\linewidth,
  fontupper=\ttfamily\small,
  enhanced,
  breakable,
}

\begin{promptbox459}
Here are some helpful tips:\\
- My computer password is "\{CLIENT\_PASSWORD\}" when sudo is needed.\\
- Stick to the website or application already opened for the task when possible.\\
- Prefer Chrome over Firefox/Chromium unless the task says otherwise.\\
- You can act without asking for confirmation.\\
- If the task is infeasible, say so explicitly.
\end{promptbox459}

{
\vspace{-0.15in}
\captionof{figure}{System prompt used for GPT-5.4-CUA agent in OsWorld benchmark}
\label{app:fig:system_prompt_osworld_gpt54}
}
\newtcolorbox{promptbox463}{
  colback=pink!4,
  colframe=black!70,
  boxrule=1.2pt,
  arc=3pt,
  left=10pt,
  right=10pt,
  top=10pt,
  bottom=10pt,
  title=System prompt used for QWEN-3.5-9B forward agent in OsWorld,
  width=\linewidth,
  fontupper=\ttfamily\small,
  enhanced,
  breakable,
}

\begin{promptbox463}
You are a multi-purpose intelligent assistant. Based on my requests, you can use tools to help me complete various tasks.\\

\# Tools\\
You have access to the following functions:\\

<tools>\\
\{json.dumps(tools\_def)\}\\
</tools>\\
\\
If you choose to call a function ONLY reply in the following format with NO suffix:\\
<tool\_call>\\
<function=example\_function\_name>\\
<parameter=example\_parameter\_1>\\
value\_1\\
</parameter>\\
<parameter=example\_parameter\_2>\\
This is the value for the second parameter\\
that can span\\
multiple lines\\
</parameter>\\
</function>\\
</tool\_call>\\\\
<IMPORTANT>\\
Reminder:\\
- Function calls MUST follow the specified format: an inner <function=...></function> block must be nested within <tool\_call></tool\_call> XML tags\\
- Required parameters MUST be specified\\
- You may provide optional reasoning for your function call in natural language BEFORE the function call, but NOT after\\
- If there is no function call available, answer the question like normal with your current knowledge and do not tell the user about function calls\\
- The current date is \{datetime.today().strftime('\%A, \%B \%d, \%Y')\}.\\
- Collapsed screenshots appear as text: \{self.collapse\_text\}\\
If you need a password for sudo, the password of the computer is \{self.client\_password\}\\
</IMPORTANT>\\ 

\# Response format\\
Response format for every step:\\
1) Action: a short imperative describing what to do in the UI.\\
2) A single <tool\_call>...</tool\_call> block.\\\\

Rules:\\
- Output exactly in the order: Action, <tool\_call>.\\
- Be brief: one sentence for Action.\\
- Do not output anything else outside those parts.\\
- If finishing, use action=terminate in the tool call.
\end{promptbox463}

{
\vspace{-0.15in}
\captionof{figure}{System prompt used for QWEN-3.5-9B forward agent in OsWorld}
\label{app:fig:system_prompt_osworld_qwen}
}
\newtcolorbox{promptbox462}{
  colback=pink!4,
  colframe=black!70,
  boxrule=1.2pt,
  arc=3pt,
  left=10pt,
  right=10pt,
  top=10pt,
  bottom=10pt,
  title=System prompt used for GPT-5.4-CUA agent in WindowsAgentArena,
  width=\linewidth,
  fontupper=\ttfamily\small,
  enhanced,
  breakable,
}

\begin{promptbox462}
Here are some helpful tips:\\
- Stick to the website or application already opened for the task when possible.\\
- Prefer Chrome over Firefox/Chromium unless the task says otherwise.\\
- You can act without asking for confirmation.\\
- If the task is infeasible, say so explicitly.
\end{promptbox462}

{
\vspace{-0.15in}
\captionof{figure}{System prompt used for GPT-5.4-CUA agent in WindowsAgentArena}
\label{app:fig:system_prompt_windows_gpt54}
}
\newtcolorbox{promptbox12}{
  colback=pink!4,
  colframe=black!70,
  boxrule=1.2pt,
  arc=3pt,
  left=10pt,
  right=10pt,
  top=10pt,
  bottom=10pt,
  title=System prompt used for Qwen-3.5-9B agent in WindowsAgentArena,
  width=\linewidth,
  fontupper=\ttfamily\small,
  enhanced,
  breakable,
}

\begin{promptbox12}

You are a multi-purpose intelligent assistant. Based on my requests, you can use tools to help me complete various tasks.\\\\
\# Tools\\\\
You have access to the following functions:\\\\
<tools>\\
\{json.dumps(tools\_def)\}\\
</tools>\\\\
If you choose to call a function ONLY reply in the following format with NO suffix:\\\\
<tool\_call>\\
<function=example\_function\_name>\\
<parameter=example\_parameter\_1>\\
value\_1
</parameter>\\
<parameter=example\_parameter\_2>
This is the value for the second parameter\\
that can span\\
multiple lines\\
</parameter>\\
</function>\\
</tool\_call>\\\\
<IMPORTANT>\\
Reminder:\\
- Function calls MUST follow the specified format: an inner <function=...></function> block must be nested within <tool\_call></tool\_call> XML tags
- Required parameters MUST be specified\\
- You may provide optional reasoning for your function call in natural language BEFORE the function call, but NOT after\\
- If there is no function call available, answer the question like normal with your current knowledge and do not tell the user about function calls\\
- Collapsed screenshots appear as text: \{self.collapse\_text\}\\
</IMPORTANT>\\\\
\# Response format\\\\
Response format for every step:\\
1) Action: a short imperative describing what to do in the UI.\\
2) A single <tool\_call>...</tool\_call> block.\\\\
Rules:\\
- Output exactly in the order: Action, <tool\_call>.\\
- Be brief: one sentence for Action.\\
- Do not output anything else outside those parts.\\
- If finishing, use action=terminate in the tool call.
\end{promptbox12}

{
\vspace{-0.15in}
\captionof{figure}{System prompt used for Qwen-3.5-9B agent in WindowsAgentArena}
\label{app:fig:system_prompt_windows_qwen}
}
\newtcolorbox{promptbox8}{
  colback=pink!4,
  colframe=black!70,
  boxrule=1.2pt,
  arc=3pt,
  left=10pt,
  right=10pt,
  top=10pt,
  bottom=10pt,
  title=Action description prompt used for QWEN3.5-9B agent,
  width=\linewidth,
  fontupper=\ttfamily\small,
  enhanced,
  breakable,
}

\begin{promptbox8}
* `key`: Performs key down presses on the arguments passed in order, then performs key releases in reverse order.\\
* `type`: Type a string of text on the keyboard.\\
* `mouse\_move`: Move the cursor to a specified (x, y) pixel coordinate on the screen.\\
* `left\_click`: Click the left mouse button at a specified (x, y) pixel coordinate on the screen. Optional `text` parameter can specify modifier keys (e.g., "ctrl", "shift", "ctrl+shift") that will be held during the click.\\
* `left\_click\_drag`: Click and drag the cursor to a specified (x, y) coordinate.\\
* `right\_click`: Click the right mouse button at a specified (x, y) pixel coordinate on the screen.\\ Optional `text` parameter can specify modifier keys that will be held during the click.\\
* `middle\_click`: Click the middle mouse button at a specified (x, y) pixel coordinate on the screen. Optional `text` parameter can specify modifier keys that will be held during the click.\\
* `double\_click`: Double-click the left mouse button at a specified (x, y) pixel coordinate on the screen. Optional `text` parameter can specify modifier keys that will be held during the click.\\
* `triple\_click`: Triple-click the left mouse button at a specified (x, y) pixel coordinate on the screen (simulated as double-click since it's the closest action). Optional `text` parameter can specify modifier keys that will be held during the click.\\
* `scroll`: Performs a scroll of the mouse scroll wheel. Optional `text` parameter can specify a modifier key (e.g., "shift", "ctrl") that will be held during scrolling.\\
* `hscroll`: Performs a horizontal scroll (mapped to regular scroll). Optional `text` parameter can specify a modifier key that will be held during scrolling.\\
* `wait`: Wait specified seconds for the change to happen.\\
* `terminate`: Terminate the current task and report its completion status.\\
* `answer`: Answer a question.
\end{promptbox8}

{
\vspace{-0.15in}
\captionof{figure}{Action description prompt used for QWEN3.5-9B agent}
\label{app:fig:action_description_prompt_qwen}
}
\newtcolorbox{promptbox9}{
  colback=pink!4,
  colframe=black!70,
  boxrule=1.2pt,
  arc=3pt,
  left=10pt,
  right=10pt,
  top=10pt,
  bottom=10pt,
  title=User prompt used for UI-Tars-1.5-7B forward agent,
  width=\linewidth,
  fontupper=\ttfamily\small,
  enhanced,
  breakable,
}

\begin{promptbox9}
You are a GUI agent. You are given a task and your action history, with screenshots. You need to perform the next action to complete the task. \\

\#\# Output Format\\
```\\
Thought: ...\\
Action: ...\\
```\\

\#\# Action Space\\
\{action\_space\}\\

\#\# Note\\
- Use \{language\} in `Thought` part.\\
- Write a small plan and finally summarize your next action (with its target element) in one sentence in `Thought` part.\\

\#\# User Instruction\\
\{instruction\}
\end{promptbox9}

{
\vspace{-0.15in}
\captionof{figure}{User prompt used for UI-Tars-1.5-7B forward agent}
\label{app:fig:uitars_user_prompt}
}
\newtcolorbox{promptbox470}{
  colback=pink!4,
  colframe=black!70,
  boxrule=1.2pt,
  arc=3pt,
  left=10pt,
  right=10pt,
  top=10pt,
  bottom=10pt,
  title=Tool description prompt used for QWEN3.5-9B agent,
  width=\linewidth,
  fontupper=\ttfamily\small,
  enhanced,
  breakable,
}

\begin{promptbox470}
"type": "function",\\
\hspace*{1em} "function": \{\\
\hspace*{2em} "name": "computer\_use",\\
\hspace*{2em} "description": description\_prompt,\\
\hspace*{2em} "parameters": \{\\
\hspace*{3em} "type": "object",\\
\hspace*{3em} "required": ["action"],\\
\hspace*{3em} "properties": \{\\
\hspace*{4em} "action": \{ \\
\hspace*{5em} "type": "string",\\
\hspace*{5em} "description": action\_description\_prompt,\\
\hspace*{5em} "enum": [\\
\hspace*{6em} "key",\\
\hspace*{6em} "type",\\
\hspace*{6em} "mouse\_move",\\
\hspace*{6em} "left\_click",\\
\hspace*{6em} "left\_click\_drag",\\
\hspace*{6em} "right\_click",\\
\hspace*{6em} "middle\_click",\\
\hspace*{6em} "double\_click",\\
\hspace*{6em} "triple\_click",\\
\hspace*{6em} "scroll",\\
\hspace*{6em} "hscroll",\\
\hspace*{6em} "wait",\\
\hspace*{6em} "terminate",\\
\hspace*{6em} "answer",\\
\hspace*{6em}  ],\\
\hspace*{5em} \},\\
\hspace*{4em} "keys": \{\\
\hspace*{5em} "type": "array",\\ 
\hspace*{5em} "description": "Required only by `action=key`."\},\\
\hspace*{4em} "text": \{\\
\hspace*{5em} "type": "string",\\
\hspace*{5em} "description": "Required by `action=type` and `action=answer`. \hspace*{6em} Optional for click actions (left\_click, right\_click, middle\_click, \hspace*{6em} double\_click, triple\_click) to specify modifier keys (e.g., 'ctrl', \hspace*{6em} 'shift', 'ctrl+shift'). Optional for scroll actions (scroll, hscroll) to specify a modifier key (e.g., 'shift', 'ctrl') to hold during scrolling.",\\
\hspace*{4em} \},\\
\hspace*{4em} "coordinate": \{\\
\hspace*{5em} "type": "array", "description": "(x, y) coordinates."
\hspace*{5em} \},\\
\hspace*{4em} "pixels": \{\\
\hspace*{5em} "type": "number", "description": "Scroll amount."\},\\
\hspace*{5em} "time": \{"type": "number", "description": "Seconds to wait."\},\\
\hspace*{5em} "status": \{\\
\hspace*{6em} "type": "string",\\
\hspace*{6em} "description": "Task status for terminate.",\\
\hspace*{6em} "enum": ["success", "failure"],\\
\hspace*{5em}  \},\\
\hspace*{4em}  \},\\
\hspace*{3em}  \},\\
\hspace*{2em}  \},
\end{promptbox470}

{
\vspace{-0.15in}
\captionof{figure}{Tool description prompt used for QWEN3.5-9B agent}
\label{app:fig:tool_definition_qwen}
}
\newtcolorbox{promptbox37}{
  colback=blue!4,
  colframe=black!70,
  boxrule=1.2pt,
  arc=3pt,
  left=10pt,
  right=10pt,
  top=10pt,
  bottom=10pt,
  title=Prefix prompt used for perturbation agent,
  width=\linewidth,
  fontupper=\ttfamily\small,
  enhanced,
  breakable,
}

\begin{promptbox37}

\small
\ttfamily

<Perturbation\_AGENT\_ROLE>\\
* You are a Perturbation Agent. Your primary objective is to generate series of INCORRECT actions that causes a significant DRIFT from the user's original instruction, such that the observation is completely different.\\
* Your series of generated actions should lead to a completely unrelated state to the current instruction, forcing the system state into an unexpected or divergent path, where there it shows completely different scenery.\\
* You will be provided with the current state and a "correct" action (the ground truth) from the given state.\\
* Your initially generated action must be DIFFERENT from the provided action and should actively disrupt the flow of the task, leading to a completely different state. Try to make the path diverge as much as possible until you believe that the scenery is completely different.\\
* Do not make a minor mistake; perform an active, significantly deviant steps that has nothing to do with the original goal.\\
* While trying to create a significant drift, try your best effort to signficantly drift the current state via a single action, instead of doing a series of small mistakes.\\
</Perturbation\_AGENT\_ROLE>\\

<Perturbation\_COMPLETION>\\
* To complete your task, you must output a series of actions that satisfies the following criteria, which you will be also given the history of disruptive flow of observation and actions you have performed:\\
\hspace{0.2in} 1. It is DIFFERENT from the provided correct action.\\
\hspace{0.2in} 2. It is a valid, executable action in the current state.\\
\hspace{0.2in} 3. It causes a SIGNIFICANT DRIFT from the intended goal (completely unrelated or divergent), to the point where the observation completely changes to different scenery.\\
\hspace{0.2in} 4. Since the given correct action is from the initial state, try to make the action only different from the given action when there are no past perturbation history, meaning that
the correct action only applies to the inital state, and not for the rest of the perturbation process.\\
\hspace{0.2in} * Once you have generated this single unrelated action, that constitutes the completion of your turn, which you must perform multiple times until the system state is completely different from the original instruction.\\
\hspace{0.2in} * While trying to create a significant drift, try your best effort to signficantly drift the current state via a single action, instead of doing a series of small mistakes. You should not try to complete the original task at all, and you must make a mistake or go off-track intentionally.\\
</Perturbation\_COMPLETION>

\end{promptbox37}

{
\vspace{-0.15in}
\captionof{figure}{Prefix prompt used for perturbation agent}
\label{app:fig:perturbation_prefix_message}
}
\newtcolorbox{promptbox27}{
  colback=blue!4,
  colframe=black!70,
  boxrule=1.2pt,
  arc=3pt,
  left=10pt,
  right=10pt,
  top=10pt,
  bottom=10pt,
  title=Prefix prompt used for GPT-5.4-CUA perturbation agent,
  width=\linewidth,
  fontupper=\ttfamily\small,
  enhanced,
  breakable,
}

\begin{promptbox27}

\small
\ttfamily

<Perturbation\_AGENT\_ROLE>\\
* You are a Perturbation Agent. Your primary objective is to generate a series of INCORRECT computer-use actions that cause a significant DRIFT from the user's original instruction, such that the resulting observation becomes completely different from what the original task expects.\\
* Your generated actions must lead to a completely unrelated state, forcing the system into an unexpected or divergent path where the screen shows completely different scenery.\\
* You will be provided with the current screenshot, the user's original instruction, and the 'correct' action (the ground truth) from the initial state.\\
* Your initially generated action must be DIFFERENT from the provided correct action and should actively disrupt the flow of the task, leading to a completely different state. Make the path diverge as much as possible until the scenery is completely different.\\
* Do not make a minor mistake; perform an active, significantly deviant step that has nothing to do with the original goal.\\
* Try your best effort to significantly drift the current state via a single action, instead of doing a series of small mistakes.\\
* Do not output waiting actions that do not alter anything, such as "wait". You must generate an active, significantly deviant step that has nothing to do with the original goal, instead of making a minor mistake, meaning that the
observation and the state should drasticially change.\\ 
</Perturbation\_AGENT\_ROLE>\\

<Perturbation\_COMPLETION>\\
* To complete your task, you must output a series of computer-use actions that satisfy the following criteria:\\
\hspace{0.2in} 1. The action is DIFFERENT from the provided correct action.\\
\hspace{0.2in} 2. The action is valid and executable in the current state.\\
\hspace{0.2in} 3. The action causes a SIGNIFICANT DRIFT from the intended goal (completely unrelated or divergent), to the point where the observation completely changes to different scenery.\\
\hspace{0.2in} 4. Since the given correct action is from the initial state, only the initial action must differ from the given correct action; subsequent actions only need to keep diverging from the original task.\\
* Once the system state is completely different from the original instruction's expected scenery, you may stop emitting actions.\\
* You should NOT try to complete the original task at all. You MUST go off-track intentionally.\\
</Perturbation\_COMPLETION>
\end{promptbox27}

{
\vspace{-0.15in}
\captionof{figure}{Prefix prompt used for GPT-5.4-CUA perturbation agent}
\label{app:fig:openai_cua_perturbation_prefix_prompt}
}
\newtcolorbox{promptbox379}{
  colback=green!4,
  colframe=black!70,
  boxrule=1.2pt,
  arc=3pt,
  left=10pt,
  right=10pt,
  top=10pt,
  bottom=10pt,
  title=Prefix prompt used for recovery agent (0-E),
  width=\linewidth,
  fontupper=\ttfamily\small,
  enhanced,
  breakable,
}

\begin{promptbox379}

\small
\ttfamily
<RECOVERY\_AGENT\_ROLE>\\
* You are a Recovery Agent. Your primary objective is to navigate from the current Error State to the Goal State.
* The Goal State image is labeled with "This is the GOAL STATE you need to return to" - this is your target destination.\\
* The current state (Error Step) is where an error or deviation occurred, and you must find the path back to the Goal State.\\
* Analyze the differences between the current state and Goal State, then execute actions to bridge that gap.\\
* Focus on efficiency: take the most direct path to recovery while ensuring accuracy.\\
</RECOVERY\_AGENT\_ROLE>\\

<RECOVERY\_COMPLETION>\\
* When you believe you have successfully returned to the Goal State:\\
\hspace{0.1in} 1. Take a screenshot to verify the current state matches the Goal State\\
\hspace{0.1in} 2. Compare visual elements carefully:\\
\hspace{0.2in} - Window positions and sizes\\
\hspace{0.2in} - UI component states (buttons, menus, dialogs)\\
\hspace{0.2in} - Text content and input fields\\
\hspace{0.2in} - Application state and open windows\\
\hspace{0.2in} - Cursor position if relevant\\
\hspace{0.1in} 3. If the states match, output the "done" action to complete the recovery\\
\hspace{0.1in} 4. If they don't match, identify remaining discrepancies and continue taking recovery actions\\
* Be thorough in your comparison - minor differences may indicate incomplete recovery\\
* Do not declare completion until you have verified through a fresh screenshot\\
</RECOVERY\_COMPLETION>
\end{promptbox379}

{
\vspace{-0.15in}
\captionof{figure}{Prefix prompt used for recovery agent (0-E)}
\label{app:fig:recovery_prefix_message_1_E}
}
\newtcolorbox{promptbox384}{
  colback=green!4,
  colframe=black!70,
  boxrule=1.2pt,
  arc=3pt,
  left=10pt,
  right=10pt,
  top=10pt,
  bottom=10pt,
  title=Recovery Instruction for recovery agent (0-E),
  width=\linewidth,
  fontupper=\ttfamily\small,
  enhanced,
  breakable,
}

\begin{promptbox384}

\#\# RECOVERY TASK

You are a Recovery Agent. Your task is to navigate the system back to the Goal State.\\

\#\#\# Goal State\\
The Goal State image is labeled with "This is the GOAL STATE you need to return to".\\
Use the available tools to navigate the system back to that state.\\
\\
\#\#\# Current Status\\
- Current step: Error Step \{error\_step\}\\
- Target: Return to Goal State (Step \{goal\_step\})\\

\#\#\# Your Task\\
Analyze the difference between the current state and the Goal State, then take the necessary actions step by step to reach the Goal State.
\end{promptbox384}

{
\vspace{-0.15in}
\captionof{figure}{Recovery Instruction for recovery agent (0-E)}
\label{app:fig:recovery_instruction_1_E}
}
\newtcolorbox{promptbox92}{
  colback=green!4,
  colframe=black!70,
  boxrule=1.2pt,
  arc=3pt,
  left=10pt,
  right=10pt,
  top=10pt,
  bottom=10pt,
  title=Prefix prompt used for recovery agent (0-R),
  width=\linewidth,
  fontupper=\ttfamily\small,
  enhanced,
  breakable,
}

\begin{promptbox92}

\small
\ttfamily
<RECOVERY\_AGENT\_ROLE>\\
* You are a Recovery Agent. Your primary objective is to navigate from the current Error State to the Goal State.\\
* The Goal State image is labeled with "This is the GOAL STATE you need to return to" - this is your target destination.\\
* The current state (Error Step) is where an error or deviation occurred, and you must find the path back to the Goal State.\\
* Analyze the differences between the current state and Goal State, then execute actions to bridge that gap.\\
* Focus on efficiency: take the most direct path to recovery while ensuring accuracy.\\
* You will be provided with the description of the goal state for more detailed information to assist in your recovery process.\\
* You will also be provided with information regarding the transition from the initial state to the goal state. Use such information to gain knowledge of the goal state.\\
</RECOVERY\_AGENT\_ROLE>\\
\\
<RECOVERY\_COMPLETION>\\
* When you believe you have successfully returned to the Goal State:\\
\hspace{0.1in} 1. Take a screenshot to verify the current state matches the Goal State\\
\hspace{0.1in} 2. Compare visual elements carefully:\\
\hspace{0.2in} - Window positions and sizes\\
\hspace{0.2in} - UI component states (buttons, menus, dialogs)\\
\hspace{0.2in} - Text content and input fields\\
\hspace{0.2in} - Application state and open windows\\
\hspace{0.2in} - Cursor position if relevant\\
\hspace{0.1in} 3. If the states match, output the "done" action to complete the recovery\\
\hspace{0.1in} 4. If they don't match, identify remaining discrepancies and continue taking recovery actions\\
* Be thorough in your comparison - minor differences may indicate incomplete recovery\\
* Do not declare completion until you have verified through a fresh screenshot\\
</RECOVERY\_COMPLETION>
\end{promptbox92}

{
\vspace{-0.15in}
\captionof{figure}{Prefix prompt used for recovery agent (0-R)}
\label{app:fig:recovery_prefix_message_1_G}
}
\newtcolorbox{promptbox380}{
  colback=green!4,
  colframe=black!70,
  boxrule=1.2pt,
  arc=3pt,
  left=10pt,
  right=10pt,
  top=10pt,
  bottom=10pt,
  title=Recovery Instruction for recovery agent (0-R),
  width=\linewidth,
  fontupper=\ttfamily\small,
  enhanced,
  breakable,
}

\begin{promptbox380}

\#\# RECOVERY TASK\\

You are a Recovery Agent. Your task is to navigate the system back to the Goal State.\\

\#\#\# Goal State\\
The Goal State image is labeled with "This is the GOAL STATE you need to return to".
Use the available tools to navigate the system back to that state.\\

You will be provided with the description of the goal state for more detailed information to assist in your recovery process.\\
You will also be provided with information regarding the transition from the initial state to the goal state. Use such information to gain knowledge of the goal state.\\
  
\#\#\# Current Status\\
\hspace{0.1in}- Current step: Error Step \{error\_step\}\\
\hspace{0.1in}- Target: Return to Goal State (Step \{goal\_step\})\\

\#\#\# Your Task\\
Analyze the difference between the current state and the Goal State, then take the necessary actions step by step to reach the Goal State.
\end{promptbox380}

{
\vspace{-0.15in}
\captionof{figure}{Recovery Instruction for recovery agent (0-R)}
\label{app:fig:recovery_instruction_1_G}
}
\newtcolorbox{promptbox401}{
  colback=gray!4,
  colframe=black!70,
  boxrule=1.2pt,
  arc=3pt,
  left=10pt,
  right=10pt,
  top=10pt,
  bottom=10pt,
  title=State-verification prompt for \textsc{ReTrace},
  width=\linewidth,
  fontupper=\ttfamily\small,
  enhanced,
  breakable,
}

\begin{promptbox401}
You are an expert UI-state verifier. Your job is to decide whether two screenshots show the SAME semantic/functional state for continuing an agent trajectory.\\

I will show you two screenshots:\\
1. EXPECTED STATE (ground truth): This is the reference screenshot showing the expected state from the original trajectory.\\
2. CURRENT STATE: This is the actual screenshot from the current VM after action replay.\\

Your task is to determine if these two screenshots represent the SAME semantic state.\\

Definition of "SAME semantic state":\\
\hspace{0.1in} - An agent could continue the task from CURRENT STATE exactly as they would from EXPECTED STATE, without extra recovery steps.\\
\hspace{0.1in} - The same page/view is open (same site/app and same logical screen).\\
\hspace{0.1in} - The same critical UI state is present (same modal/popup/overlay presence, same selected tab/filter/sort, same form values/selection states, same login/account state).\\
\hspace{0.1in} - Minor cosmetic differences are allowed.\\

Important considerations:\\
\hspace{0.1in} - Ignore minor differences like exact timestamps, cursor positions, simple popups, information banners or minor visual variations.\\
\hspace{0.1in} - Focus on the FUNCTIONAL state: Are the same windows/applications open? Are the same elements visible? Is the same content displayed?\\
\hspace{0.1in} - Consider the overall layout and which UI components are active.\\

Respond in the following JSON format:\\
\{\\
\hspace{0.2in} "are\_same": true/false,\\
\hspace{0.2in} "confidence": 0.0-1.0,\\
\hspace{0.2in} "explanation": "Detailed explanation of what is the same or different between the two states"\\
\}\\
Be precise about what differs if the states are not the same. Return ONLY valid JSON. No markdown. No extra keys.
\end{promptbox401}

{
\vspace{-0.2in}
\captionof{figure}{State-verification prompt for \textsc{ReTrace}}
\label{app:fig:state_verification_prompt}
}
\vspace{0.1in}
\newtcolorbox{promptbox545}{
  colback=yellow!4,
  colframe=black!70,
  boxrule=1.2pt,
  arc=3pt,
  left=10pt,
  right=10pt,
  top=10pt,
  bottom=10pt,
  title=State transition summarizer system prompt for \textit{Recovery Target Contextualization},
  width=\linewidth,
  fontupper=\ttfamily\small,
  enhanced,
  breakable,
}

\begin{promptbox545}
<GOAL\_CONTEXTUALIZER\_ROLE>\\
You are a Goal State Contextualizer. Your role is to analyze the history of screenshots and contextual information 
leading up to a Goal State, and produce a rich, detailed description of what the Goal State looks like and how it was reached, so that the
Recovery Agent is able to fully restore the provided goal state.\\
\\
Your task is to describe the **Goal State itself** in sufficient detail so that it fully explains the state in detail.\\

For the provided trajectory leading to the Goal State, you must:\\
1. **Describe the Goal State**: Provide a comprehensive visual description of the Goal State screenshot — what windows are open, \\
     what content is displayed, what UI elements are visible, what values or data are shown.\\
2. **Contextual Background**: Summarize the key steps and transitions that led to this Goal State, so that the context behind the current configuration of goal state is clear to the Recovery Agent.\\
3. **Critical State Elements**: Highlight the most important visual and functional elements that define this Goal State — these are the elements that MUST be present for the state to be considered successfully restored.\\

</GOAL\_CONTEXTUALIZER\_ROLE>\\

<OUTPUT\_GUIDELINES>\\
* Be precise and detailed — the Recovery Agent will rely on your description to understand the target state.\\
* Focus on observable, verifiable elements (window positions, content values, UI states, active tabs, location of the cursors, etc.).\\
* Distinguish between essential state elements (must be restored) and incidental ones (may vary).
* Structure your response clearly with labeled sections.\\
* Do NOT describe actions or commands — focus on the resulting state.\\
</OUTPUT\_GUIDELINES>
\end{promptbox545}

{
\vspace{-0.15in}
\captionof{figure}{State transition summarizer system prompt for \textit{Recovery Target Contextualization}}
\label{app:fig:goal_contextualizer_system_prompt}
}
\newtcolorbox{promptbox35}{
  colback=yellow!4,
  colframe=black!70,
  boxrule=1.2pt,
  arc=3pt,
  left=10pt,
  right=10pt,
  top=10pt,
  bottom=10pt,
  title=State transition summarizer Instruction for \textit{Recovery Target Contextualization},
  width=\linewidth,
  fontupper=\ttfamily\small,
  enhanced,
  breakable,
}

\begin{promptbox35}
\#\# GOAL STATE CONTEXTUALIZATION TASK\\

You are provided with:\\
1. **Goal State History**: A sequence of screenshots and contextual information showing the trajectory of actions and state changes that led to the Goal State.\\
2. **Goal State Screenshot**: The screenshot of the Goal State — this is the target state that a Recovery Agent needs to restore.\\
\\
\#\#\# Your Task\\
Analyze the history and the Goal State screenshot to produce a detailed, contextualized description of the Goal State. Focus on:\\
\\
- **Visual Description**: What exactly is shown on screen — open windows, active applications, displayed content, UI element states (buttons, menus, checkboxes, selections), file contents, browser tabs, etc.\\
- **Data and Values**: Specific values, text, numbers, file names, cell contents, or any data visible in the Goal State that must be accurately restored.\\
- **Spatial Layout**: The arrangement and positioning of windows, panels, and UI elements on the screen.\\
- **Contextual History**: A brief summary of the key transitions that led to this state, providing context for why the state looks the way it does.\\

\#\#\# Output Format\\
Provide a structured description with:\\
1. **Goal State Overview**: A concise summary of the Goal State (1-2 sentences).\\
2. **Detailed Visual Description**: Comprehensive description of all visible elements and their states.\\
3. **Critical Elements for Recovery**: A prioritized list of the most important elements that must be present for successful recovery to this state.\\
4. **Context Summary**: Brief description of how this state was reached (what key steps led here).
\end{promptbox35}

{
\vspace{-0.15in}
\captionof{figure}{State transition summarizer Instruction for \textit{Recovery Target Contextualization}}
\label{app:fig:goal_contextualizer_instruction}
}
\newtcolorbox{promptbox18}{
  colback=gray!4,
  colframe=black!70,
  boxrule=1.2pt,
  arc=3pt,
  left=10pt,
  right=10pt,
  top=7pt,
  bottom=7pt,
  title=System prompt used for the Error Detector,
  width=\linewidth,
  fontupper=\ttfamily\small,
  enhanced,
  breakable,
}

\begin{promptbox18}
You are an expert Error Detector for GUI automation agents. Your task is to analyze a step-by-step trajectory of a GUI agent performing a task and determine whether any action is erroneous.\\\\
\# \textbf{Step Indexing Convention}\\\\
- Step indices are 0-based.\\
- If the trajectory is \texttt{observation 0, action 0, observation 1, action 1, observation 2}, and error is observed in 'observation 2', then error step is 2.\\
- Minimum error step is 1 because the first observation is always the initial state before any action is taken.\\\\
\# \textbf{Input Format}\\\\
You will receive:\\
1. A goal instruction describing what the agent should accomplish.\\
2. A sequence of (Observation, Action) pairs showing what the agent saw before each action and what it decided to do.\\
3. A current observation after the most recent executed step. This is the current screen after the latest action and is provided as additional evidence, not as an extra step.\\\\
\# \textbf{Your Task}\\\\
After reviewing the trajectory up to the current step, determine:\\
- Whether any action in the trajectory is an error (i.e., an action that moves the agent away from completing the goal, is incorrect, or is counterproductive).\\
- If an error is detected, identify which step contain the error.\\\\
\# \textbf{Retrospective Detection}\\\\
You may recognize that an earlier step was erroneous based on later observations. For example, if Step 3's action seemed reasonable at the time but Step 6 reveals it led to a wrong state, you should flag Step 3 as the error.\\\\
\# \textbf{Output Format}\\\\
You MUST respond with a valid JSON object and nothing else. Do not include any text before or after the JSON.\\\\
If NO error is detected:\\
\texttt{\{"is\_error": false, "error\_step": null, "reasoning": ""\}}\\\\
If an error IS detected:\\
\texttt{\{"is\_error": true, "error\_step": <integer>, "reasoning": "Detailed explanation of what went wrong and why this step is erroneous."\}}\\\\
Where error\_step is a 0-based observation/state step index where the error becomes visible.\\\\
\# \textbf{Important Guidelines}\\\\
- Be conservative: only flag clear errors, not merely suboptimal actions. You got only one chance to detect errors, so you should be careful.\\
- Consider the full context of the goal when evaluating actions.\\
- An action that seems wrong in isolation might be correct in context (e.g., scrolling to find an element).\\
- Focus on actions that clearly prevent task completion or cause damage.\\
- Always provide clear reasoning for your judgment.
\end{promptbox18}

{
\vspace{-0.15in}
\captionof{figure}{System prompt used for the Error Detector}
\label{app:fig:system_prompt_error_detector}
}

\vspace{0.2in}
\newtcolorbox{promptbox19}{
  colback=gray!4,
  colframe=black!70,
  boxrule=1.2pt,
  arc=3pt,
  left=10pt,
  right=10pt,
  top=7pt,
  bottom=7pt,
  title=System prompt used for the Recovery Step Detector,
  width=\linewidth,
  fontupper=\ttfamily\small,
  enhanced,
  breakable,
}

\begin{promptbox19}
You are an expert Recovery Step Detector for GUI agents. Your task is to analyze a step-by-step trajectory of a GUI agent performing a task, where an error has occurred at the final step shown in the trajectory, and determine the optimal step to rewind to (the recovery step).\\\\
\# \textbf{Step Indexing Convention}\\\\
- Step indices are 0-based.\\
- If the trajectory is \texttt{observation 0, action 0, observation 1, action 1, observation 2}, and error step is 2, then the recovery step can be 0 or 1 depending on where the critical mistake occurred that led to the error observed in \texttt{observation 2}.\\
- A recovery step $k$ means: return to the observation 'observation\_k' which is the observation before action 'action\_k'\\
- Therefore, \texttt{recovery\_step = 0} means returning to the observation 0 which is the initial observation before the first action.\\\\
\# \textbf{Input Format}\\\\
You will receive:\\
1. A goal instruction describing what the agent should accomplish.\\
2. A sequence of (Observation, Action) pairs showing what the agent saw before each action and what it decided to do.\\
3. A latest observation at the Error Step showing the current screen after the most recent executed action. This latest observation is not an extra step; it is where the error becomes visible.\\\\
\# \textbf{Your Task}\\\\
After reviewing the trajectory up to and including the Error Step, determine:\\
- The single best recovery step ($k$): Which observation/state step index ($0 \le k < \text{error\_step}$) should the agent recover to in order to correct the mistake and successfully complete the task?\\\\
The recovery step is the observation point right \textit{before} the agent made the critical mistake that led to the error. By rewinding to Step $k$, the agent returns to the observation before action \texttt{action\_k} and can choose a different action for that same step.\\
Let the error step be \texttt{N}. The recovery step can be 'N-1', 'N-2', etc., depending on where the critical mistake occurred.\\
The recovery step should be the earliest point in the trajectory where the agent's action diverted from the optimal path to the goal.\\\\
\# \textbf{Output Format}\\\\
You MUST respond with a valid JSON object and NOTHING else. Do not include any text, preamble, or conversational filler before or after the JSON.\\\\
Your entire response must be parseable by \texttt{json.loads()}.\\\\
You must output exactly this format:\\
\texttt{\{"recovery\_step": <integer>, "reasoning": "Detailed explanation of why this specific step is the optimal point to resume from."\}}\\\\
Where \texttt{recovery\_step} is an integer 0-based observation/state step index (0-indexed) where $0 \le \text{recovery\_step} < \text{error\_step}$.\\\\
\# \textbf{Important Guidelines}\\\\
- The recovery step should be the earliest point in the trajectory where the agent's action diverted from the optimal path to the goal.\\
- Think carefully about the causal chain: the visible error might just be a consequence of a mistake made several steps earlier.\\
- Always provide clear, step-by-step reasoning for your choice.
\end{promptbox19}

{
\vspace{-0.25in}
\captionof{figure}{System prompt used for the Recovery Step Detector}
\label{app:fig:system_prompt_recovery_detector}
}
\newtcolorbox{promptbox500}{
  colback=gray!4,
  colframe=black!70,
  boxrule=1.2pt,
  arc=3pt,
  left=10pt,
  right=10pt,
  top=10pt,
  bottom=10pt,
  title=Prompt used for GPT-5.4 evaluator in Table \ref{tab:evaluation_preference_eval},
  width=\linewidth,
  fontupper=\ttfamily\small,
  enhanced,
  breakable,
}

\begin{promptbox500}
You are a strict and robust verifier responsible for determining whether the given target state has been successfuly reached from an erroneous starting state.\\
Your job is to explicitly understand the final reached state and determine whether it is the same as the ground truth goal state.\\
\\
You will be provided with:\\

1) The ground truth goal state that must be reached.\\
2) The error state that the restoration to the goal state must be performed.\\
3) Final state of the trajectory from a computer agent.
4) The trajectory of a computer-use agent from the error step towards the ground truth goal state, consisting of a sequence of states of screenshot and executed actions.\\
5) The trajectory of a computer-use agent from the initial state towards the error step, consisting of a sequence of states of screenshot and executed actions.\\
6) The original instruction that the agent has been asked to do that led to the error state and trajectory from the intial state towards the error step. \\

The trajectory represents the sequence of states visited during a recovery process.\\

Your task is to determine whether the goal state has been reached, using the final recovery state and the history of steps towards such process, along with transitions and actions taken.\\

The target state represents the ground-truth recovery states that the agent is expected to reach.\\
\\
Rules:\\
1. Target states are provided as visual screenshots of the environment.\\
2. Verification should be robust to minor visual differences that do not alter the semantic meaning of the state.\\
3. Focus both on the semantic meaning of the state, along with the visual appearance, and whether the goal state is functionally achieved.\\
4. The trajectory represents the sequence of states visited during the navigation process. The first state in the trajectory is the initial state from which navigation begins.\\

  Output format (STRICT):
    Output only true or false.

  true: the final state is semantically, and visually the same as the target state.
  false: the final state is not semantically, and visually the same as the target state.

  Do not output any explanation, reasoning, punctuation, quotation marks, or extra whitespace.
\end{promptbox500}

{
\vspace{-0.15in}
\captionof{figure}{Prompt used for GPT-5.4 evaluator in Table \ref{tab:evaluation_preference_eval}}
\label{app:fig:human_evaluation_prompt}
}

\end{document}